\documentclass[final]{cvpr}

\usepackage{times}
\usepackage{epsfig}
\usepackage{graphicx}
\usepackage{diagbox}

\usepackage[caption=false]{subfig}
\usepackage{url}            
\usepackage{amsfonts}       
\usepackage{nicefrac}       
\usepackage{microtype}      
\usepackage{comment}
\usepackage{amsmath,amssymb,bm} 
\usepackage{multirow}
\usepackage{color,colortbl}
\usepackage{cuted}
\usepackage[ruled,vlined]{algorithm2e}

\definecolor{Gray}{gray}{0.9}

\newcommand{\figref}[1]{Fig.~\ref{fig:#1}}
\newcommand{\tabref}[1]{Table~\ref{tab:#1}}

\newcommand{\secref}[1]{Sec.~\ref{sec:#1}}
\newcommand{\appref}[1]{Appx.~\ref{sec:#1}}
\newcommand{\algref}[1]{Algorithm~\ref{alg:#1}}
\newcommand{\eq}[1]{\eqref{eq:#1}}

\newcommand{\xx}[0]{\mathbf{x}}
\newcommand{\X}[0]{\mathbf{X}}
\newcommand{\bR}[0]{\mathbb{R}}

\newcommand{\T}[0]{\mathbf{T}}
\newcommand{\bS}[0]{\mathbf{S}}

\newlength\savedwidth
\newcommand\whline[1]{\noalign{\global\savedwidth\arrayrulewidth
                               \global\arrayrulewidth #1} %
                      \hline
                      \noalign{\global\arrayrulewidth\savedwidth}}

\makeatletter
\renewcommand{\paragraph}{%
  \@startsection{paragraph}{4}%
  {\z@}{0.60ex \@plus 1ex \@minus .15ex}{-0.8em}%
  {\normalfont\normalsize\bfseries}%
}
\makeatother

\usepackage[pagebackref=true,breaklinks=true,colorlinks,bookmarks=false]{hyperref}

\def\cvprPaperID{7612} 
\def\confYear{CVPR 2021}

\begin{document}

\title{Linear Semantics in Generative Adversarial Networks}

\author{Jianjin Xu \thanks{Jianjin Xu is currently an assistant research scientist at Panzhihua University.}\\
Columbia University\\
{\tt\small jx2386@columbia.edu}
\and
Changxi Zheng\\
Columbia University\\
{\tt\small cxz@cs.columbia.edu}
}

\maketitle

\begin{abstract}
   Generative Adversarial Networks (GANs) are able to generate high-quality 
   images, but it remains difficult to explicitly specify the semantics of synthesized images.
   In this work, we aim to better understand the semantic representation of GANs, and 
   thereby enable semantic control in GAN's generation process.
   Interestingly, we find that a well-trained GAN encodes image semantics 
   in its internal feature maps in a surprisingly simple way: a linear 
   transformation of feature maps suffices to extract the generated image semantics.
To verify this simplicity, we conduct extensive experiments on various GANs and datasets; 
and thanks to this simplicity, we are able to learn a semantic segmentation model 
   for a trained GAN from a small number (e.g., 8) of labeled images.
   Last but not least, leveraging our findings, we propose two few-shot image editing
   approaches, namely Semantic-Conditional Sampling and Semantic Image
   Editing.  Given a trained GAN and as few as eight semantic annotations, the user is able to 
   generate diverse images subject to a user-provided semantic layout, and control 
   the synthesized image semantics. 
   We have made the code publicly available\footnote{https://github.com/AtlantixJJ/LinearGAN}.
\end{abstract}

\vspace{-2mm}
\section{Introduction}
Recent years have witnessed 
the striking success of Generative Adversarial Networks
(GANs)~\cite{goodfellow2014generative} in various image synthesis tasks:
to generate human faces, animals, cars, and interior scenes~\cite{radford2015unsupervised,denton2015deep,huang2017stacked,karras2019style}.
Apart from improving the generated image quality, recent research has been directed
toward the \emph{control} of GAN's image generation process\textemdash for example,
to enforce the generated images having user specified attributes, colors, and layouts.

Toward this goal, a fundamental question remains unanswered: 
how does a well-trained GAN encodes image semantics\textemdash such as hair, nose, and hats in a facial image \textemdash in its image generation process?
Motivated by this question, we aim to extract image semantics
from a GAN's internal data, namely its feature maps.
If we can well extract image semantics and understand the extraction process,
we can develop insight on how the image semantics are encoded.

Our finding is surprisingly simple: a linear transformation 
on the GAN's internal feature maps suffices to extract the generated
image semantics. In stark contrast to GAN's highly nonlinear image generation process,
this simple linear transformation is easy to understand and has a clear geometric interpretation (see \secref{linear_transform}).

We refer to this linear transformation process as linear semantic
extraction (LSE). To verify its performance, we conduct extensive experiments
on various GANs and datasets, including PGGAN~\cite{karras2017progressive},
StyleGAN~\cite{karras2019style} and StyleGAN2~\cite{karras2019analyzing}
trained on FFHQ~\cite{karras2019style}, CelebAHQ~\cite{liu2015faceattributes},
and LSUN~\cite{yu2015lsun}'s bedroom and church dataset.  We also compare the
performance of LSE with other semantic extraction approaches which
use learned nonlinear transformations.  It turns out that LSE is
highly comparable to those more complex, nonlinear models, suggesting that image semantics
are indeed represented in a linear fashion in GANs.




Related to our study of the linear encoding of image semantics in GANs is the
work of GAN Dissection~\cite{bau2018gan}.  It identifies feature maps that have
causal manipulation ability for image semantics.
Yet, most feature maps in that approach come from middle-level layers in the GAN, 
often having much lower resolution than the output image.  
Instead, we examine the GAN's internal feature maps collectively. 
We upsample all feature maps to the output image's resolution and stack them into a tensor.
This approach allows us to study per-pixel feature vectors, that is,
feature values corresponding to a particular pixel across all internal layers,
and we are able to classify every pixel into a specific semantic class.



The linear transformation in our proposed LSE is learned under supervision.
Its training requires image semantic annotations, which are automatically generated
using a pretrained segmentation model (such as UNet~\cite{ronneberger2015u}).
Interestingly, thanks to the linearity of LSE, even a small number of annotations suffice to train
LSE well. For example, the LSE trained with 16 annotated images 
on StyleGAN2 (which itself is trained on FFHQ dataset) achieves
88.1\% performance relative to a fully trained LSE model.
Not only does this result further support our finding about the linear representation of semantics in GANs,
it also inspires new approaches for controlling image generation through few-shot training.






In particular, we explore two controlled image generation tasks:
(1) \emph{Semantic Image Editing} (SIE)
and (2) \emph{Semantic-Conditional Sampling} (SCS).
The former aims to update images based on the user's edit on the semantics of
a GAN's output (e.g., generate images in which the hair region is reshaped);
the latter is meant to generate images subject to a user specification of 
desired semantic layout (e.g., produce images of an interior room where the furnitures
are laid out according to user specification).
We demonstrate few-shot SIE and SCS models both trained with small number 
of annotated images.



Behind both SCS and SIE is the core idea of matching the generated 
image semantics with a target semantic specification.
This is done by formulating an optimization problem, one that finds
a proper latent vector for the GAN's image generation while respecting the user specification.
We also consider baselines of both tasks, which are implemented by using 
carefully trained, off-the-shelf semantic segmentation models rather than our few-shot LSE.
In comparison to the baselines, our approach with 8-shot LSE is able to generate 
comparable (and sometimes even better) image quality.


In summary, our technical contributions are twofold:
\textbf{(i)} Through extensive experiments, we show that GANs represent the image's pixel-level semantics in a linear fashion.
\textbf{(ii)} We propose an LSE with few-shot learning, which further enables two 
image synthesis applications with semantic control, namely SCS and SIE under {few-shot}
settings.
\section{Related work}\label{sec:rel}


\textbf{Generative Adversarial Networks}.
GANs~\cite{goodfellow2014generative} have achieved tremendous success in image
generation tasks, such as synthesizing photo-realistic facial images~\cite{karras2017progressive,karras2019style,karras2019analyzing}, 
cityscapes \cite{2017High,park2019gaugan} and ImageNet images~\cite{zhang2018self,brock2018large}.  Among various
GAN models, Progressively Grown GAN (PGGAN) ~\cite{karras2017progressive},
StyleGAN~\cite{karras2019style} and its improvement 
StyleGAN2~\cite{karras2019analyzing} are three of the most widely used
GAN structures. 
PGGAN shares a similar architecture as the Deep Convolution GAN
(DCGAN)~\cite{radford2015unsupervised}, trained progressively.  StyleGAN
adopts the adaptive instance normalization~\cite{2017Arbitrary} from neural
stylization literatures to improve generation quality.
Further improving on StyleGAN, StyleGAN2 is by far the state-of-the-art GAN model on various datasets.
We therefore conduct experiments on the three types of GANs.

\textbf{Interpreting GANs}.
Our study of image semantics in GAN models is related to the works toward interpreting and dissecting GANs.
Along this research direction,
existing methods can be grouped into two categories.  First are those
aiming to interpret a GAN's latent space.  Prior works~\cite{shen2019interpreting,yang2019semantic} 
find that there exist linear
boundaries in latent space that separate positive and negative 
attributes of image samples.  Others works~\cite{2020Closed,jahanian2019steerability,voynov2020unsupervised} 
propose to 
find linear trajectories of attributes in the latent space in an unsupervised way.  
Second, interpreting the feature maps of GANs.
GAN Dissection \cite{bau2018gan} identifies convolution units that have causality with semantics in the generated images.
Collins et al. \cite{Collins20} find that the clusters of semantics can be found by k-means and matrix factorization in GAN's feature maps.
Our differences are two-fold.
First, we study the high-resolution semantic masks extracted from the generator, which is rarely touched in existing works.
Second, the SIE and SCS applications derived from our discoveries are novel in terms of their few-shot settings.

\textbf{Controlling GANs}. 
Methods to enable GAN's controllability can be divided into two streams.
First, training new GANs with architectures specifically designed to enable controllability.
Conditional GAN (cGAN) and its variant \cite{odena2016cgan,dumoulin2016adversarially,chen2016infogan} are proposed to enable GAN's controllability for category.
StackGAN \cite{zhang2017stackgan} extends cGAN by using the embedding of natural language to control the synthesis.
The image-to-image translation networks can map semantic masks to images \cite{Zhu2017Unpaired,Isola2017pix,park2019gaugan,2019SEAN}.
They allow explicit control of semantic structures but need expensive labeled data for training.
Second, interpreting or devising auxilary architectures to exploit the controllability of pretrained GAN models.
The controllability for \emph{global attributes} is studied by many interpretation-based editing methods \cite{shen2019interpreting,yang2019semantic,2020Closed,jahanian2019steerability,voynov2020unsupervised}.
Besides interpretation, other methods propose auxilary networks for the controllability for attributes \cite{adbal2020styleflow} or 3D characteristics \cite{ghosh2020gif,tewari2020stylerig,zhang2020image}.
The controllability for \emph{local image editing} also receives much research attention.
The latent code optimization methods \cite{Zhu2016Generative,Brock2017NeuralPE} can make the image resemble the color strokes drawn by users, but the precision of editing is limited.
The feature map substitution methods \cite{Rameen2019edit,Suzuki2018collaging,collins2018} can edit a localized region of an image precisely, but the editing operation requires users to find a source image for reference.
GAN Dissection \cite{bau2018gan} succeed in editing the semantics of images, but its resolution and diversity are limited.
Bau et al. \cite{bau2020rewriting} rewrite the weight of a generator to change its generation pattern.

The semantic controllability studied in our work differs from previous works in two aspects.
First, previous SCS models in the context of image-to-image translation require extensively labeled images, 
whereas our approach requires only a few annotations in its training.
Second, previous SIE models (such as~\cite{bau2018gan}) are mainly concerned with the control of semantic morphology,
not the user's ability to fully specify semantic regions. 
As a result, our approach requires no reference image, and thereby eases the user editing process.


\begin{figure*}
    \centering
    \includegraphics[width=0.84\linewidth]{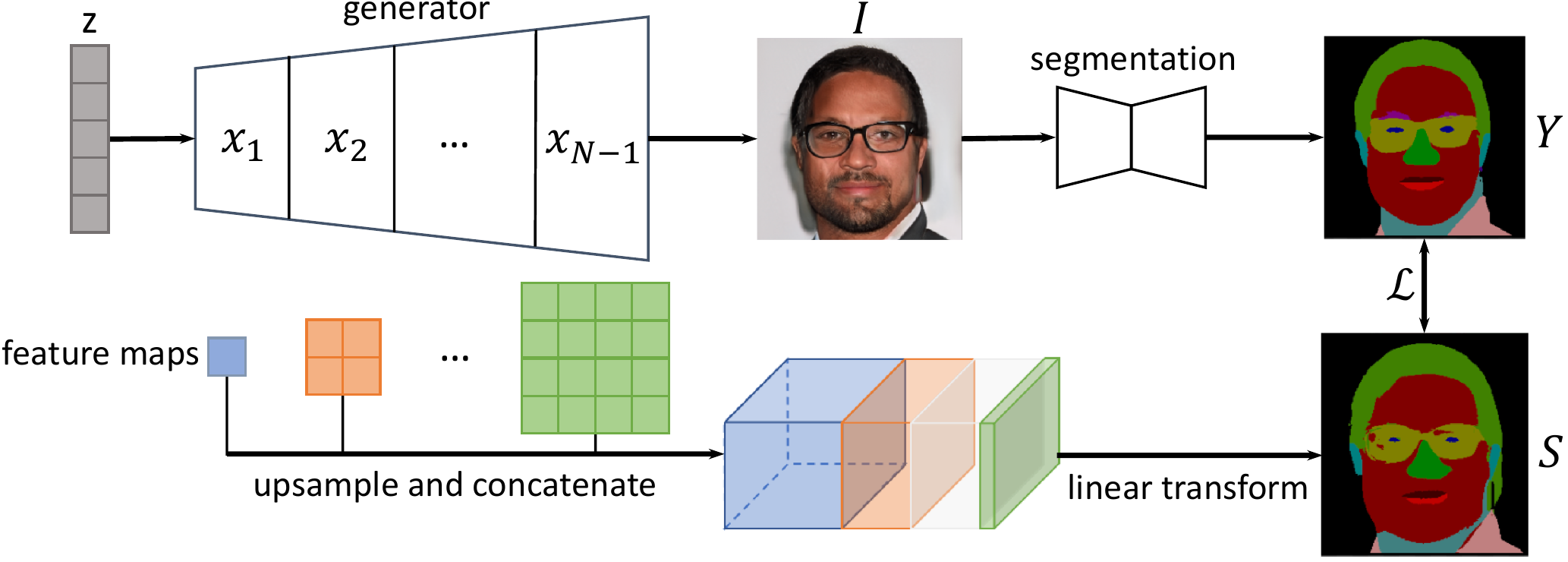}
    \caption{
    When synthesizing an image $I$ from a latent vector $z$, the generator
builds a series of internal feature maps. 
Provided a well-trained GAN model,
we decode its
feature maps $\{\xx_i\}_{i=1}^{N-1}$ to extract the output image's semantic
segmentation $\mathbf{S}$.  This is done by learning a simple linear
transformation applied on the feature maps. 
Learning the linear transformation is supervised by a pretrained segmentation model.}
    \label{fig:se_pipeline}
    \vspace{-1mm}
\end{figure*}

\section{GAN's Linear Embedding of Semantics}\label{sec:linear}







We aim to decode a GAN's internal representation of image semantics in its
image synthesis process.  Our finding is surprisingly simple: a \emph{linear
transformation} on the GAN's feature maps suffices to reveal its synthesized
image semantics.  In this section, we first construct such a linear
transformation (\secref{linear_transform}), and then conduct experiments
(\secref{expr_setup}) to examine its competence for revealing image semantics
(\secref{suffice}).

\subsection{Linear Transformation on Feature Maps}\label{sec:linear_transform}
A well-trained GAN model maps a randomly chosen latent vector to a realistic image.
Structurally, a GAN model concatenates a series of network layers. 
Provided a latent vector, each layer $i$ outputs a feature map $\xx_i$, which is in turn fed into the next layer. 
We denote the width, height, and depth of $\xx_i$ using $w_i$, $h_i$ and $c_i$, respectively (i.e., $\xx_i\in\bR^{c_i\times w_i\times h_i}$).


It is unsurprising at all that one can deduce from the feature maps
the generated image semantics. After all, feature maps represent the GAN's
internal data flow that results in the final image.  As images can be
semantically segmented using pretrained networks, the feature map can also be
segmented with appropriate networks.  More interesting is the question of how
\emph{easily} we can learn from feature maps about the generated image
semantics.  A straightforward relation between feature maps and image semantics
could be easy to understand, and inspire new theories and applications.

\paragraph{Objective.}
Consider a GAN model consisting of $N$ layers and producing images with $m$ semantic classes (such as hair, face, and cloth). 
We seek the simplest possible relation between its feature maps and output image semantics\textemdash a linear transformation matrix $\T_i$ applied to each feature map $\xx_i$ to predict a semantic map of the layer $i$. 
By accumulating all the maps, we wish to predict a semantic segmentation $\bS$ of the GAN's output image (see ~\figref{se_pipeline}).
Formally, $\bS$ is just a linear transformation of all feature maps, defined as
\begin{equation}\label{eq:linear}
    \mathbf{S} = \sum_{i=1}^{N-1} \mathsf{u}_{i}^\uparrow(\mathbf{T}_i\cdot\xx_i),
\end{equation}
where $\T_i\in \bR^{m\times c_i}$ converts $\xx_i\in\bR^{c_i\times w_i\times
h_i}$ into a semantic map $\mathbf{T}_i\cdot\xx_i\in\bR^{m\times w_i\times
h_i}$ through a tensor contraction along the depth axis. The result from each
layer is then upsampled (denoted by $\mathsf{u}_{i}^\uparrow$) to the output image
resolution.  The summation extends over all internal layers, excluding the 
last layer (layer $N$), which outputs the final image.  The result
$\bS\in\bR^{m\times w\times h}$ has the same spatial resolution $w\times h$ as
the output image.  Each pixel $\bS_{ij}$ is a $m\times 1$ vector, indicating
the pixel's unnormalized logarithmic probabilities representing each of the $m$
semantic classes. We refer to this method as Linear Semantic Extractor
(LSE).

\paragraph{Optimizing $\T_i$.}
The training process of LSE is supervised by pixel-level annotation of semantics.
Yet, it is impractical to manually annotate a large set of images that are automatically generated by a GAN model. 
Instead, we leverage off-the-shelf pretrained segmentation models for semantic annotation.
In practice, we use UNet~\cite{ronneberger2015u} to segment facial images (into 
the nose, eye, ear, and other semantic regions), and 
DeepLabV3~\cite{2017Rethinking} with ResNeSt backbone~\cite{2020ResNeSt} for bedroom and church images.

Concretely, provided a well-trained GAN model, we randomly sample its latent space to produce a set $\mathbb{S}$ of synthetic images.
When synthesizing every image in $\mathbb{S}$, we also record the model's feature maps $\{\xx_i\}_{i=1}^{N-1}$.
These feature maps are linearly transformed using \eq{linear} to predict a
semantic mask of the image, which is then compared with the result from the
pretrained semantic segmentation network to form the standard cross-entropy
loss function:
\begin{equation*}\label{eq:loss}
    \mathcal{L} = \frac{1}{w\cdot h} \sum_{ \substack{1\le i \le w \\1\le j\le h}}
    \left[ \log \left( \sum_{k=1}^m \exp\left(\bS_{ij}[k]\right)\right)- \bS_{ij}[Y_{ij}] \right],
\end{equation*}
where $Y_{ij}$ is the semantic class at pixel $(i,j)$ indicated by the
supervisor network, and $\bS_{ij}[k]$ is the corresponding unnormalized logarithmic 
probability for the $k$-th semantic class predicted by the LSE.

Lastly, the linear matrices $\T_i$ are optimized by minimizing the expected loss (estimated by taking the average loss over image batches in $\mathbb{S}$).
Details of the training process are provided in \appref{setup}.

\paragraph{Geometric picture.}\label{sec:geo}

\begin{figure}[t]
    \centering
    \subfloat[upsampled feature maps $\X$\label{fig:concat_b}]{
        \includegraphics[width=0.48\linewidth]{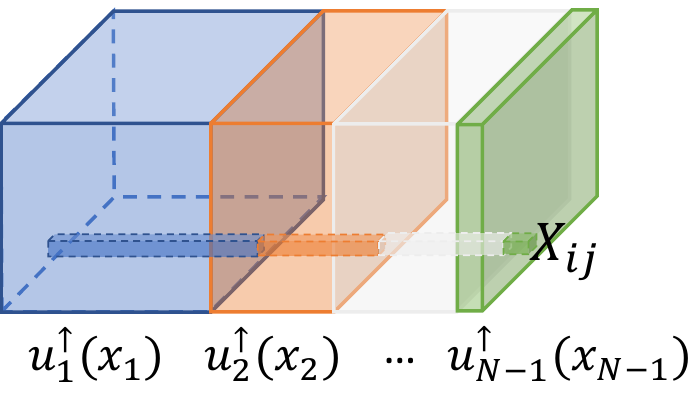}}
    \subfloat[dot product $\mathbf{S}_{ij} = \T \cdot \X_{ij}$\label{fig:concat_c}]{
        \includegraphics[width=0.48\linewidth]{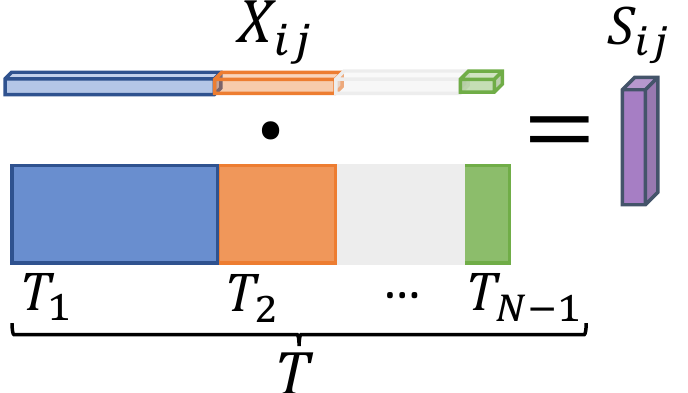}}
    \vspace{1mm}
    \caption{The visualization of the linear transformation, with upsample placed before convolution.}
    \vspace{-3mm}
\end{figure}

The linear relation \eq{linear} allows us to draw an intuitive geometric
picture of how image semantics are encoded in the generator's feature maps.

First, notice that $\T_i$ applied on $\xx_i$ can be viewed as a $1\times1$ convolutional filter with stride 1.
The filter operation is commutative with the upsample operation $\mathsf{u}_{i}^\uparrow(\cdot)$ 
(see \appref{comm} for a proof of this commutative property).
As a result, we can rewrite the semantic prediction $\bS$ in \eq{linear} as
\begin{equation}\label{eq:geo_interp}
    \mathbf{S} = \sum_{i=1}^{N-1} \mathbf{T}_i\cdot\mathsf{u}_{i}^\uparrow(\xx_i)
     = \T \cdot \X,
\end{equation} %
where $\T = \begin{bmatrix} \T_1 & \dots & \T_{N-1} \end{bmatrix}$ 
is an $m\times n$ matrix with $n=\sum_{i=1}^{N-1}c_i$ being the total layer
depth.  $\X\in\bR^{n\times w\times h}$ is a tensor concatenating all upsampled
$\xx_i$ (i.e., $\mathsf{u}_{i}^\uparrow(\xx_i)$ with resolution $c_i\times
w\times h$) along the depth axis (see \figref{concat_b}).

Now, consider a pixel $(i,j)$ in the output image.
To predict its semantic class, Equation \eq{geo_interp} shows that we can take
the corresponding $n \times 1$ vector $\X_{ij}$ that stacks the pixel's
features resulted from all GAN layers, and dot product it with each row of $\T$
(see \figref{concat_c}): $\mathbf{S}_{ij} = \T \cdot \X_{ij}$.  In other words,
each row $\T^{(k)}$ of $\T$ defines a direction representing the semantic class
$k$ in the $n$-dimensional feature space.

If the linear transformation can classify features with high accuracy, then 
the feature vectors of different semantic classes are linearly
separable. Define the set of all vectors that are classified into class $k$ as
\begin{equation}\label{eq:region}
    \mathcal{R}_k = \{\bm{x} | \T^{(k)} \bm{x} > \T^{(j)} \bm{x}, \forall j \neq k\},
\end{equation}
where $\T^{(k)}$ is the k-th row of the tensor $\T$.
This definition shows that the subspace of each semantic class forms a
hyper-cone originating from the origin.

An intuitive geometric picture is as follows.
Consider a unit $n$-sphere at the origin.
The intersection of a semantic class $i$'s hyper-cone and the sphere surface encloses a convex area $A_i$.
Then, take the feature values at a pixel and normalize it into a unit vector.
If that vector falls into the convex area $A_i$, then the pixel is classified as the class $i$.
In other words, the surface of $n$-sphere is divided into $k$ convex areas, each representing a
semantic class.
From this geometric perspective, we can even infer a pixel's semantic class without training the linear model~\eq{linear}.
Rather, we locate a semantic \emph{center} $c_i$ for each convex area $A_i$ on the $n$-sphere surface.
For example, the semantic centers can be estimated by a clustering algorithm (such as $k$-means clustering).
A pixel is classified as class $i$ if its feature vector is closest to $c_i$ (among all semantic centers) 
on the $n$-sphere.
In \secref{geo_expr}, we show that the class centers
can segment images reasonably well, supporting our hyper-cone interpretation.

\paragraph{Nonlinear semantic extraction.}

If LSE can extract the semantics of generated images, a further question is to
what extent the semantics can be better extracted by nonlinear models.  The
answer to this question provides further support on whether or not feature maps
in GANs indeed encode image semantics linearly.  If they do,
then nonlinear models would perform \emph{no} significantly better than
our linear model.

We propose two nonlinear extraction models for this study.
The first Nonlinear Semantic Extractor (which we referred to as NSE-1) 
transforms each layer's feature maps through three convolutional layers interleaved with ReLU activations. Each transformed feature map 
is upsampled using the same interpolation
$\mathsf{u}_{i}^\uparrow(\cdot)$ as in~\eq{linear}.
The second model (NSE-2) transforms feature maps into hidden layers and refines
them as the resolution increases, resembling the approach in 
DCGAN~\cite{radford2015unsupervised}. See \appref{nonlinear} for
details of both models.

There are other nonlinear models---for example, one that concatenates a
generative model with a full-fledged semantic segmentation model (such as
UNet~\cite{ronneberger2015u}).  However, such a model provides no clue about
how feature maps encode image semantics.  We therefore choose not to consider
them in our studies.

\subsection{Experiment Setup} \label{sec:expr_setup}
We conduct experiments on various GANs and datasets to 
examine our LSE model.
We choose PGGAN \cite{karras2017progressive}, StyleGAN \cite{karras2019style},
and StyleGAN2~\cite{karras2019analyzing} trained on three datasets.
Specifically, we use StyleGAN trained on the facial image dataset CelebAHQ~\cite{liu2015faceattributes},
and StyleGAN2 trained on FFHQ~\cite{karras2019style} and
separately on a bedroom and church dataset from LSUN~\cite{yu2015lsun}.
Instead of training those GAN models from scratch,
we use the existing pretrained GANs\footnote{These pretained GANs 
are publicly available \href{https://github.com/genforce/genforce}{here}.}.


Training LSE is supervised by pretrained semantic segmentation models.  For
facial images, we use a UNet trained on CelebAMask-HQ~\cite{lee2019maskgan}
with manually labeled semantic masks, and it segments a facial image into 15
semantic regions.  For bedroom and church images, we use the publicly
available DeepLabV3~\cite{2017Rethinking} trained on ADE20K~\cite{2017Scene}
dataset.  
DeepLabV3 predicts 150 classes, most of which are not present in the GAN's output images. 
Thus, we consider a subset of classes for generated bedroom and church images.
Our choice of the classes is described in \appref{category_selection}.

In each experiment, the training data of LSE consists of 51,200 images sampled
by a GAN model (i.e., PGGAN, StyleGAN, or StyleGAN2),
and the images are semantically labeled by a pretrained segmentation model.
Meanwhile, we record the GAN model's feature maps in each image generation.  
The semantic masks together with the feature maps are then used
to train the transformation matrix $\T_i$ for every GAN layer (see \appref{setup} for more details).  

After training, we evaluate our LSE on a separate set of 10,000 generated images.
During the generation of each image, we use LSE (and NSE-1 and NSE-2 for comparison) on
the generator's feature maps to predict a semantic segmentation,
which is in turn compared with the segmentation labels to compute an IoU score (defined in \appref{iou}). 

\subsection{Results}\label{sec:suffice}
We now present empirical results to back our proposed linear semantic extraction~\eq{linear}.

\begin{figure}[h]
    \centering
    \includegraphics[trim=20 10 20 10,clip,width=\linewidth]{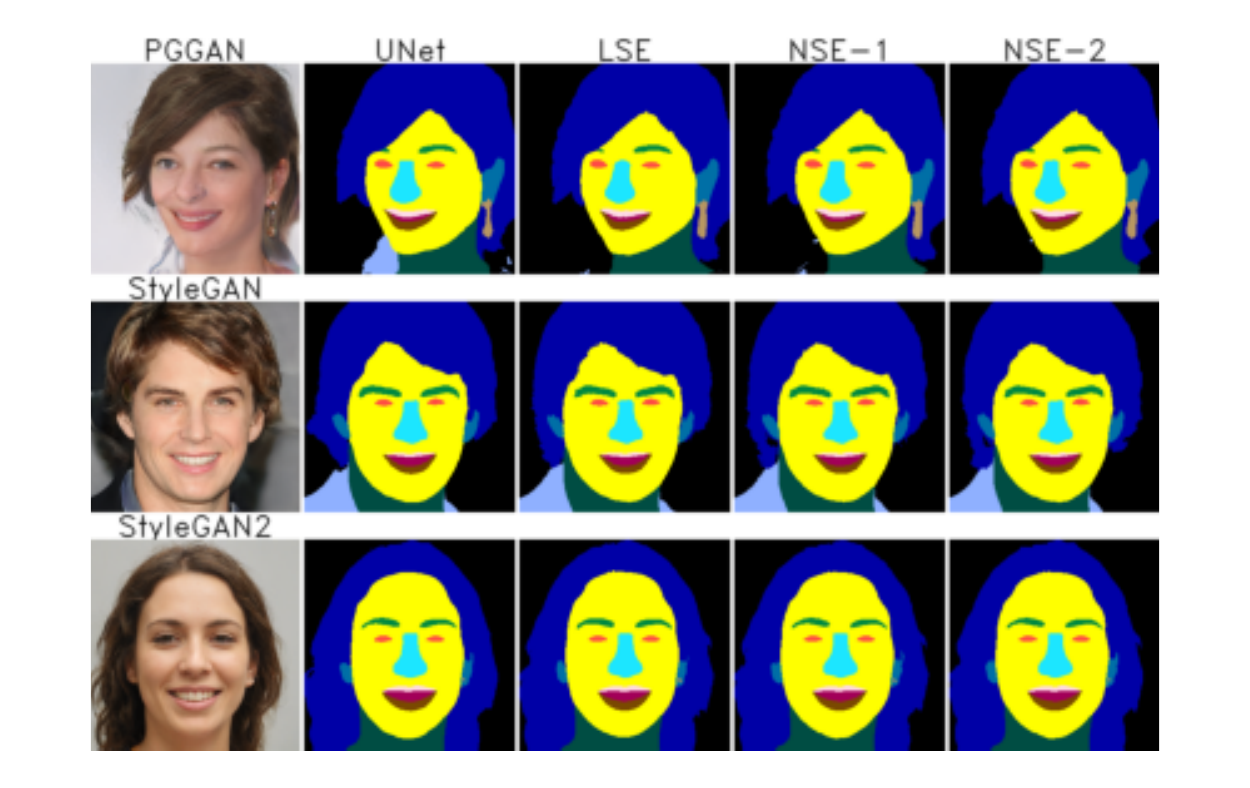}
    \includegraphics[trim=20 10 20 10,clip,width=\linewidth]{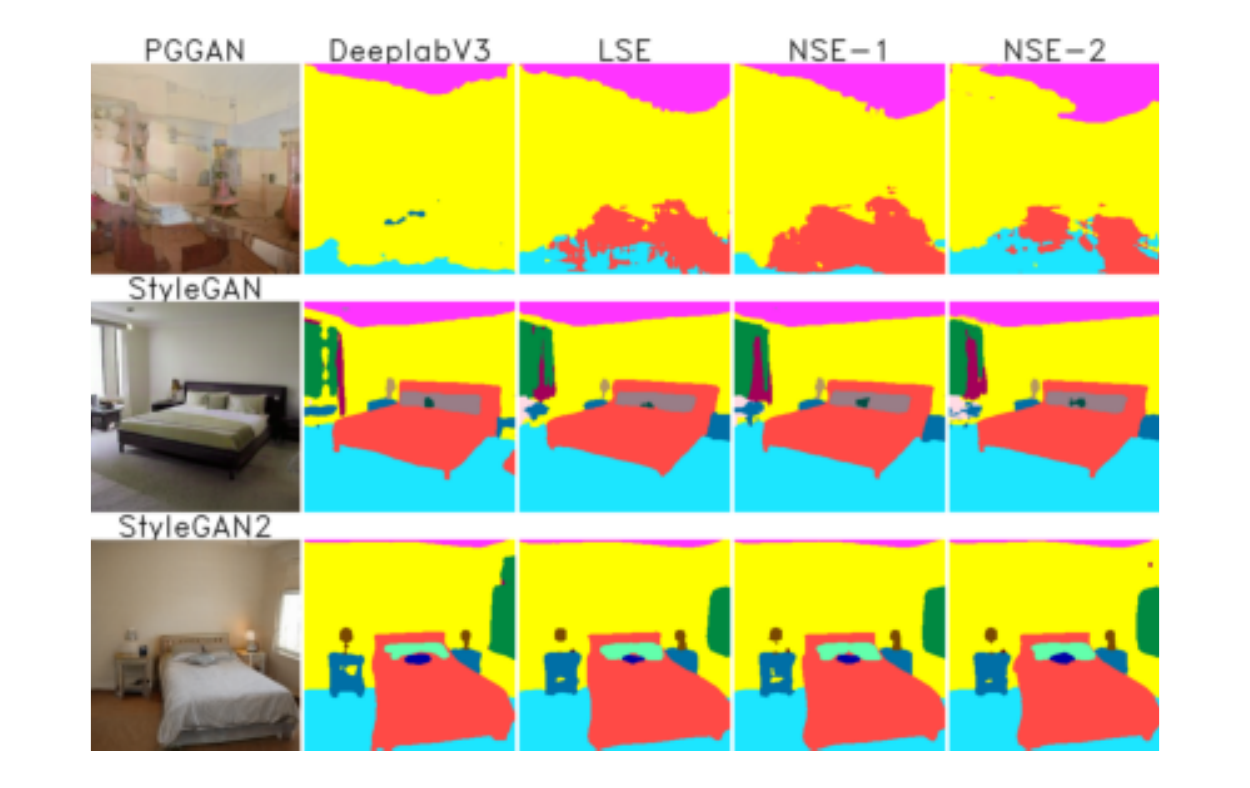}
    \includegraphics[trim=20 10 20 10,clip,width=\linewidth]{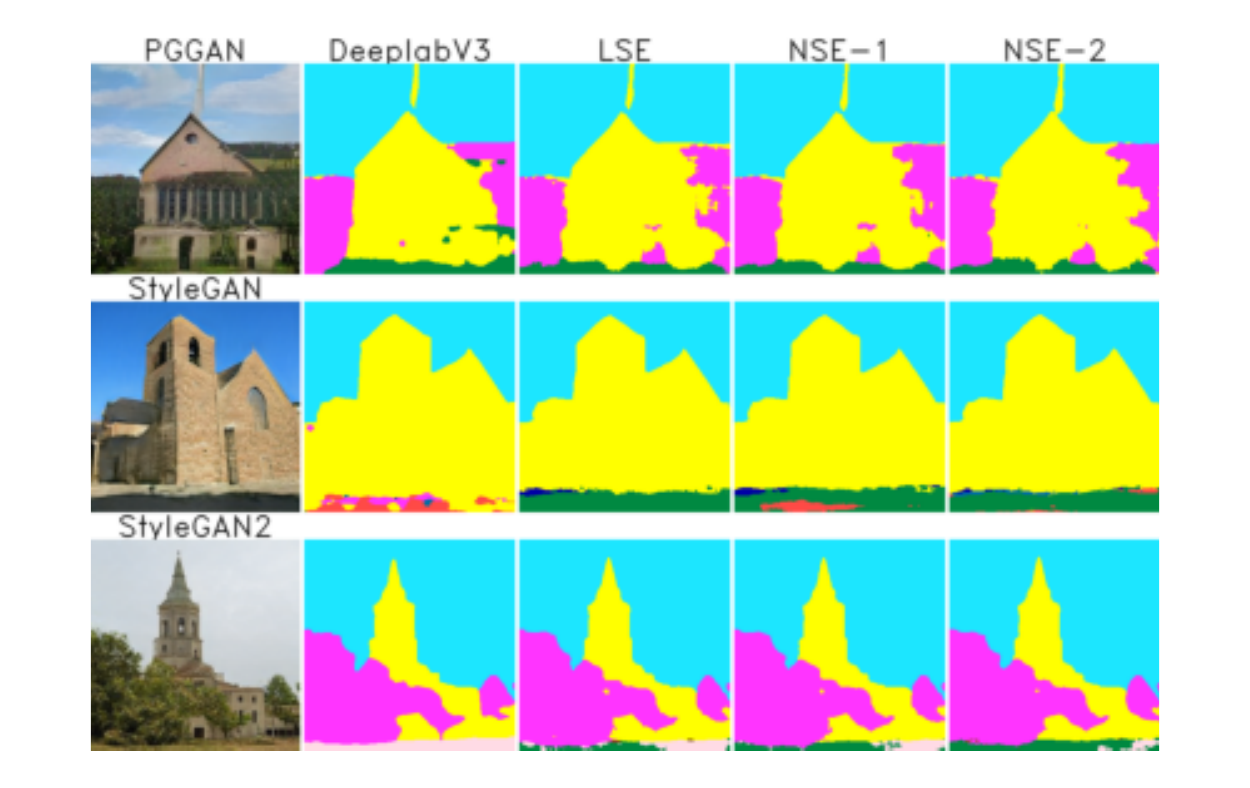}
    \vspace{-0.5cm}
    \caption{Qualitative comparison of LSE, NSE-1 and NSE-2. From top to bottom, every 3 rows are from GAN models trained on the same dataset (face, bedroom, church images, respectively). Images are sampled randomly rather than cherry-picked.}
    \label{fig:qualitative}
    \vspace{-2mm}
\end{figure}

\begin{table*}
    \centering
    \resizebox{\linewidth}{!}{
    \begin{tabular}{c|ccc|ccc|ccc}
    \whline{1.0pt}
     & \multicolumn{3}{c|}{PGGAN} & \multicolumn{3}{c|}{StyleGAN} & \multicolumn{3}{c}{StyleGAN2} \\
    Dataset & CelebAHQ & Bedroom & Church & CelebAHQ & Bedroom & Church & FFHQ & Bedroom & Church \\\hline
    LSE & 65.5 (-1.6) & 33.2 (-3.2) & 51.3 (-3.2) & 69.1 (-1.9) & 39.9 (-7.8) & 35.4 (-6.3) & 79.7 (-1.7) & 53.9 (-3.4) & 37.7 (-2.6) \\
    NSE-1 & \textbf{66.5} & \textbf{34.3} & \textbf{53.0} & \textbf{70.5} & \textbf{43.3} & \textbf{37.8} & \textbf{81.0} & \textbf{55.8} & \textbf{38.7} \\
    NSE-2 & 65.9 (-0.9) & 30.7 (-10.5) & 49.5 (-6.6) & 70.1 (-0.5) & 38.9 (-10.2) & 34.0 (-10.1) & 80.2 (-1.1) & 52.1 (-6.8) & 35.3 (-8.8) \\
    \whline{1.0pt}
    \end{tabular}}
    \caption{\normalsize The mIoU (\%) of LSE, NSE-1, and NSE-2 
        trained with off-the-shelf semantic segmentation models 
        (UNet for CelebAHQ and FFHQ, DeepLabV3 for bedroom and church dataset). 
        ``Bedroom'' and ``Church'' images are subsets of the LSUN \cite{yu2015lsun} dataset. 
    The numbers in brakets are the performance difference relative to the best model highlighted in bold.}
    \label{tab:quant_se}
    \vspace{-3mm}
\end{table*}


\paragraph{Evaluation of LSE.}
Figure~\ref{fig:qualitative} compares qualitatively semantic segmentation of LSE to other methods.
The quantitative results in terms of mIoU scores are reported in \tabref{quant_se},
from which it is evident that our simple LSE
is comparable to more complex, nonlinear semantic classifiers.
The relative performance gap between LSE and NSEs (NSE-1 and NSE-2) is within 3.5\%.
Results on StyleGAN-Bedroom and StyleGAN-Church have a slightly larger gap ($< 8$\%).
We present additional qualitative results and IoU for each category in \appref{supp_results}.

\textit{Takeaway.}
Our experiments show that LSE is capable of extracting image semantics from
the feature maps of the GANs.
Further, the close performance of LSE to NSEs suggests that a well-trained GAN encodes
the image semantics in its feature maps in a linear way.

Our approach differs from the prior GAN Dissection work~\cite{bau2018gan}, which 
identifies feature maps correlating with a specific semantic class.  These
feature maps are primarily found in middle-level feature maps, resulting in a
lower resolution segmentation than the network output.  Also, the
per-pixel semantic classification remains unexplored.  
In contrast, the
semantics extracted by LSE are of high resolution (the same as the output
image) and have sharp boundaries.  

\paragraph{Geometrical evidence.} \label{sec:geo_expr}
The geometric interpretation of \eq{linear} indicates that
features of a semantic class fall into a convex surface area on an $n$-sphere.
To verify this intuition, we test a stronger hypothesis \textemdash 
the features of individual pixels can be clustered around class centers.
If the clusters are well-formed, we should be able to find a convex hull to identify individual classes.

\begin{figure}[t]
    \centering
    \includegraphics[width=\linewidth]{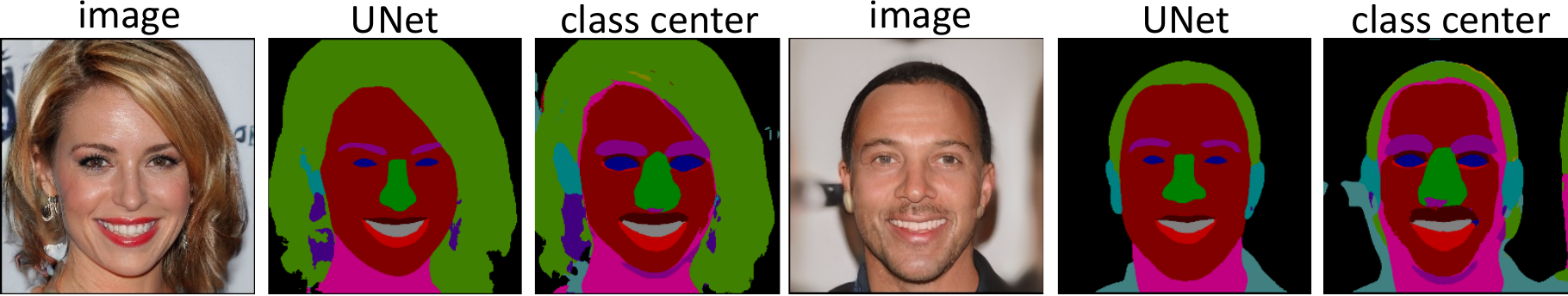}
    \caption{Forging LSE's parameter $\T$ using the statistical centers of features. Experiment is done on StyleGAN-CelebAHQ.}
    \label{fig:mean_weight}
    \vspace{-3mm}
\end{figure}

To estimate the class centers, we randomly generate 3000 images using
StyleGAN-CelebAHQ, and obtain their semantic masks using UNet.  
All per-pixel feature vectors
from the same semantic class are collected and normalized onto the unit $n$-sphere.
The vectors are then averaged and renormalized on the $n$-sphere. The resulting vector is then treated 
as a class center to determine each pixel's semantic class.
Some segmentation results are shown in \figref{mean_weight}, 
suggesting that this approach indeed segments images reasonably. 
The segmentation error (e.g., in \figref{mean_weight}) may be attributed to the inaccurate boundaries
between classes, as they are not explicitly trained to separate different
semantic classes.  Nevertheless, this experiment
confirms our geometric intuition about the feature maps' linear embedding of semantics.

We further compute the statistics of cosine similarities of feature vectors
that are within the same semantic class and that are in different classes.  
We show that feature vector's cosine similarities between pixels within the same class are indeed
higher.  Details are reported in \appref{cosine_similarity}.

\textit{Takeaway.}
Statistical centers of feature vectors can segment images reasonably well,
suggesting that feature vectors from different classes are well separated on the $n$-sphere. 
The relatively larger cosine similarities between different classes also backs our intuition.
These are further evidence indicating linear encoding of image semantics in
the feature maps of GANs.

\begin{table}[b]
    \vspace{-2mm}
    \centering
    \resizebox{\linewidth}{!}{
    \begin{tabular}{c|ccc}
    \whline{1.0pt}
    N & FFHQ & Bedroom & Church \\\hline
    1 & 55.6 (69.8) $\pm$ 5.2 & 21.5 (39.8) $\pm$ 3.7 & 19.7 (52.2) $\pm$ 3.4 \\
    4 & 64.8 (81.4) $\pm$ 1.0 & 36.5 (67.8) $\pm$ 2.7 & 24.2 (64.3) $\pm$ 1.4 \\
    8 & 68.4 (85.8) $\pm$ 2.6 & 38.6 (71.6) $\pm$ 2.4 & 26.3 (69.7) $\pm$ 0.8 \\
    16 & 70.2 (88.1) $\pm$ 3.0 & 42.2 (78.3) $\pm$ 1.1 & 27.7 (73.5) $\pm$ 0.8 \\\hline
    full & 79.7\% & 53.9\%  & 37.7\% \\
    \whline{1.0pt}
    \end{tabular}}
    \vspace{1mm}
    \caption{The evaluation of few-shot LSEs for StyleGAN2. Each model is trained 5 times.
    Both the mean and maximum deviation of the 5 repeats are shown.
    The numbers in parentheses indicate the ratio of the mean performance over the fully trained model's performance listed in the last row.
    }
    \label{tab:fewshot_eval}
\end{table}

\paragraph{Few-shot LSEs.} \label{sec:fewshot_LSE}
Thanks to the linearity of LSE, we can also train it using a small number of examples. 
We refer to this approach as the \emph{few-shot} LSE.
Here, we experiment the training of LSE with only 1, 4, 8, and 16 annotated images, respectively.
For each few-shot LSE setup, the training is repeated for five times, and we report the average 
performance of the five trained models. 

Table~\ref{tab:fewshot_eval} reports the quantitative evaluation results.
First, the extreme case, one-shot LSE, already shows plausible performance, achieving 69.8\%, 39.8\%, and 52.5\% mIoU scores relative
to the fully trained model.  
The 16-shot LSE further improves the mIoU scores to
88.1\%, 78.3\%, and 73.5\% relative to the fully trained model.  

\textit{Takeaway.}
The few-shot LSEs have already achieved performance comparable to fully supervised LSEs.
Not only do they enable a low-cost way of extracting semantics from GANs, 
the results further support our hypothesis that image semantics are linearly embedded in feature maps.


\section{Applications}\label{sec:app}
In this section, we leverage the simplicity of LSE to control 
image semantics of GAN's generation process.

\subsection{Few-shot Semantic Editing}\label{sec:sie}

\begin{figure*}[t]
  \centering
  \includegraphics[width=0.96\linewidth]{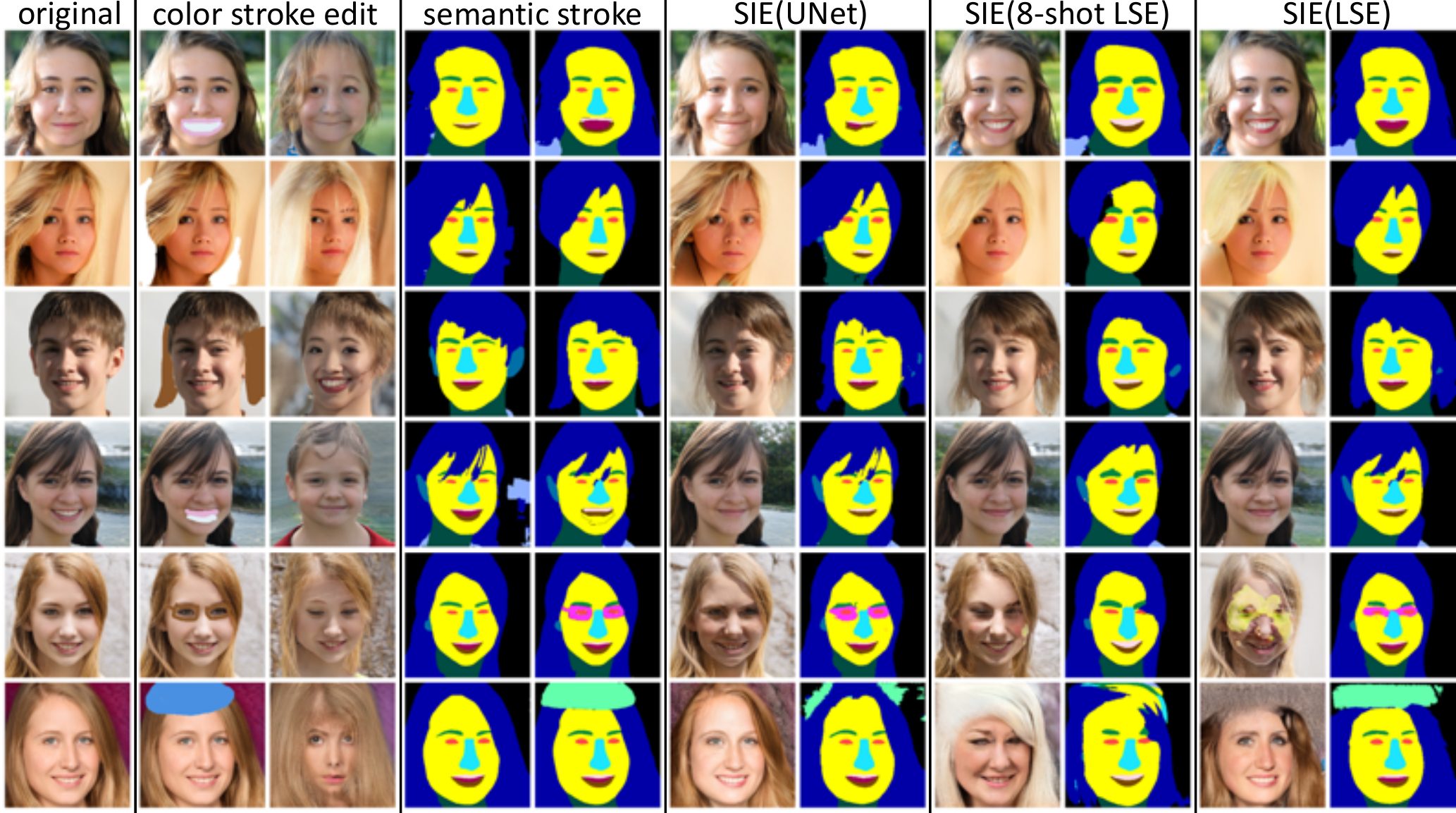}
  \caption{Results of Semantic Image Editing (SIE) on StyleGAN2-FFHQ. Images edited with color strokes are shown in col 2 and 3. Col 4 and 5 show LSE's original segmentation mask and the user-editied semantic masks. The rest of columns show the results of SIE(UNet), SIE(8-shot LSE), and SIE(LSE), respectively.}
  \label{fig:sie}
  \vspace{-3mm}
\end{figure*}

In many cases, the user may want to control a GAN's image generation process.
For example, they might want to adjust the hair color of a generated facial image from blond to red;
and the user may draw a red stroke on the hair to easily specify their intent.
Existing approaches, such as 
color space editing~\cite{Zhu2016Generative,Brock2017NeuralPE,2019Image2StyleGAN,2020Indomain},
aim to find a latent vector that generates an image better matching the user specification.
The latent vector is often found by minimizing a distance measure between
the generated image and the user's strokes in color space.

However, without explicit notion of semantics, the minimization process 
may not respect image semantics, leading to undesired changes of shapes and textures. 
For example, in the 2nd row and 2nd and 3rd columns of \figref{sie},
the user wishes to remove the hair in generated images,
but the color space editing methods tend to just lighten the hair color rather than removing it.


Leveraging LSE, we propose an approached called Semantic
Image Editing (SIE) to enable semantic-aware image generation.
We define a semantic edit loss $L_s = \mathcal{L}(P(G(z)), Y)$, where $\mathcal{L}(\cdot)$ is the
cross-entropy loss, $Y$ is the target semantic mask, and $G$ is the generator that takes
the latent vector $z$ as input.
$P$ is a pretrained segmentation model such as our LSE.  
Starting from an image's latent vector $z$,
we find an output image's latent vector $z'$ by minimizing the loss.
The details are presented in \appref{detail_sie}.

Here, we compare the results of the method using different segmentation models,
including UNet, our 8-shot LSE, and fully trained LSE.
The qualitative results are shown in \figref{sie}.
For each instance, we include both the results of color-space editing and semantic editing,
and more results are presented in~\appref{supp_results}.

First, SIE(UNet) controls image generation and better preserves semantics than color-space editing.  In
comparison to the results of color-space editing, the undesired changes in output
images are greatly reduced, although SIE(UNet) may still fail to transform the image's semantics:
for instance, in the 1st and 2nd row of \figref{sie}, SIE(UNet) barely changes the original image.
We speculate that this is because the gradient from $L_s$ is carried through the entire UNet, 
making the optimization process more difficult. 

SIE(8-shot LSE) edits the semantics better than SIE(UNet): it preserves the semantic regions
not intended by the user. 
However, in 5th and 6th row, SIE(8-shot LSE) produces lower image quality.
We conjecture that this is due to highly unbalanced data distribution: the semantic classes ``hat'' and ``eyeglasses''
occur sparsely in the training dataset.
As a result, those classes can not be well represented in the GAN model\textemdash leading to the well-known 
mode collapse problem.
Lastly, SIE(8-shot LSE) has a similar performance to SIE(LSE), 
although its LSE is trained with much fewer annotations. 

\subsection{Few-shot Conditional Generation}\label{sec:scs}
Semantic-Conditional Sampling (SCS) aims to synthesize an image subject to a semantic mask.
It offers the user more control over the image
generation process. SCS has been explored~\cite{park2019gaugan,2019SEAN},
but most previous works rely on large annotated datasets to train their models.
Thanks to its simplicity, our LSE can be trained with a small set of annotated images (recall our few-shot LSE).
Here we leverage it to build a few-shot SCS model.
It is the need of only a few labeled images that differs our method from existing image-to-image translation methods \cite{Isola2017pix,Zhu2017Unpaired,2019SEAN,park2019gaugan,2017High}.


Our few-shot SCS finds output latent vector by formulating a minimization problem 
similar to SIE discussed in \secref{sie}. 
But unlike SIE, which takes the input image's latent vector,
it needs to choose a proper initial latent vector.
The details of the minimization process are presented in \appref{detail_scs}.

\begin{figure}[t]
  \centering
  \includegraphics[trim=0 275 810 815,clip,width=\linewidth]{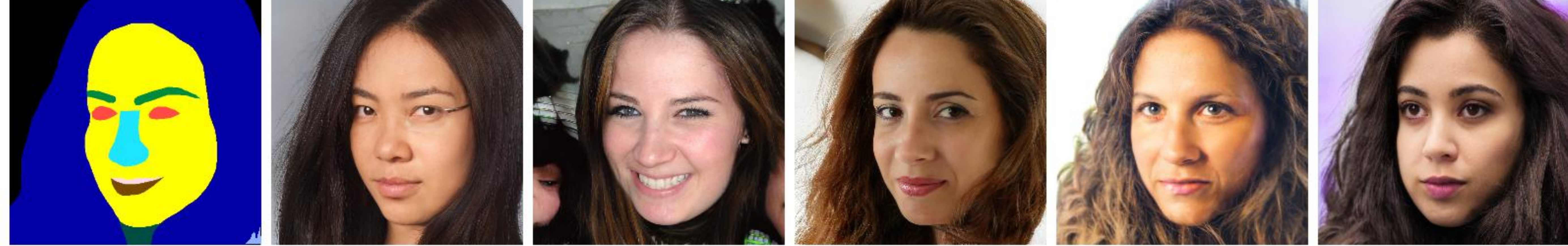} \\
  \includegraphics[trim=0 275 810 815,clip,width=\linewidth]{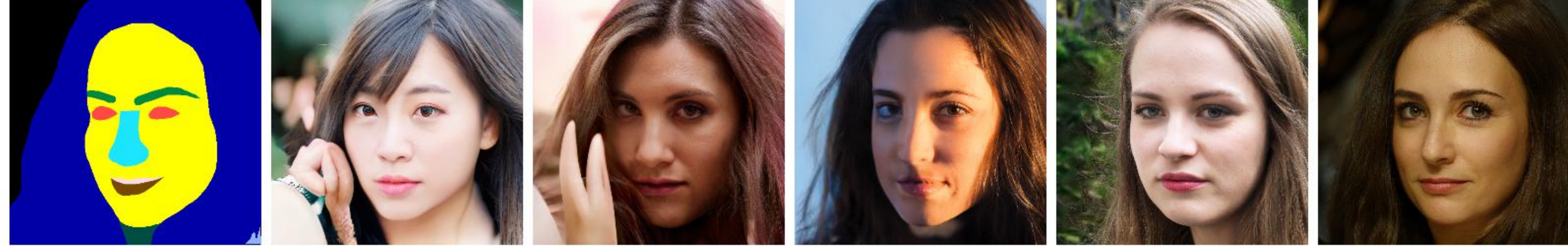} \\
  \includegraphics[trim=0 5 810 1085,clip,width=\linewidth]{figures/stylegan2_ffhq_r0_n8trunc.pdf} \\
  \includegraphics[trim=0 0 810 1085,clip,width=\linewidth]{figures/stylegan2_ffhq_baselinetrunc.pdf} \\

  \includegraphics[trim=0 275 810 810,clip,width=\linewidth]{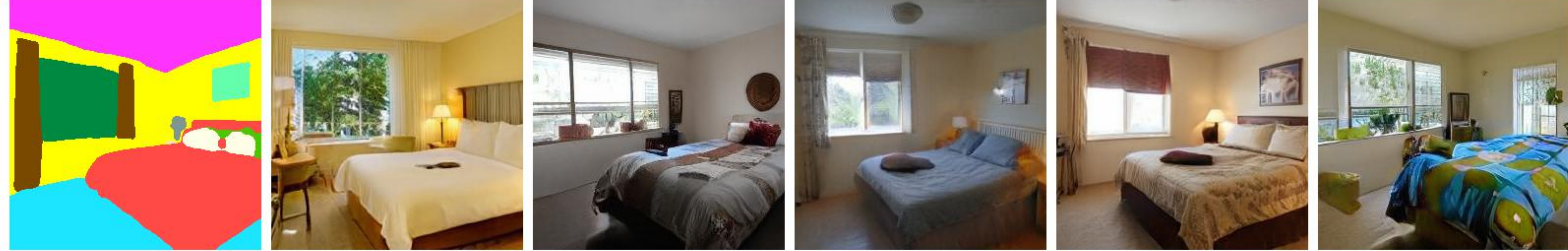} \\
  \includegraphics[trim=0 275 810 815,clip,width=\linewidth]{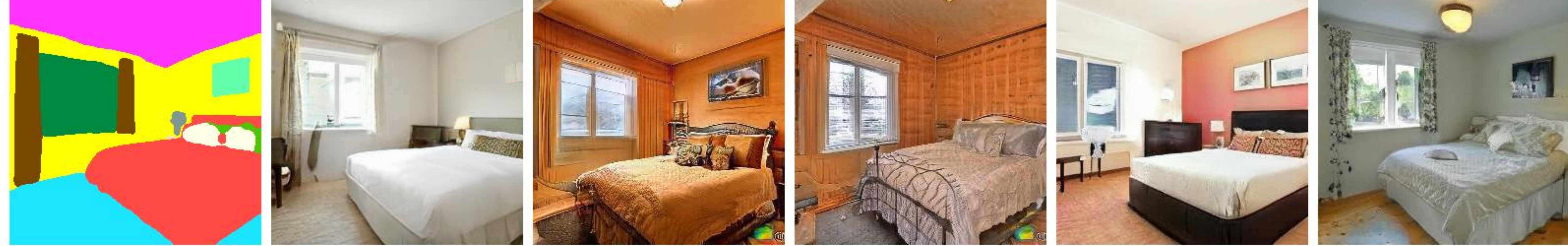} \\
  \includegraphics[trim=0 5 810 1085,clip,width=\linewidth]{figures/stylegan2_bedroom_r0_n8trunc.pdf} \\
  \includegraphics[trim=0 0 810 1085,clip,width=\linewidth]{figures/stylegan2_bedroom_baselinetrunc.pdf} \\

  \includegraphics[trim=0 275 810 810,clip,width=\linewidth]{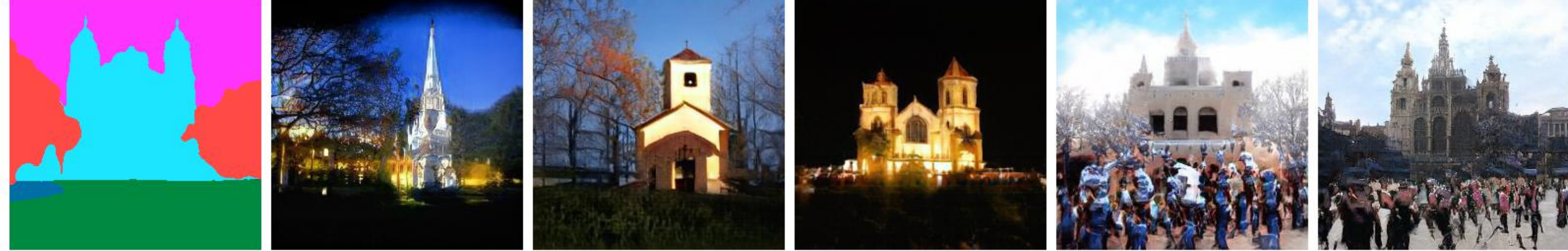} \\
  \includegraphics[trim=0 275 810 815,clip,width=\linewidth]{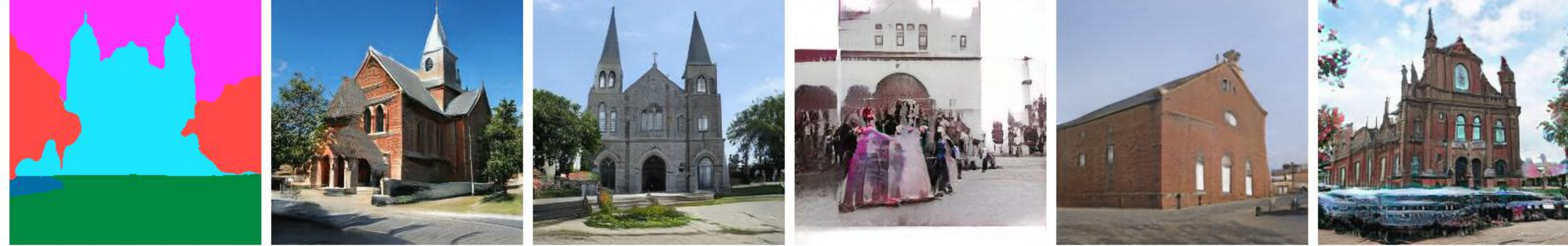} \\
  \includegraphics[trim=0 5 810 1085,clip,width=\linewidth]{figures/stylegan2_church_r0_n8trunc.pdf} \\
  \includegraphics[trim=0 0 810 1085,clip,width=\linewidth]{figures/stylegan2_church_baselinetrunc.pdf} \\
  \caption{SCS results on StyleGAN2. Every pair of rows are to compare SCS(8-shot LSE) results (shown
      in the first row of each pair) with SCS using a pretrained segmentation model (shown in 
  the second row of each pair). }
  \label{fig:scs}
  \vspace{-3mm}
\end{figure}

\paragraph{Qualitative results.}
Figure~\ref{fig:scs} shows conditionally sampled images using 8-shot LSEs.
The eight image labels are produced by a pretained semantic segmentation model:
for facial images, we use UNet;
and for bedroom and church images, we use DeepLabV3.

While the generated images are diverse, they all respect well the provided semantic targets (first column of \figref{scs}).
For facial image generation (1st to 4th row), faces are well matched to the
provided semantic masks. 
For bedroom images (5th to 8th row), the location and
orientation of beds, windows, and walls all match well to the target semantic layout.  
For church images (9th to 12th row), the church geometries are mostly matched.
For instance, in the first group of church images, three of the five samples have two tall towers and 
one short tower in between.  In the second group, all the samples have a tall tower on the left side,
matching the provided semantic mask.

The SCS(8-shot LSE) and the baseline SCS(UNet) have comparable semantic quality in generated images.
We notice that in 3rd and 4th rows, 
SCS(8-shot LSE) appears less successful than SCS(UNet) to respect the provided semantic targets. 
We believe that this is again due to the data imbalance:
``hat'' and ``eyeglasses'' occur must less frequently in the training dataset than other semantic classes. 

\paragraph{Quantitative results.}
We compute the semantic accuracy of SCS, which measures the 
discrepancy between the semantic target and the segmentation of a generated image.
We present the formal definition of the accuracy in \eq{scs_miou} of the appendix,
and report the results in \tabref{quant_scs}.
On the church dataset, the SCS(few-shot LSE) performs slightly worse than SCS(UNet),
while on the bedroom and face datasets, our method with 8-shot (and 16-shot) LSE is even better than
SCS(UNet).

\begin{table}[t]
  \centering
  \begin{tabular}{cccc}
  \whline{1.0pt}
  N & Church & Bedroom & FFHQ \\\hline
  1 & 16.0 $\pm$ 1.4 & 17.5 $\pm$ 2.0 & 37.2 $\pm$ 0.8 \\
  4 & 18.0 $\pm$ 1.3 & 21.6 $\pm$ 0.9 & 39.1 $\pm$ 0.5 \\
  8 & 19.6 $\pm$ 0.5 & 21.7 $\pm$ 0.8 & 39.4 $\pm$ 0.9 \\
  16 & 20.4 $\pm$ 0.6 & 22.3 $\pm$ 0.4 & 40.0 $\pm$ 0.2 \\\hline
  baseline & 23.1 & 17.3 & 34.3 \\
  \whline{1.0pt}
  \end{tabular}
  \vspace{1mm}
  \caption{The semantic accuracy measures the
      semantic agreement between generated images and targets. 
  For SCS with few-shot LSEs, each model is trained for 5 times with different training data to account for the training data variance.
  The numbers before $\pm$ sign are the average results of the 5 repeats, and the numbers following $\pm$ indicate the maximum deviations from the average.
  Experiments are done on StyleGAN2.}
  \label{tab:quant_scs}
  \vspace{-2mm}
\end{table}


\section{Conclusions}\label{sec:conclusion}
In this work, we study how the image semantics are embedded in GAN's feature maps.
We propose a Linear Semantic Extractor (LSE) to extract image semantics modeled by GANs.  Experiments on
various GANs show that LSE can indeed reveal the semantics from feature maps.
We also study the class centers and cosine
similarities between different classes to provide geometric interpretation of our LSE. 
Therefore, it is well-backed that GANs use a linear notion to encode semantics.
Then, we successfully train LSEs in few-shot settings.
Using only 16 training annotations, we obtain 73.5\%, 78.3\%,
and 88.1\% performance relative to fully supervised LSEs on the church,
bedroom, and face images.
Finally, we build two novel applications based on few-shot LSEs:
the few-shot Semantic-Conditional Sampling and the few-shot Semantic
Image Editing.
Our methods can match or surpass the baselines using fully supervised
segmentation networks.
Using the proposed methods, users can exert precise and diverse 
spatial semantic controllability over pretrained GAN models with only a few annotations.

\paragraph{Acknowledgements.} 
This work was supported in part
by the National Science Foundation (1910839, 1717178, 1816041).
Any opinions, findings, and conclusions or recommendations expressed in this material are those
of the authors and do not necessarily reflect the views of the
National Science Foundation or others.

{\small
\bibliographystyle{ieee_fullname}
\bibliography{bib}

\begin{thebibliography}{10}\itemsep=-1pt

\bibitem{Rameen2019edit}
Rameen Abdal, Yipeng Qin, and Peter Wonka.
\newblock Image2stylegan++: How to edit the embedded images?
\newblock {\em CoRR}, abs/1911.11544, 2019.

\bibitem{2019Image2StyleGAN}
Rameen Abdal, Yipeng Qin, and Peter Wonka.
\newblock Image2stylegan++: How to edit the embedded images?
\newblock 2019.

\bibitem{adbal2020styleflow}
Rameen {Abdal}, Peihao {Zhu}, Niloy {Mitra}, and Peter {Wonka}.
\newblock {StyleFlow: Attribute-conditioned Exploration of StyleGAN-Generated
  Images using Conditional Continuous Normalizing Flows}.
\newblock {\em arXiv e-prints}, page arXiv:2008.02401, Aug. 2020.

\bibitem{bau2020rewriting}
David Bau, Steven Liu, Tongzhou Wang, Jun-Yan Zhu, and Antonio Torralba.
\newblock Rewriting a deep generative model.
\newblock In {\em Proceedings of the European Conference on Computer Vision
  (ECCV)}, 2020.

\bibitem{bau2018gan}
David Bau, Jun-Yan Zhu, Hendrik Strobelt, Bolei Zhou, Joshua~B Tenenbaum,
  William~T Freeman, and Antonio Torralba.
\newblock Gan dissection: Visualizing and understanding generative adversarial
  networks.
\newblock {\em arXiv preprint arXiv:1811.10597}, 2018.

\bibitem{brock2018large}
Andrew Brock, Jeff Donahue, and Karen Simonyan.
\newblock Large scale gan training for high fidelity natural image synthesis.
\newblock {\em arXiv preprint arXiv:1809.11096}, 2018.

\bibitem{Brock2017NeuralPE}
A. Brock, T. Lim, J.~M. Ritchie, and N. Weston.
\newblock Neural photo editing with introspective adversarial networks.
\newblock {\em ArXiv}, abs/1609.07093, 2017.

\bibitem{2017Rethinking}
Liang~Chieh Chen, George Papandreou, Florian Schroff, and Hartwig Adam.
\newblock Rethinking atrous convolution for semantic image segmentation.
\newblock 2017.

\bibitem{chen2016infogan}
Xi {Chen}, Yan {Duan}, Rein {Houthooft}, John {Schulman}, Ilya {Sutskever}, and
  Pieter {Abbeel}.
\newblock {InfoGAN: Interpretable Representation Learning by Information
  Maximizing Generative Adversarial Nets}.
\newblock {\em arXiv e-prints}, page arXiv:1606.03657, June 2016.

\bibitem{collins2018}
Edo Collins, Radhakrishna Achanta, and Sabine Susstrunk.
\newblock Deep feature factorization for concept discovery.
\newblock In {\em The European Conference on Computer Vision (ECCV)}, 2018.

\bibitem{Collins20}
Edo Collins, Raja Bala, Bob Price, and Sabine S{\"u}sstrunk.
\newblock Editing in style: Uncovering the local semantics of {GANs}.
\newblock In {\em IEEE Conference on Computer Vision and Pattern Recognition
  (CVPR)}, 2020.

\bibitem{denton2015deep}
Emily~L Denton, Soumith Chintala, Rob Fergus, et~al.
\newblock Deep generative image models using a laplacian pyramid of adversarial
  networks.
\newblock In {\em Advances in neural information processing systems}, pages
  1486--1494, 2015.

\bibitem{dumoulin2016adversarially}
Vincent Dumoulin, Ishmael Belghazi, Ben Poole, Olivier Mastropietro, Alex Lamb,
  Martin Arjovsky, and Aaron Courville.
\newblock Adversarially learned inference.
\newblock {\em arXiv preprint arXiv:1606.00704}, 2016.

\bibitem{ghosh2020gif}
Partha Ghosh, Pravir~Singh Gupta, Roy Uziel, Anurag Ranjan, Michael Black, and
  Timo Bolkart.
\newblock Gif: Generative interpretable faces, 2020.

\bibitem{goodfellow2014generative}
Ian Goodfellow, Jean Pouget-Abadie, Mehdi Mirza, Bing Xu, David Warde-Farley,
  Sherjil Ozair, Aaron Courville, and Yoshua Bengio.
\newblock Generative adversarial nets.
\newblock In {\em Advances in neural information processing systems}, pages
  2672--2680, 2014.

\bibitem{2017Arbitrary}
Xun Huang and Serge Belongie.
\newblock Arbitrary style transfer in real-time with adaptive instance
  normalization.
\newblock 2017.

\bibitem{huang2017stacked}
Xun Huang, Yixuan Li, Omid Poursaeed, John Hopcroft, and Serge Belongie.
\newblock Stacked generative adversarial networks.
\newblock In {\em Proceedings of the IEEE conference on computer vision and
  pattern recognition}, pages 5077--5086, 2017.

\bibitem{Isola2017pix}
P. {Isola}, J. {Zhu}, T. {Zhou}, and A.~A. {Efros}.
\newblock Image-to-image translation with conditional adversarial networks.
\newblock In {\em 2017 IEEE Conference on Computer Vision and Pattern
  Recognition (CVPR)}, pages 5967--5976, 2017.

\bibitem{jahanian2019steerability}
Ali Jahanian, Lucy Chai, and Phillip Isola.
\newblock On the''steerability" of generative adversarial networks.
\newblock {\em arXiv preprint arXiv:1907.07171}, 2019.

\bibitem{karras2017progressive}
Tero Karras, Timo Aila, Samuli Laine, and Jaakko Lehtinen.
\newblock Progressive growing of gans for improved quality, stability, and
  variation.
\newblock {\em arXiv preprint arXiv:1710.10196}, 2017.

\bibitem{karras2019style}
Tero Karras, Samuli Laine, and Timo Aila.
\newblock A style-based generator architecture for generative adversarial
  networks.
\newblock In {\em Proceedings of the IEEE Conference on Computer Vision and
  Pattern Recognition}, pages 4401--4410, 2019.

\bibitem{karras2019analyzing}
Tero Karras, Samuli Laine, Miika Aittala, Janne Hellsten, Jaakko Lehtinen, and
  Timo Aila.
\newblock Analyzing and improving the image quality of stylegan.
\newblock {\em arXiv preprint arXiv:1912.04958}, 2019.

\bibitem{kingma2014adam}
Diederik~P Kingma and Jimmy Ba.
\newblock Adam: A method for stochastic optimization.
\newblock {\em arXiv preprint arXiv:1412.6980}, 2014.

\bibitem{lee2019maskgan}
Cheng-Han Lee, Ziwei Liu, Lingyun Wu, and Ping Luo.
\newblock Maskgan: towards diverse and interactive facial image manipulation.
\newblock {\em arXiv preprint arXiv:1907.11922}, 2019.

\bibitem{liu2015faceattributes}
Ziwei Liu, Ping Luo, Xiaogang Wang, and Xiaoou Tang.
\newblock Deep learning face attributes in the wild.
\newblock In {\em Proceedings of International Conference on Computer Vision
  (ICCV)}, December 2015.

\bibitem{odena2016cgan}
Augustus {Odena}, Christopher {Olah}, and Jonathon {Shlens}.
\newblock {Conditional Image Synthesis With Auxiliary Classifier GANs}.
\newblock {\em arXiv e-prints}, page arXiv:1610.09585, Oct. 2016.

\bibitem{park2019gaugan}
Taesung Park, Ming-Yu Liu, Ting-Chun Wang, and Jun-Yan Zhu.
\newblock Gaugan: semantic image synthesis with spatially adaptive
  normalization.
\newblock In {\em ACM SIGGRAPH 2019 Real-Time Live!}, pages 1--1. 2019.

\bibitem{radford2015unsupervised}
Alec Radford, Luke Metz, and Soumith Chintala.
\newblock Unsupervised representation learning with deep convolutional
  generative adversarial networks.
\newblock {\em arXiv preprint arXiv:1511.06434}, 2015.

\bibitem{ronneberger2015u}
Olaf Ronneberger, Philipp Fischer, and Thomas Brox.
\newblock U-net: Convolutional networks for biomedical image segmentation.
\newblock In {\em International Conference on Medical image computing and
  computer-assisted intervention}, pages 234--241. Springer, 2015.

\bibitem{shen2019interpreting}
Yujun Shen, Jinjin Gu, Xiaoou Tang, and Bolei Zhou.
\newblock Interpreting the latent space of gans for semantic face editing.
\newblock {\em arXiv preprint arXiv:1907.10786}, 2019.

\bibitem{2020Closed}
Yujun Shen and Bolei Zhou.
\newblock Closed-form factorization of latent semantics in gans.
\newblock 2020.

\bibitem{Suzuki2018collaging}
Ryohei {Suzuki}, Masanori {Koyama}, Takeru {Miyato}, Taizan {Yonetsuji}, and
  Huachun {Zhu}.
\newblock {Spatially Controllable Image Synthesis with Internal Representation
  Collaging}.
\newblock {\em arXiv e-prints}, page arXiv:1811.10153, Nov. 2018.

\bibitem{tewari2020stylerig}
Ayush Tewari, Mohamed Elgharib, Gaurav Bharaj, Florian Bernard, Hans-Peter
  Seidel, Patrick Pérez, Michael Zollhöfer, and Christian Theobalt.
\newblock Stylerig: Rigging stylegan for 3d control over portrait images, 2020.

\bibitem{voynov2020unsupervised}
Andrey Voynov and Artem Babenko.
\newblock Unsupervised discovery of interpretable directions in the gan latent
  space.
\newblock {\em arXiv preprint arXiv:2002.03754}, 2020.

\bibitem{2017High}
Ting~Chun Wang, Ming~Yu Liu, Jun~Yan Zhu, Andrew Tao, Jan Kautz, and Bryan
  Catanzaro.
\newblock High-resolution image synthesis and semantic manipulation with
  conditional gans.
\newblock 2017.

\bibitem{yang2019semantic}
Ceyuan Yang, Yujun Shen, and Bolei Zhou.
\newblock Semantic hierarchy emerges in deep generative representations for
  scene synthesis.
\newblock {\em arXiv preprint arXiv:1911.09267}, 2019.

\bibitem{yu2015lsun}
Fisher Yu, Ari Seff, Yinda Zhang, Shuran Song, Thomas Funkhouser, and Jianxiong
  Xiao.
\newblock Lsun: Construction of a large-scale image dataset using deep learning
  with humans in the loop.
\newblock {\em arXiv preprint arXiv:1506.03365}, 2015.

\bibitem{zhang2018self}
Han Zhang, Ian Goodfellow, Dimitris Metaxas, and Augustus Odena.
\newblock Self-attention generative adversarial networks.
\newblock {\em arXiv preprint arXiv:1805.08318}, 2018.

\bibitem{2020ResNeSt}
Hang Zhang, Chongruo Wu, Zhongyue Zhang, Yi Zhu, Zhi Zhang, Haibin Lin, Yue
  Sun, Tong He, Jonas Mueller, and R~and Manmatha.
\newblock Resnest: Split-attention networks.
\newblock 2020.

\bibitem{zhang2017stackgan}
Han Zhang, Tao Xu, Hongsheng Li, Shaoting Zhang, Xiaogang Wang, Xiaolei Huang,
  and Dimitris~N Metaxas.
\newblock Stackgan: Text to photo-realistic image synthesis with stacked
  generative adversarial networks.
\newblock In {\em Proceedings of the IEEE international conference on computer
  vision}, pages 5907--5915, 2017.

\bibitem{zhang2020image}
Yuxuan Zhang, Wenzheng Chen, Huan Ling, Jun Gao, Yinan Zhang, Antonio Torralba,
  and Sanja Fidler.
\newblock Image gans meet differentiable rendering for inverse graphics and
  interpretable 3d neural rendering, 2020.

\bibitem{2017Scene}
Bolei Zhou, Hang Zhao, Francesco Xavier~Puig Fernandez, Sanja Fidler, and
  Antonio Torralba.
\newblock Scene parsing through ade20k dataset.
\newblock In {\em 2017 IEEE Conference on Computer Vision and Pattern
  Recognition (CVPR)}, 2017.

\bibitem{2020Indomain}
Jiapeng Zhu, Yujun Shen, Deli Zhao, and Bolei Zhou.
\newblock In-domain gan inversion for real image editing.
\newblock 2020.

\bibitem{Zhu2016Generative}
Jun~Yan Zhu, Philipp Krhenbühl, Eli Shechtman, and Alexei~A. Efros.
\newblock Generative visual manipulation on the natural image manifold.
\newblock 2016.

\bibitem{Zhu2017Unpaired}
Jun~Yan Zhu, Taesung Park, Phillip Isola, and Alexei~A. Efros.
\newblock Unpaired image-to-image translation using cycle-consistent
  adversarial networks.
\newblock 2017.

\bibitem{2019SEAN}
Peihao Zhu, Rameen Abdal, Yipeng Qin, and Peter Wonka.
\newblock Sean: Image synthesis with semantic region-adaptive normalization.
\newblock 2019.

\end{thebibliography}
}

\clearpage
\setcounter{page}{1}

\begin{strip}
    \vspace{-6mm}
\begin{center}
 \Large
 \textbf{Supplementary Document}\\
 \smallskip
 \smallskip
 \textbf{Linear Semantics in Generative Adversarial Networks}
 \smallskip
\end{center}
\end{strip}

\appendix
\section{Definition of IoU.} \label{sec:iou}

Intersection-over-Union (IoU) is a widely used metric in semantic segmentation literature.
A segmentation of a category is represented as a set of pixels among all pixels that belong to this category.
Suppose we have a segmentation $A$ and $B$, their IoU is $\text{IoU}(A, B) = \frac{|A \cap B|}{|A \cup B|}$.
Taking the average across a set of segmentations $\mathcal{A}=\{A_i\}$ and $\mathcal{B}=\{B_i\}$, we get the average IoU on this dataset:

\begin{equation}
    \displaystyle
    IoU(\mathcal{A}, \mathcal{B})=\frac{1}{N} \sum_{A_i \cup B_i \neq \emptyset} IoU(A_i, B_i)
\end{equation}

IoU evaluates how well the segmentation is for a particular category. Mean IoU (mIoU) evaluates the overall performance of multi-class segmentation.
It is calculated as the mean of IoUs over all categories.

\section{Proof of commutative property}\label{sec:comm}

Now we are going to prove that $\mathsf{u}^\uparrow(\T_i\cdot\xx_i)=\T_i\cdot\mathsf{u}^\uparrow(\xx_i)$.

Suppose that a pixel $p$ that we want to interpolate lies in the rectangle of four pixels $(x_{11}, x_{12}, x_{21}, x_{22})$ and its relative position is described by $(\alpha, \beta)$ as distance ratio to the edges of the rectangle.
The interpolated value is

\begin{equation}
\begin{split}
    & \mathsf{u}_p^\uparrow(x_{11}, x_{12}, x_{21}, x_{22}, \alpha, \beta) \\
    = & (1 - \beta)[(1 - \alpha) x_{11} + \alpha x_{12}] \\
    & + \beta [(1 - \alpha)x_{21} + \alpha x_{22}]
\end{split}
\end{equation}

When we do convolution then upsample, we get the following result

\begin{equation}
\begin{split}
& \mathsf{u}_p^\uparrow(\T_i x_{11}, \T_i x_{12}, \T_i x_{21}, \T_i x_{22}, \alpha, \beta) \\
= & (1 - \beta)[(1 - \alpha) \T_i x_{11} + \alpha \T_i x_{12}] \\
& + \beta [(1 - \alpha) \T_i x_{21} + \alpha \T_i x_{22}] \\
= & \T_i (1 - \beta) [(1 - \alpha) x_{11} + \alpha x_{12}] \\
& + \T_i \beta [(1 - \alpha)x_{21} + \alpha x_{22}] \\
= & \T_i \mathsf{u}_p^\uparrow(x_{11}, x_{12}, x_{21}, x_{22}, \alpha, \beta)
\end{split}
\end{equation}

which is exactly upsampling then convoluting.

\section{Nonlinear semantic extractors}\label{sec:nonlinear}

\begin{figure}
    \centering
    \subfloat[NSE-1]{
        \includegraphics[width=0.44\linewidth]{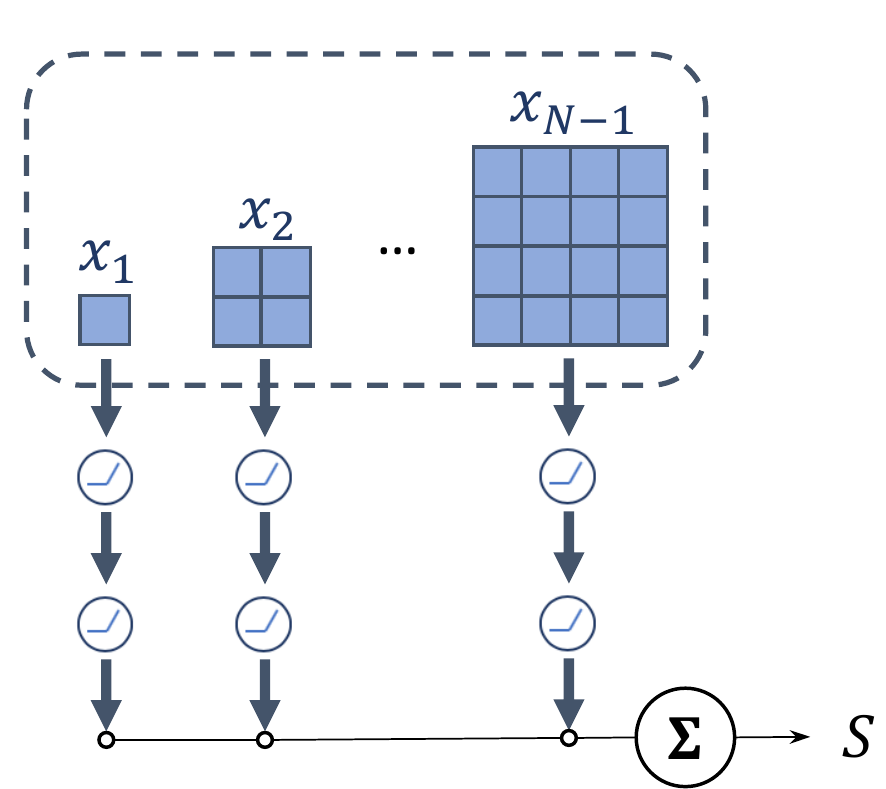}}
    \subfloat[NSE-2]{
        \centering
        \includegraphics[width=0.53\linewidth]{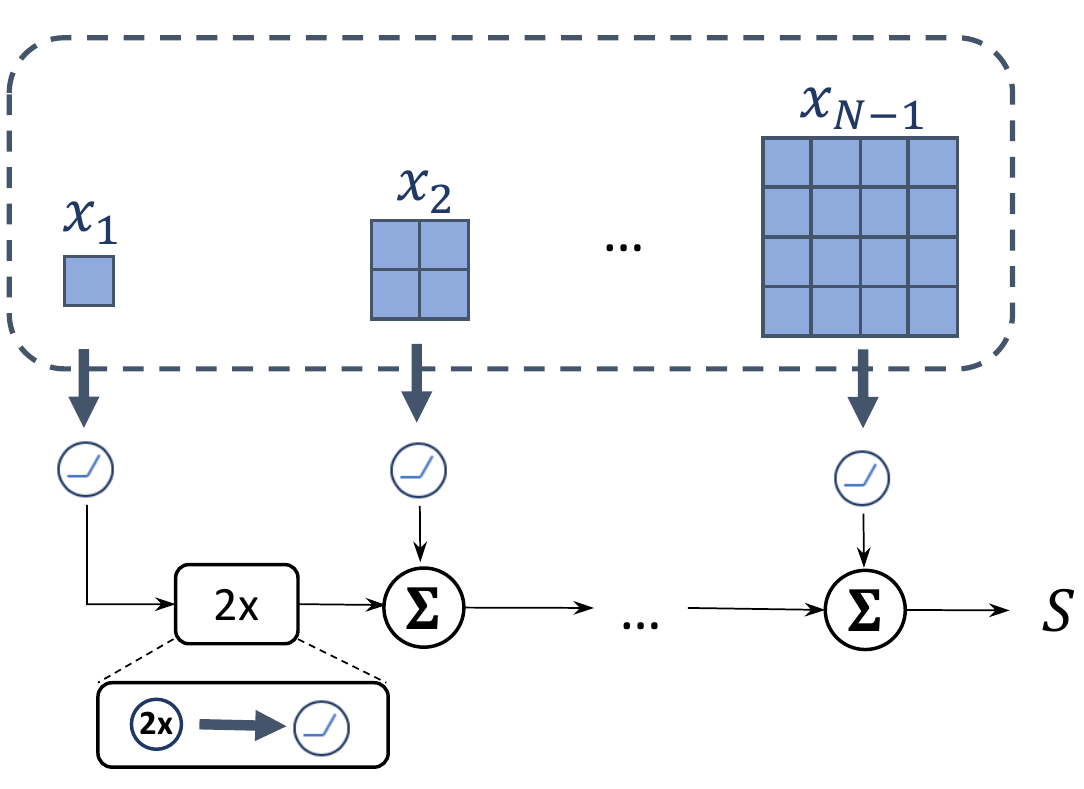}}
    \caption{The architecture of NSEs. The thick blue arrow refers to $3\times3$ convolution with stride 1. ``2x'' refers to nearest upsampling with factor 2.}
    \label{fig:nse_arch}
\end{figure}

The architectures of NSEs are shown in ~\figref{nse_arch}.
NSE-1 is a direction generalization from LSE. Instead of extracting semantics from each layer linearly, NSE-1 extract semantics with 3 nonlinear convolution layers from each layer.
The results from each layer are upsampled and summed up the same as LSE.
The architecture of NSE-2 is inspired by DCGAN, where the resolution gradually increases.
The output from the last layer of NSE-2 is upsampled and summed with embedding extracted from GAN's feature maps.

\section{Experiment details}\label{sec:setup}

\begin{figure*}[h]
    \centering
    \includegraphics[width=0.9\linewidth]{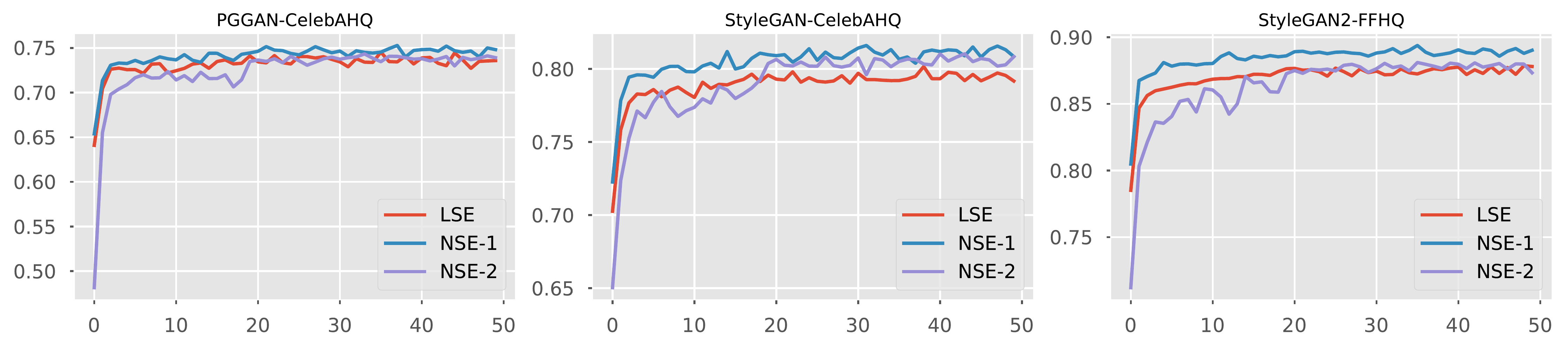} \\
    \includegraphics[width=0.9\linewidth]{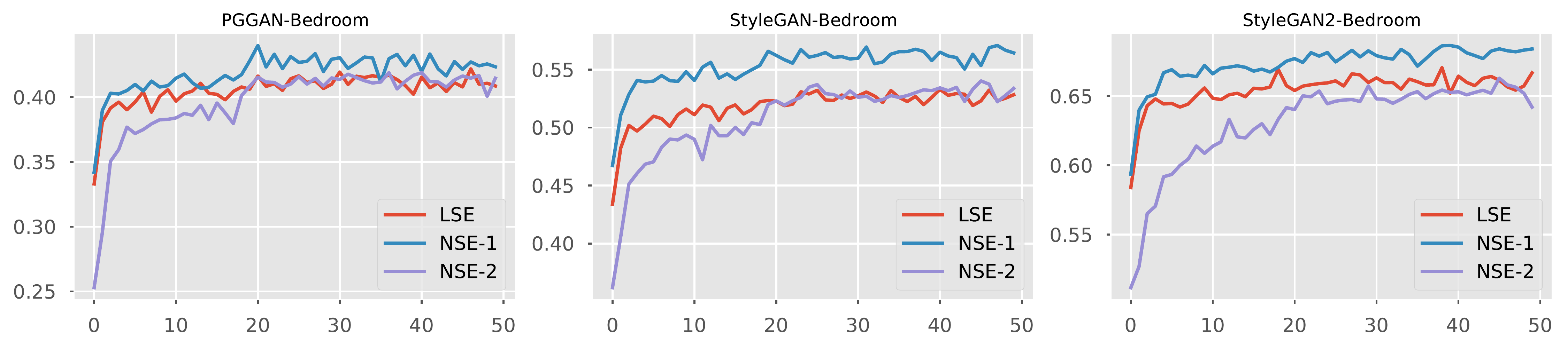} \\
    \includegraphics[width=0.9\linewidth]{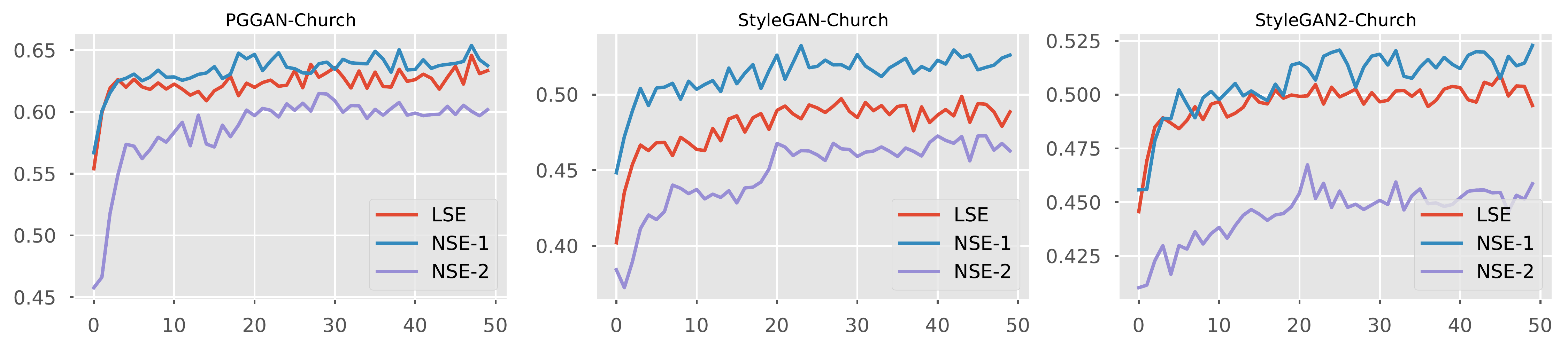}
    \caption{Training evolution of mIoU of all the semantic extractors on GANs.}
    \label{fig:train_eval}
\end{figure*}

\paragraph{Pretrained networks}\label{sec:setup_pretrain}

For segmentation on facial images, we train a UNet~\cite{ronneberger2015u} on CelebAMask-HQ~\cite{lee2019maskgan} dataset to perform semantic segmentation.
The training script is adapted from the project repo \footnote{https://github.com/switchablenorms/CelebAMask-HQ} of MaskGAN~\cite{lee2019maskgan}.
Our UNet follows standard UNet architecture and takes $512 \times 512$ images as input.
It is trained using Adam optimizer~\cite{kingma2014adam} for 40 epochs (about 76k iterations), with learning rate $3 \times 10^{-4}$, $\beta_1 = 0.9$, $\beta_2 = 0.999$ and batch size 16.

The CelebAMask-HQ dataset contains {30k} human-labeled face-segmentation pairs.
The face images are aligned to the center and have $1024 \times 1024$ resolution.
The semantic labeling's resolution is $512 \times 512$, consisting of 19 semantic categories.
However, there are duplicate semantic concepts like ``right ear'' and ``left ear'', ``right eye'' and ``left eye'', ``right brow'' and ``left brow''.
In those pairs, as both categories differ only in spatial location, we unify them into ``ear'', ``eye'', and ``brow''.
Besides, only 50 instances are labeled with ``necklace'', thus we remove it by merging ``necklace'' into ``neck''.
As a result, we get 15 semantic categories listed in category results \tabref{cat_face}.

For segmentation on GANs trained on LSUN's bedroom and church datasets, we use the DeepLabV3~\cite{2017Rethinking} with ResNeSt~\cite{2020ResNeSt} backbone trained on ADE20k dataset~\cite{2017Scene}. Model parameters are obtained from here\footnote{https://github.com/zhanghang1989/PyTorch-Encoding}.
However, the DeepLabV3 predicts in total 150 categories, where most are not present in generated images, because GANs are train on LSUN datasets rather than the ADE20k dataset.
We apply a category selection process (detailed in \appref{category_selection}) to remove irrelevant semantic categories.

All the pretrained GAN models are adapted from GenForce\footnote{https://github.com/genforce/genforce}.
The image resolution of GANs trained on face datasets are $1024\times1024$, and the rest are $256 \times 256$.

\paragraph{Training}\label{sec:setup_train}

\begin{table*}[h]
    \centering
    \begin{tabular}{c|cc|cc|cc}
    \whline{1.0pt}
    Generator & \multicolumn{2}{c|}{PGGAN} & \multicolumn{2}{c|}{StyleGAN} & \multicolumn{2}{c}{StyleGAN2} \\\hline
    Dataset & Bedroom & Church & Bedroom & Church & Bedroom & Church \\
    LSE & 32.4 (-4.8) & 49.4 (-2.4) & 39.8 (-8.0) & 34.8 (-7.0) & 54.3 (-4.2) & 36.8 (-3.7) \\
    NSE-1 & \textbf{34.1} & \textbf{50.6} & \textbf{43.2} & \textbf{37.4} & \textbf{56.6} & \textbf{38.2} \\
    NSE-2 & 28.9 (-15.1) & 45.1 (-10.8) & 39.5 (-8.7) & 33.3 (-11.0) & 51.9 (-8.4) & 34.5 (-9.8) \\\hline
    \multicolumn{7}{c}{Results adaped from \tabref{quant_se}.} \\ \hline
    LSE & 33.2 (-3.2) & 51.3 (-3.2) & 39.9 (-7.8) & 35.4 (-6.3) & 53.9 (-3.4) & 37.7 (-2.6) \\
    NSE-1 & \textbf{34.3} & \textbf{53.0} & \textbf{43.3} & \textbf{37.8} & \textbf{55.8} & \textbf{38.7} \\
    NSE-2 & 30.7 (-10.5) & 49.5 (-6.6) & 38.9 (-10.2) & 34.0 (-10.1) & 52.1 (-6.8) & 35.3 (-8.8) \\
    \whline{1.0pt}
    \end{tabular}
    \caption{The evaluation of LSE, NSE-1, and NSE-2 trained on the full list of ADE20K 150 classes.
    The mIoU(\%) is calculated on the final selected categories, which is the same as the the categories used in the paper.
    The results of models re-trained on the selected categories are also shown in the last three rows for reference.}
    \label{tab:catselect_global}
\end{table*}

For fully supervised training, we sample 51,200 images from the GAN and record their feature maps.
These images are then semantically segmented with an off-the-shelf segmenter in the corresponding data domain.
The semantic masks and feature maps are then used to train the transformation matrix $\T_i$ for every GAN layer.
To be specific, the total matrix $\T$ (defined in \eq{geo_interp}) for StyleGAN2-FFHQ are of size $15 \times 5568$.
For StyleGAN2-Bedroom, $\T$ is shaped as $16 \times 5376$.

$\T_i$ are optimized with Adam \cite{kingma2014adam} with $\beta_1 = 0.9$, $\beta_2 = 0.999$ and initial learning rate $10^{-3}$.
The training takes 50 epochs in total, where each epoch consists of 1,024 samples.
The learning rate is reduced with a factor of 10 at epoch 20.
For the first two epochs, the batch size is 1.
For the next 16 epochs (3 to 19), the batch size is set to 4. For epoch 20 to 50, the batch size is 64.
The total optimization iterations are $1024 \times 2 + \frac{1024}{4} \times 16 + \frac{1024}{64} \times 32 = 6,656$. LSE, NSE-1, and NSE-2 are trained in the same settings.

We record the mIoU of training samples, and show the evolution of training mIoU in \figref{train_eval}.
All the semantic extractors converge during the training.

For the few-shot training of LSEs, we also sample the latent space and segment the images.
The difference is that only a few annotations are made available to the LSE.
We experimented with 1, 4, 8, 16 samples.
The resultant models are named as the one-shot, 4-shot, 8-shot and 16-shot LSEs, respectively.
For the one-shot LSE, the training takes 2000 iterations with batch size 1.
For 4, 8, and 16 samples, the training uses batch sizes 4, 8, and 16 and iteration numbers 2000, 1000, and 500, respectively.
For PGGAN, each batch is exactly the same.
For StyleGAN and StyleGAN2, the layer noises are re-sampled in each batch.
The optimizer setting is the same as in full supervision.

\paragraph{Evaluation}\label{sec:setup_eval}

Conventionally, semantic segmentation methods are evaluated on real image-segmentation datasets.
However, our semantic extractors cannot take real images as input.
One may invert real images in GAN's representation, but the inversion is another challenging problem, thus we do not consider this approach.
As a result, the evaluation cannot be conducted on the common annotated dataset.
Ideally, we should annotate synthetic images manually, but the cost would then be prohibitive.
Therefore, we choose to use the prediction from the off-the-shelf segmenter as the ground-truth for evaluation.

We sample and segment another 10,000 images different from those used in training.
Every time GAN generates an image, we apply the semantic extractor to the generator's feature maps to predict a semantic mask.
The segmentation is compared with the pretrained segmenter's prediction to compute the IoU.

As some datasets (e.g., LSUN's bedroom dataset) may be more difficult to segment than some others (e.g., the CelebAHQ dataset),
we compute relative performance differences between semantic extractors.
Concretely, for each GAN model, there are three semantic extractors to be evaluated, which are LSE, NSE-1, and NSE-2.
Denoting their mIoUs with the pretrained segmenter as $y_i$, and the highest mIoU among the three as $y^*$,
the relative performance difference of each semantic extractor is defined as $\frac{y_i - y^*}{y^*}$.

\section{Category selection}\label{sec:category_selection}

For GANs trained on bedroom and church images, we rely on DeepLabV3 trained on ADE20K to provide the training supervision.
In this section, we aim to remove categories that are not generated by GANs.

First of all, we train and evaluate semantic extractors on the full set of 150 classes.
Then, we remove all the categories that are predicted with mIoU $<10\%$ by all the semantic extractors.
In other words, a category will be selected as long as any of the LSE, NSE-1, and NSE-2 predicts it with mIoU $>10\%$.
In this way, the list of selected categories for each GAN model are decided.
The formal results are obtained by training and evaluating on the selected categories under the same settings.

The evaluation of LSE, NSE-1, and NSE-2 trained on full categories is shown in \tabref{catselect_global}.
Generally, the re-trained semantic extractors obtain slightly better performance, which is expected.
We also show the IoU for each category in \tabref{cat_bedroom_church}, where the table headers also list the final selected categories for each GAN model.

\begin{table*}[t]
    \centering
    \subfloat[StyleGAN2-Bedroom]{
    \resizebox*{\textwidth}{!}{
    \scriptsize
    \begin{tabular}{c|cccccccccccccccc}
    \whline{1.0pt}
    & wall & floor & ceiling & bed & win. & table & curtain & painting & lamp & cushion & pillow & flower & light & chdr. & fan & clock \\\hline
    LSE & 91.29 & 85.56 & 88.47 & 90.53 & 75.58 & 67.19 & 43.13 & 68.79 & 59.43 & 32.29 & 46.05 & 12.98 & 36.73 & 18.06 & 37.79 & 14.77 \\
    NSE-1 & 92.13 & 87.06 & 89.79 & 91.99 & 76.40 & 69.98 & 46.14 & 73.71 & 62.70 & 34.21 & 48.10 & 15.76 & 40.04 & 20.34 & 45.62 & 12.39 \\
    NSE-2 & 91.56 & 86.22 & 88.78 & 91.15 & 72.94 & 67.52 & 42.77 & 69.43 & 58.61 & 30.87 & 46.20 & 4.76 & 26.75 & 12.51 & 35.14 & 5.39 \\\hline
    LSE & 91.29 & 85.95 & 88.27 & 90.94 & 76.23 & 67.31 & 42.08 & 69.29 & 59.28 & 31.17 & 45.57 & 11.77 & 36.17 & 18.07 & 35.72 & 13.62 \\
    NSE-1 & 92.21 & 87.12 & 89.05 & 91.80 & 76.37 & 69.62 & 45.74 & 71.69 & 62.34 & 33.22 & 47.98 & 12.40 & 39.10 & 18.57 & 43.28 & 12.74 \\
    NSE-2 & 91.66 & 86.31 & 88.65 & 91.21 & 74.79 & 68.85 & 43.66 & 70.70 & 60.51 & 26.36 & 45.34 & 4.01 & 31.83 & 11.22 & 30.61 & 7.13 \\
    \whline{1.0pt}
    \end{tabular}}}
    
    \subfloat[StyleGAN-Bedroom]{
    \resizebox*{\textwidth}{!}{
    \scriptsize
    \begin{tabular}{c|ccccccccccccccccccc}
        \whline{1.0pt}
        & wall & floor & ceiling & bed & win. & table & curtain & chair & painting & rug & wdrb. & lamp & cushion & chest & pillow & flower & light & chdr. & fan \\\hline
        LSE & 82.30 & 74.12 & 73.79 & 88.28 & 55.34 & 47.58 & 37.13 & 9.10 & 64.01 & 8.93 & 11.07 & 42.91 & 33.42 & 19.09 & 46.51 & 9.02 & 11.45 & 22.70 & 18.55 \\
        NSE-1 & 83.78 & 75.50 & 76.55 & 89.40 & 57.24 & 50.63 & 40.96 & 10.52 & 67.41 & 11.54 & 12.31 & 48.90 & 36.62 & 18.66 & 49.07 & 10.67 & 25.08 & 26.49 & 29.66 \\
        NSE-2 & 83.05 & 74.61 & 75.44 & 89.34 & 53.99 & 47.98 & 39.63 & 6.03 & 64.04 & 5.11 & 8.78 & 45.89 & 35.95 & 17.45 & 48.88 & 3.02 & 12.32 & 16.52 & 21.49 \\\hline
        LSE & 83.12 & 74.70 & 73.90 & 88.82 & 56.80 & 45.78 & 37.19 & 10.24 & 62.87 & 9.35 & 10.81 & 42.43 & 33.26 & 20.91 & 46.19 & 7.17 & 11.01 & 25.37 & 18.18 \\
        NSE-1 & 84.37 & 75.01 & 76.30 & 89.90 & 58.27 & 49.78 & 39.98 & 11.95 & 66.87 & 12.77 & 12.04 & 49.36 & 35.88 & 22.91 & 48.95 & 10.32 & 24.10 & 25.93 & 27.83 \\
        NSE-2 & 83.81 & 74.19 & 75.38 & 89.64 & 55.27 & 47.43 & 40.03 & 11.23 & 63.60 & 6.76 & 9.27 & 46.61 & 34.22 & 15.45 & 47.84 & 0.66 & 8.92 & 14.58 & 13.85 \\
        \whline{1.0pt}
    \end{tabular}}}

    \subfloat[PGGAN-Bedroom]{
    \resizebox*{\textwidth}{!}{
    \scriptsize
    \begin{tabular}{c|ccccccccccccc}
        \whline{1.0pt}
        & wall & floor & ceiling & bed & windowpane & table & curtain & painting & lamp & pillow & light & chandelier & fan \\\hline
        LSE & 69.46 & 45.07 & 54.05 & 68.39 & 36.38 & 12.58 & 25.77 & 32.67 & 17.18 & 16.10 & 13.65 & 12.14 & 18.06 \\
        NSE-1 & 70.75 & 46.78 & 57.01 & 70.35 & 38.84 & 14.78 & 29.07 & 35.66 & 18.35 & 18.86 & 15.42 & 9.49 & 17.36 \\
        NSE-2 & 68.60 & 45.19 & 54.56 & 68.12 & 33.03 & 12.06 & 27.67 & 34.59 & 15.31 & 16.79 & 0.11 & 0.00 & 0.00 \\\hline
        LSE & 71.91 & 47.88 & 54.53 & 70.29 & 37.06 & 11.71 & 25.39 & 33.74 & 16.82 & 17.90 & 15.57 & 10.83 & 17.94 \\
        NSE-1 & 72.91 & 49.01 & 56.42 & 71.69 & 38.90 & 14.35 & 28.34 & 35.39 & 18.21 & 19.31 & 13.58 & 10.45 & 17.08 \\
        NSE-2 & 72.11 & 47.90 & 54.72 & 71.41 & 38.46 & 13.14 & 27.77 & 35.41 & 16.83 & 18.65 & 0.00 & 1.22 & 1.30 \\
        \whline{1.0pt}
    \end{tabular}}}

    \subfloat[StyleGAN2-Church]{
    \resizebox*{0.9\textwidth}{!}{
    \small
    \begin{tabular}{c|ccccccccccc}
    \whline{1.0pt}
    & building & sky & tree & road & grass & sidewalk & person & earth & plant & car & stairs \\\hline
    LSE & 85.94 & 97.52 & 76.51 & 24.19 & 40.16 & 16.71 & 15.78 & 13.72 & 8.92 & 12.22 & 13.43 \\
    NSE-1 & 86.93 & 97.87 & 78.59 & 25.76 & 44.45 & 17.73 & 17.03 & 14.04 & 10.62 & 13.30 & 14.39 \\
    NSE-2 & 86.65 & 97.71 & 77.96 & 22.16 & 37.87 & 10.78 & 12.92 & 8.87 & 7.52 & 8.21 & 8.91 \\
    \hline
    LSE & 87.96 & 97.46 & 76.31 & 27.32 & 41.61 & 17.08 & 17.62 & 14.21 & 8.28 & 12.84 & 14.23 \\
    NSE-1 & 88.75 & 97.70 & 78.16 & 27.14 & 44.96 & 18.82 & 16.76 & 15.92 & 9.99 & 12.68 & 15.26 \\
    NSE-2 & 88.77 & 97.69 & 77.66 & 22.33 & 39.78 & 13.38 & 12.52 & 10.80 & 6.31 & 8.58 & 10.81 \\
    \whline{1.0pt}
    \end{tabular}}}

    \subfloat[StyleGAN-Church]{
    \resizebox*{0.9\textwidth}{!}{
    \small
    \begin{tabular}{c|cccccccccc}
    \whline{1.0pt}
    & building & sky & tree & road & grass & sidewalk & person & plant & signboard & path \\\hline
    LSE & 88.18 & 95.53 & 49.14 & 23.29 & 39.34 & 11.07 & 9.42 & 9.32 & 14.11 & 8.52 \\
    NSE-1 & 88.55 & 95.69 & 54.25 & 25.06 & 42.24 & 11.05 & 11.61 & 12.55 & 22.53 & 10.40 \\
    NSE-2 & 87.74 & 95.34 & 48.18 & 19.77 & 34.36 & 7.79 & 10.11 & 8.51 & 14.37 & 6.63 \\\hline
    LSE & 91.30 & 95.53 & 47.46 & 25.30 & 41.63 & 13.08 & 8.39 & 9.17 & 13.80 & 8.84 \\
    NSE-1 & 92.01 & 96.01 & 53.01 & 28.06 & 44.84 & 13.33 & 11.56 & 12.25 & 16.62 & 10.63 \\
    NSE-2 & 91.58 & 95.66 & 49.60 & 22.96 & 35.72 & 7.95 & 9.47 & 9.12 & 12.69 & 5.40 \\
    \whline{1.0pt}
    \end{tabular}}}

    \subfloat[PGGAN-Church]{
    \resizebox*{0.6\textwidth}{!}{
    \small
    \begin{tabular}{c|cccccc}
    \whline{1.0pt}
    & building & sky & tree & road & grass & signboard \\\hline
    LSE & 83.98 & 91.21 & 45.55 & 17.01 & 30.14 & 28.51 \\
    NSE-1 & 84.60 & 91.46 & 47.92 & 18.79 & 31.17 & 29.71 \\
    NSE-2 & 83.79 & 90.91 & 42.99 & 15.19 & 23.64 & 14.35 \\\hline
    LSE & 88.17 & 91.33 & 44.57 & 29.10 & 34.05 & 20.78 \\
    NSE-1 & 88.98 & 92.02 & 47.18 & 31.43 & 36.08 & 22.37 \\
    NSE-2 & 88.35 & 91.75 & 44.85 & 25.91 & 30.31 & 16.00 \\
    \whline{1.0pt}
    \end{tabular}}}

    \caption{The IoU (\%) of each category for LSE, NSE-1, and NSE-2.
    In every subtable, the first three rows show the results of models trained with full classes during the category selection process.
    The last three rows of each subtable show the results of the models used in \tabref{quant_se}, which are obtained by re-training on the selected classes. 
    The abbreviation ``win.'', ``wdrb.'', ``chdr.'' stands for ``windowpane'', ``wardrobe'', and ``chandelier'', respectively.}
    \label{tab:cat_bedroom_church}
\end{table*}

\section{Cosine similarity of categories}\label{sec:cosine_similarity}

\begin{algorithm}[b]
    \KwIn{$G$; $P$; $T_1$; $T_2$}
    \KwOut{$\{f_k\}$}
    \For{$k=1,2,\dots,M$}{
        $f_k = \emptyset$ \\
    }
    \While{$\exists k, |f_k| < T_2$}{
        $z \sim \mathcal{N}(0, I)$ \\
        $I, F = G(z)$ // $F$ denotes features \\
        $S = P(I)$ \\
        \For{$k=1,2,\dots,M$}{
            \If{$|f_k| < T_2 \text{ and } |\{p|S_p=k\}| \ge T_1$}{
                $R = choice(\{p|S_p=k\}, T_1)$ \\
                $f_k = f_k \cup \{F_p | p \in R\}$ \\
            }
        }
    }
    \caption{Fair feature sampling algorithm.}
    \label{alg:fair_sample}
\end{algorithm}

\begin{figure}
    \subfloat[StyleGAN-CelebAHQ]{\includegraphics[width=0.48\textwidth]{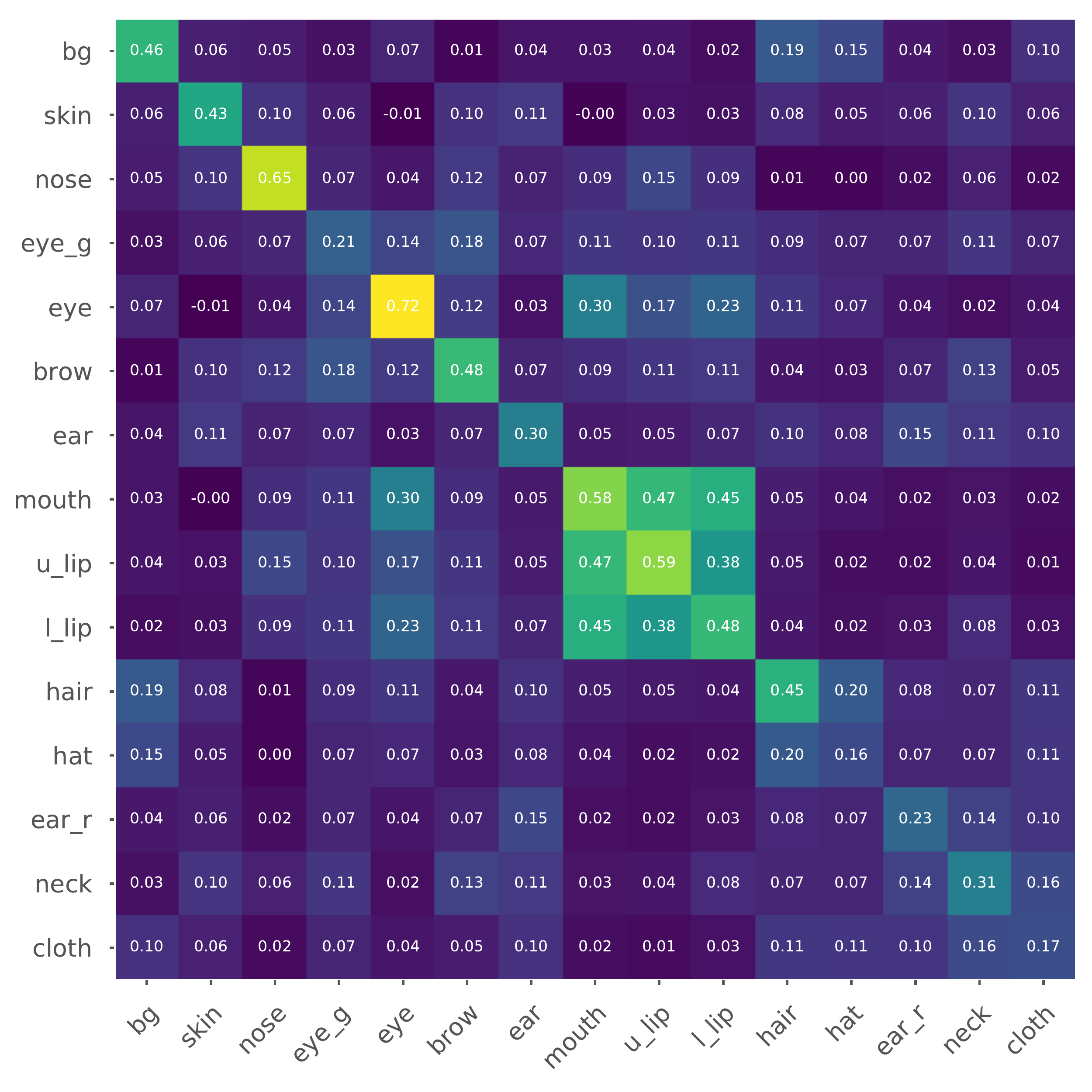}} \\
    \subfloat[StyleGAN2-FFHQ]{\includegraphics[width=0.48\textwidth]{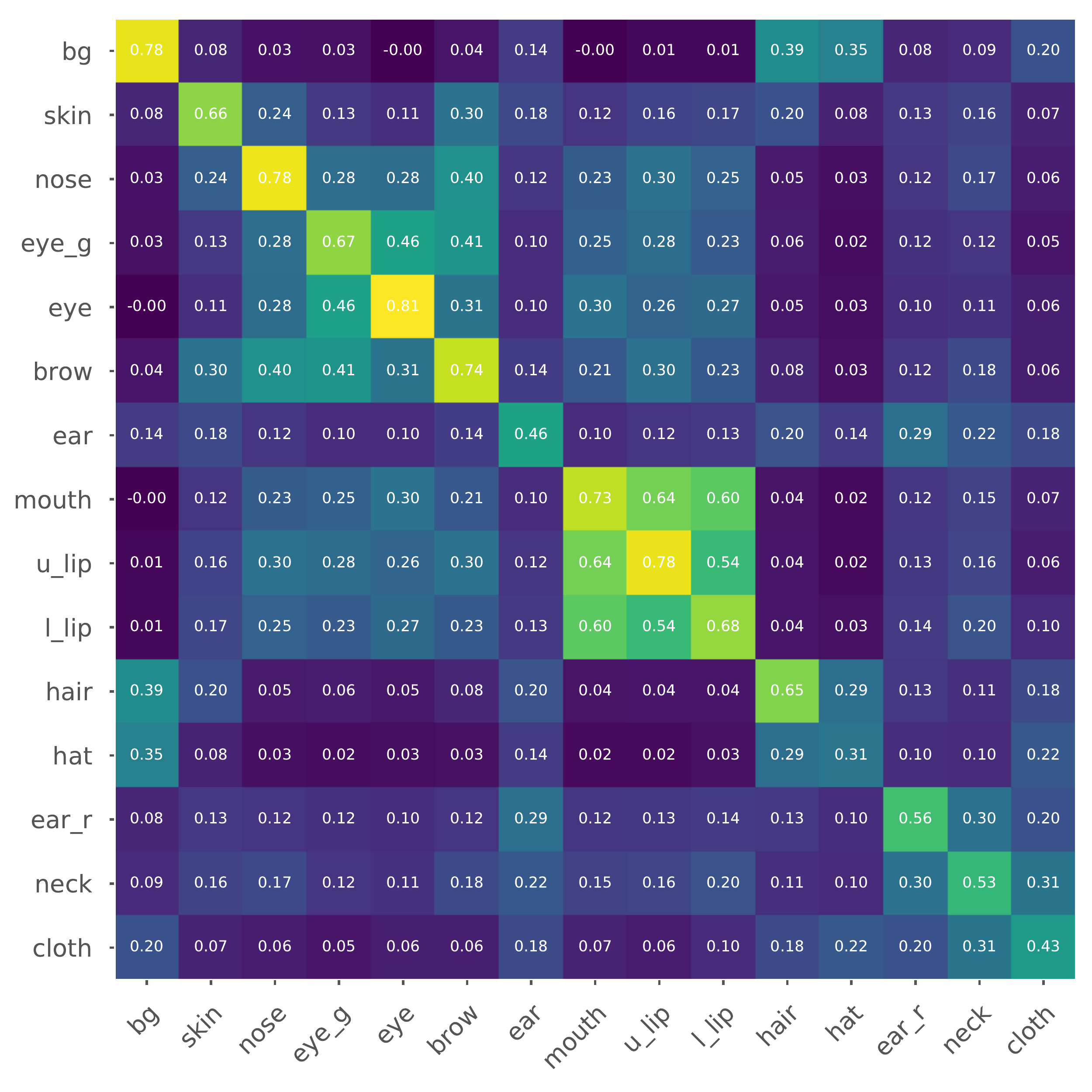}}
    \caption{The cosine similarity between categories. The features for each category are collected using \algref{fair_sample}.}
    \label{fig:cosim}
\end{figure}

As mentioned in \secref{geo}, the linear formulation indicates that the features of a particular category can be bounded by a hyper-cone.
To verify this geometric intuition, we propose to test a stronger hypothesis:
the features of different categories are well-separated.
In other words, the distances of features within a category are closer than those between different categories.
Our approach is to sample features for each category fairly, and compute the cosine distances between features.

First of all, we need to ensure the fairness of comparison for each category.
For this purpose, we propose a fair sampling algorithm (\algref{fair_sample}) which repeatedly samples images and record features fairly until enough features are collected.
In every image, if the feature number of a category is larger than a threshold $T_1$, then $T_1$ feature vectors from that category are chosen randomly without replacement (denoted by $choice(a, N)$).
The chosen vectors would be accumulated to a category feature pool until the feature number reach $T_2$.
The algorithm would terminate when all the category feature pools have collected $T_2$ features.
The fair sampling algorithm gauruantees that each category feature pool consists of $T_1$ randomly chosen vectors from $\frac{T_2}{T_1}$ randomly sampled images.
As the sampling procedure is identical for each category, the sampled features are fair for each category.
In practice, we choose $T_1=200$ and $T_2=4000$.

Second, we calculate the cosine similarity between categories using the fairly sampled features.
Specifically, we first calculate the pairwise cosine similarity between feature vectors of two pools, resulting in a $T_2 \times T_2$ confusion matrix.
The two pools can belong to different categories (inter-class) or the same category (intra-class).
The cosine similarity $cos(A, B)$ between A and B is defined as the mean of the entire matrix.

We show results of StyleGAN-CelebAHQ and StyleGAN2-FFHQ in \figref{cosim}.
Most diagonal elements of the confusion matrix have higher cosine similarity than other elements in a row.
It is indicated that the features in a category are indeed more similar to one another than features between different categories.

\section{Details of Semantic Image Editing} \label{sec:detail_sie}

\begin{algorithm}[t]
    \KwIn{$G$; $L$; $L_reg$; $N$}
    \KwOut{latent code $z$}
    \For{$i=1,\dots,N$}{
      $z_i \leftarrow z_{i-1} + \text{optimizer}(L(z_{i-1}) + L_{\text{reg}}(z_{i-1}))$
    }
    $z \leftarrow z_N$ \\
    \caption{Image editing algorithm.} \label{alg:ie}
\end{algorithm}

\paragraph{Algorithm.} 
A general image editing algorithm is shown in \algref{ie}, whose inputs are the generator $G$, the edit loss $L$, the optional regularization loss $L_{reg}$, and total iteration number $N$.

For color space editing, the editing loss will be the color editing loss $L_c$, defined as $L_c = \frac{1}{||M||_2^2}||M \odot (G(z_i) - C)||_2^2$, where $C$ is the color stroke, $M$ is the mask  of the modified region.
For semantic image editing, the editing loss will be the semantic editing loss $L_s$ as defined in \secref{sie}.
The regularization loss is composed by items including the color preservation loss $L_p = \frac{||(1 - M) \odot (G(z_i) - G(z_0))||_2^2}{||1 - M||_2^2}$, the neighbor regularization loss $L_n = ||z_i - z_0||_2^2$, and the prior regularization loss $L_z = ||z_i||_2^2$.
$z_i$ denotes the latent vector for the $i$-th iteration and $z_0$ denotes the initial latent vector.

For color space editing, its total loss is $L = L_s + 10^{-3} L_n + 10^{-3} L_z$.
For SIE, the total loss is $L = L_c + L_p + 10^{-3} L_n + 10^{-3} L_z$.
We use Adam as the optimizer with default parameters.
The optimization repeats for 50 iterations with a learning rate fixed to be $0.01$.

\paragraph{Usage.}
In practice, our image editing application works in two steps:
The first step is to annotate 1 to 8 images sampled from GAN.
The backend of the application will then train a few-shot LSE using the annotations.
The second step is to edit any sampled images.
The editing interface will provide the semantic mask extracted by the few-shot LSE along with the image.
When the user wants to edit an image, he draws some strokes on the semantic mask to form a target mask.
Then, the backend would run the SIE algorithm and return an image that is closer to the target.

\section{Semantic-Conditional Sampling.} \label{sec:detail_scs}

\paragraph{Algorithm.}
To sample an image matching the given mask, we first try to find a good initialization.
We randomly sample $n_{\text{init}}$ latent codes and select the initialization to be the one closest to the target mask.
Next, we iteratively optimize the latent code to match the target mask using the cross-entropy loss defined in \eq{loss}.
The semantic masks can be predicted using either a pretrained segmenter or a few-shot LSE. 

\begin{algorithm}[b]
    \KwIn{$G$; $P$; $Y$; $n_{\text{init}}$; $N$}
    \KwOut{latent code $z$}
    $\bar{z}_{i} \sim N(0, I)$, $i=1,\dots,n_{\text{init}}$ \\
    $S_i = P(G(\bar{z}_{i}))$ \\
    $P_i = |\{p | S_{i,p} = Y_p\}|$ \\
    $z_0 = \bar{z}_{\alpha}$, $\alpha = argmin_i P_i$ \\
    \For{$i=1,\dots,N$}{
      $L = \mathcal{L}(P(G(z_{i-1})), Y)$ \\
      $z_i = \text{optimizer}(L, z_{i-1})$ \\
    }
    $z \leftarrow z_N$ \\
    \caption{Semantic-Conditional Sampling algorithm.} \label{alg:scs}
\end{algorithm}

The SCS algorithm is defined formally in \algref{scs}.
Its inputs are the current latent code $z$, the target semantic segmentation $Y$, the generator $G$, the semantic predictor $P$, the initialization number $n_{\text{init}}$, and the iteration number $N$.
Its output will be image samples that respect the given mask $Y$.

In practice, we use $n_{\text{init}}=10$ for SCS on face images.
$n_{\text{init}}=100$ is used for bedroom and church images, as they are much more diverse than faces. The optimization is repeated for 50 iterations.
The optimizer is Adam with default hyperparameters (lr=$10^{-3}$, $\beta_1=0.9$, $\beta_2=0.999$).
These settings are manually selected without tuning.

For SCS on facial images, the target masks are selected randomly from the annotations in the CelebAMask-HQ \cite{lee2019maskgan} dataset.
For bedroom and church, the masks are predicted from images sampled from truncated latent space, which has better image quality than the full latent space \cite{karras2019style}.
The truncated latent space $\mathcal{W}^-$ is obtained by truncating the latent vectors of $\mathcal{W}$ within a distance of the statistical center.

\paragraph{Evalution.}
Our proposed method plugs in a few-shot LSE for $P$, while the baseline uses a pretrained segmentation network as $P$.
To evaluate the performance of SCS models, we again rely on a pretrained segmentation network, $P^*$.
In this work, the pretrained network used by the baseline is exactly the same as the one used in evaluation.
This is slightly biased toward the baseline, yet our method is still able to match or surpass the baseline.

Formally, let the set of targets be $\mathcal{Y}$.
The images sampled by a SCS model given a target $Y_i$ are denoted as a set $\mathcal{I}_i$.
The semantic agreement $A$ of sampled images can be measured by the mean IoU between the predicted segmentation masks and the target mask:

\begin{equation}\label{eq:scs_miou}
  A(\mathcal{I}, \mathcal{Y}; P^*) = \sum_{ \substack{1 \le i \le |\mathcal{Y}| \\ 1 \le j \le |\mathcal{I}_i|}} \frac{1}{|\mathcal{I}_i||\mathcal{Y}|} \text{mIoU}(Y_i, P^*(I_{i,j}))
\end{equation}

In practice, we select 100 target masks and conditionally sample 10 images for each target, i.e., $|\mathcal{Y}|=100$ and $|\mathcal{I}_i|=10$.
As a result, we obtain 1,000 images for the evaluation of each setting of SCS.
To account for the variance of few-shot LSEs, we repeat the training for each model 5 times, as mentioned in \secref{fewshot_LSE}.



\section{Layerwise analysis}\label{sec:layerwise}

\begin{figure*}
    \centering
    \includegraphics[trim=20 0 20 10,clip,width=0.49\linewidth]{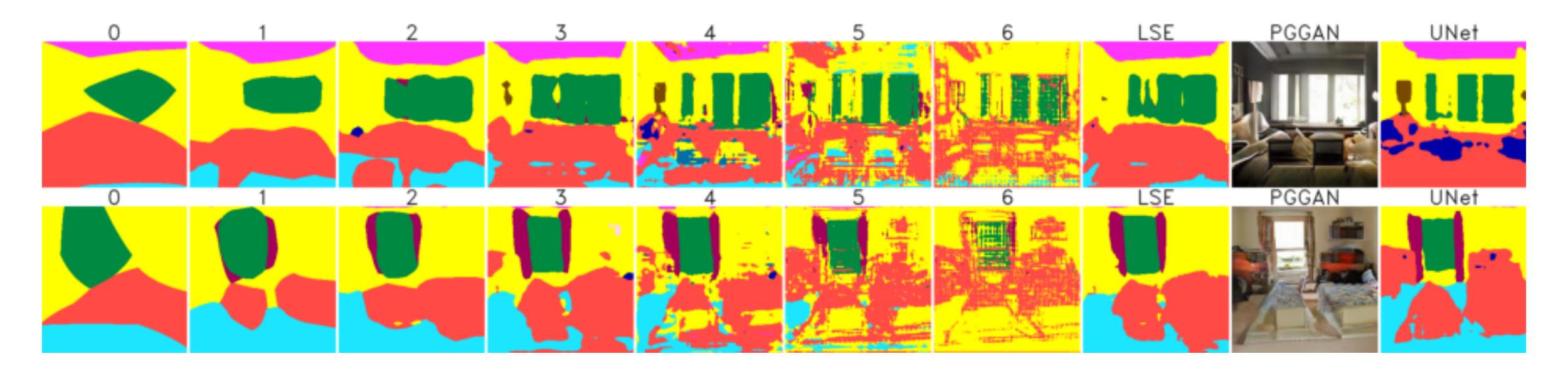}
    \includegraphics[trim=20 0 20 10,clip,width=0.49\linewidth]{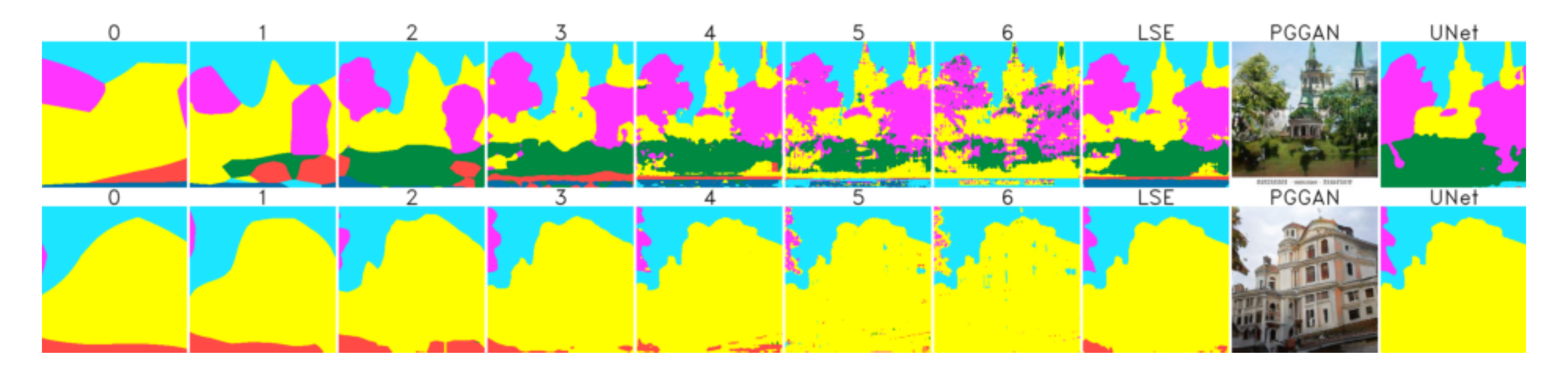}
    \includegraphics[trim=20 5 20 10,clip,width=0.49\linewidth]{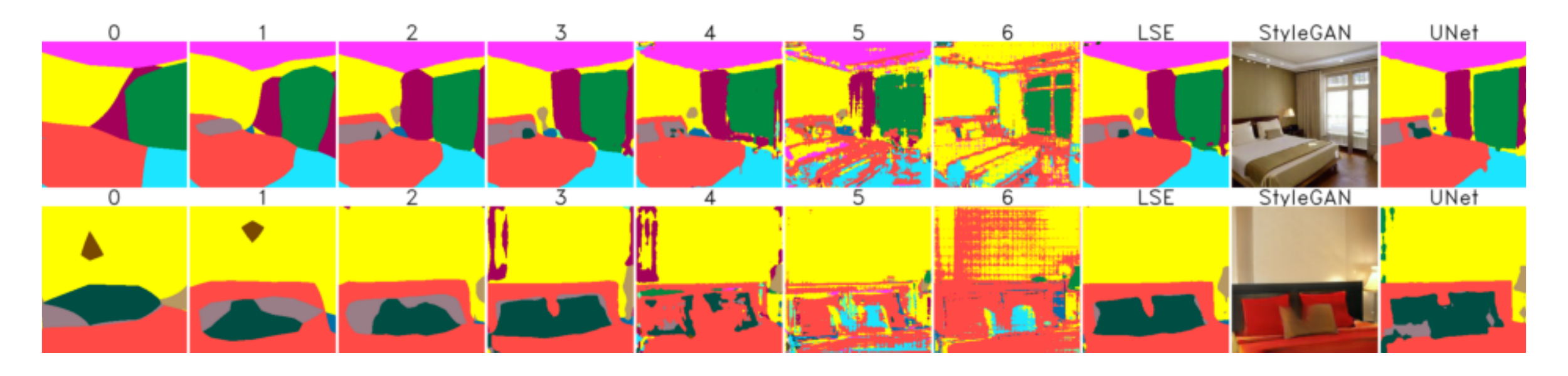}
    \includegraphics[trim=20 5 20 10,clip,width=0.49\linewidth]{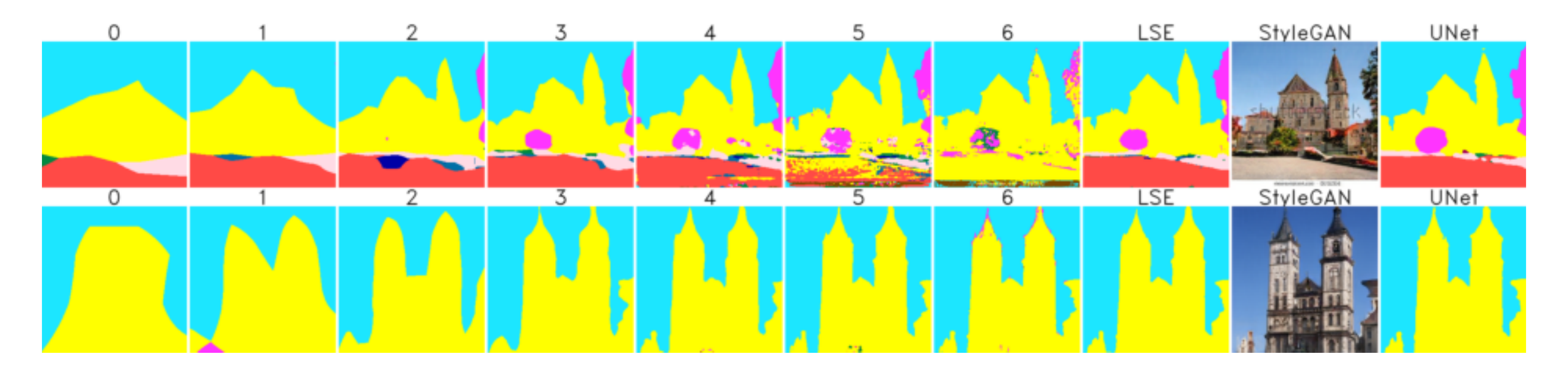}
    \includegraphics[trim=10 10 10 10,clip,width=0.49\linewidth]{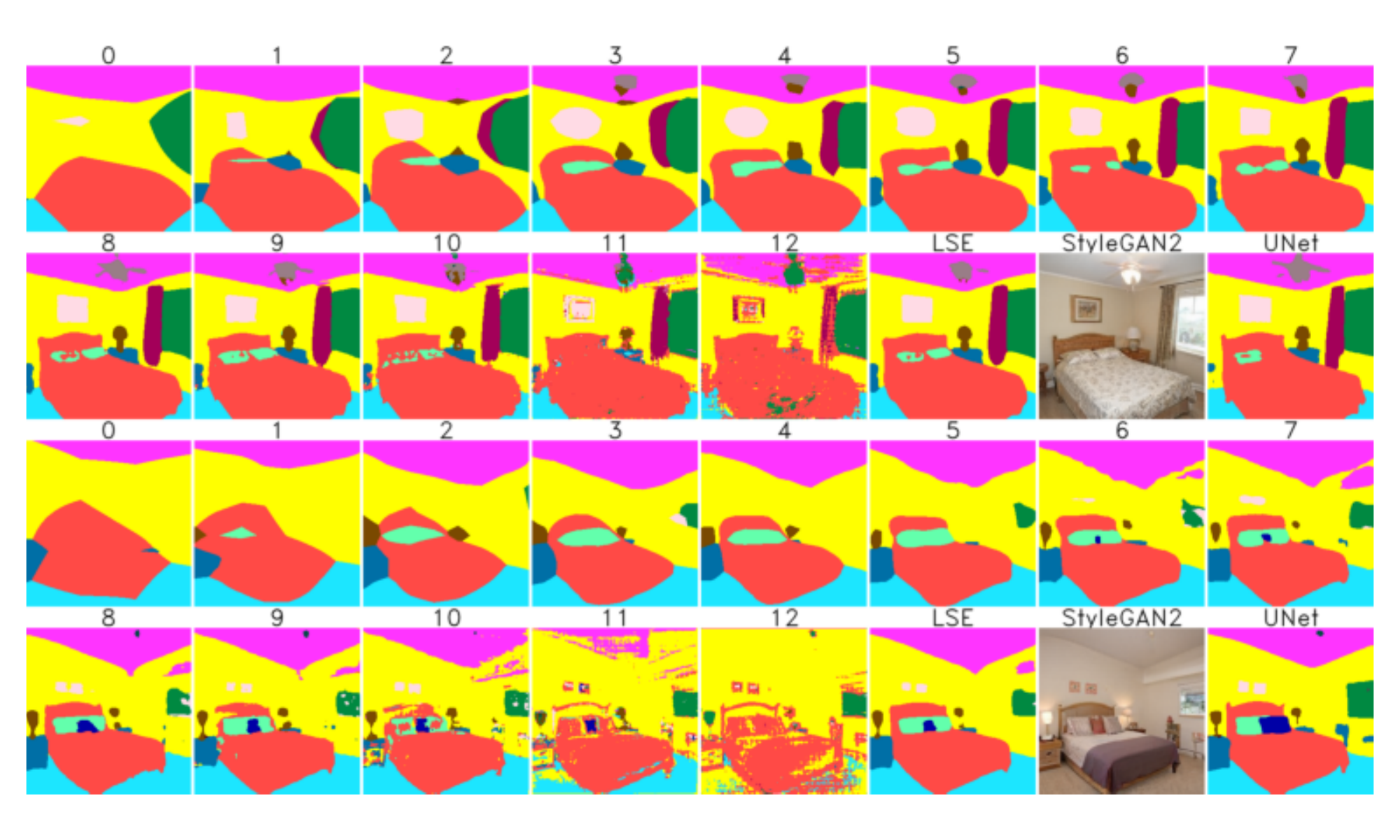}
    \includegraphics[trim=10 10 10 10,clip,width=0.49\linewidth]{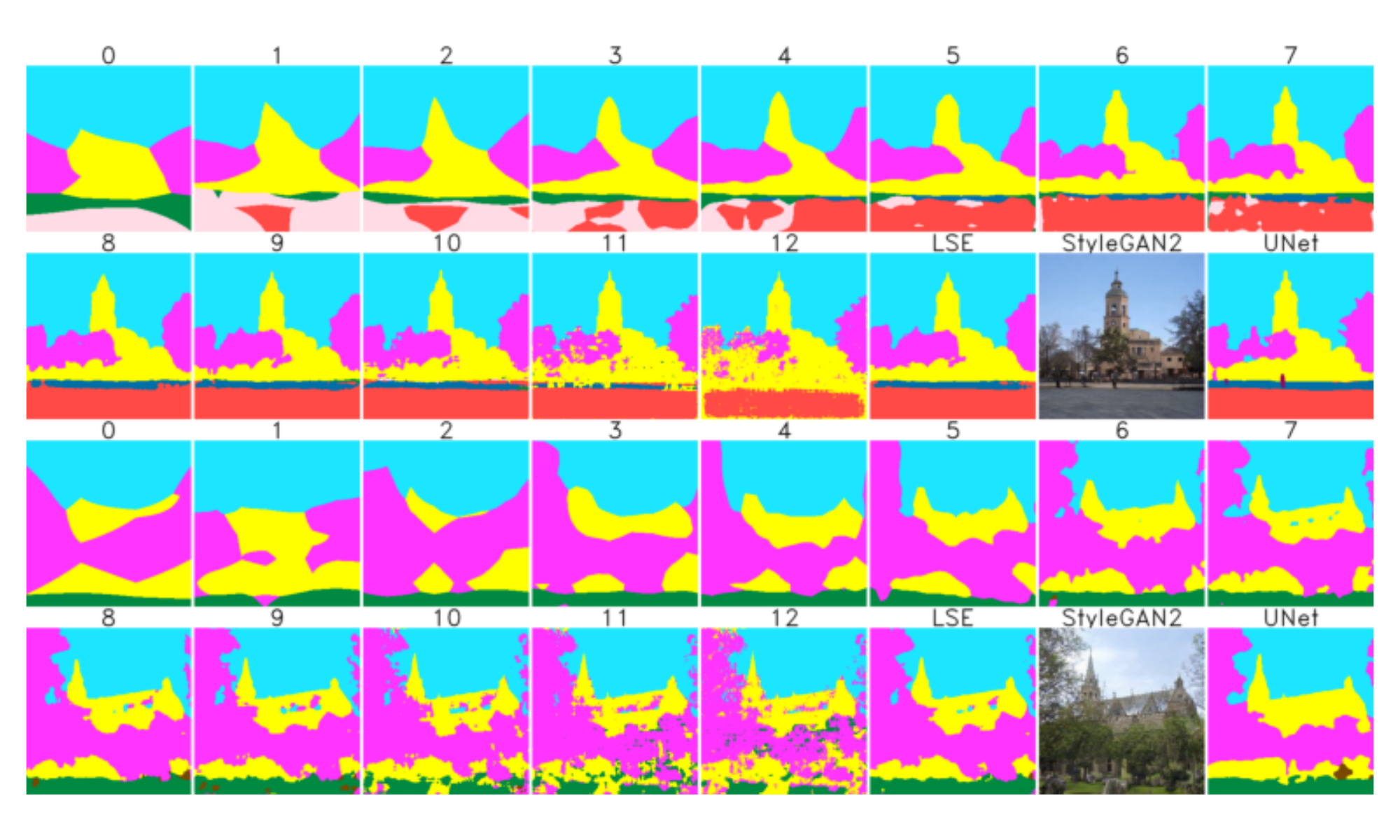}
    \includegraphics[trim=10 15 10 20,clip,width=0.49\linewidth]{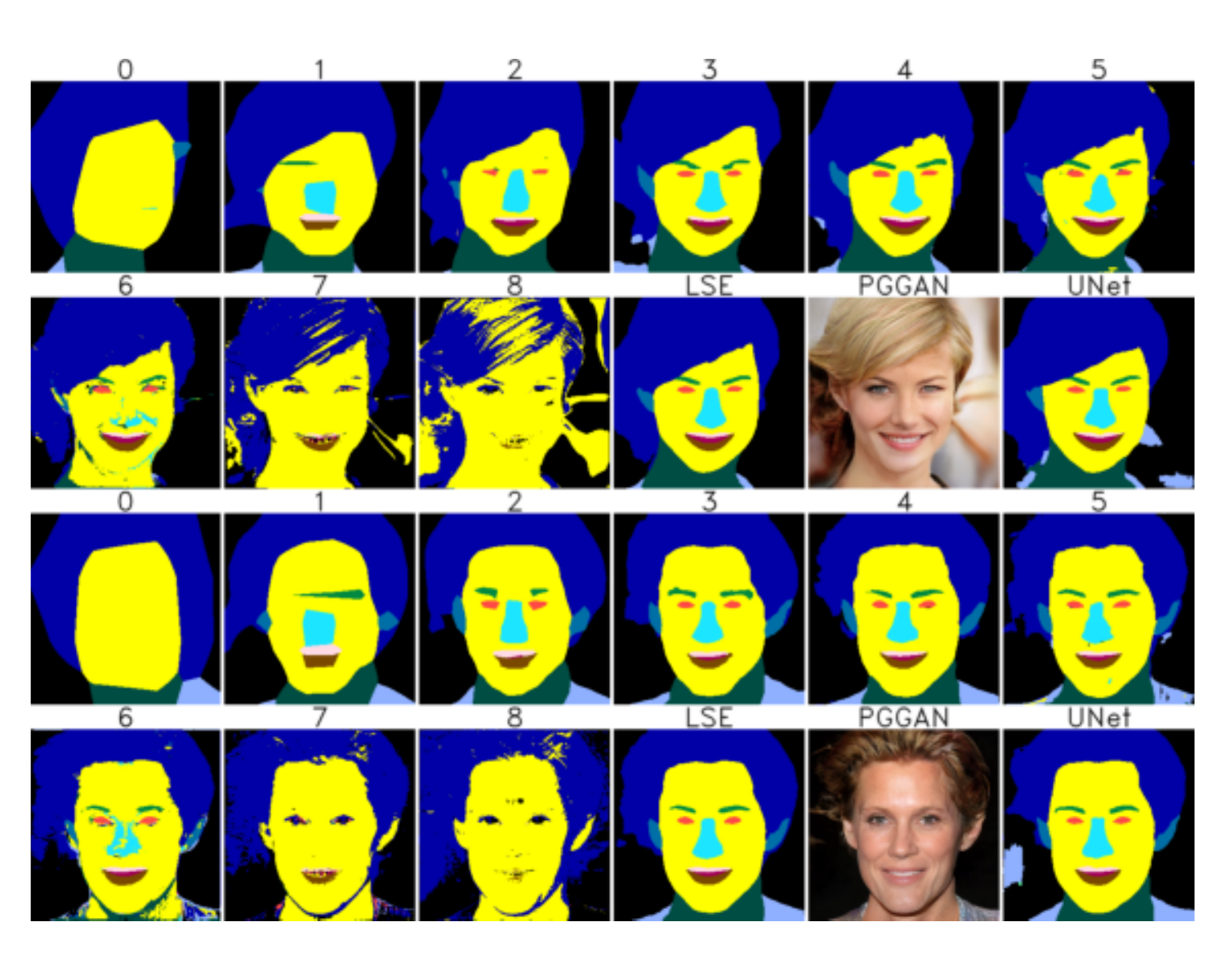}
    \includegraphics[trim=10 15 10 20,clip,width=0.49\linewidth]{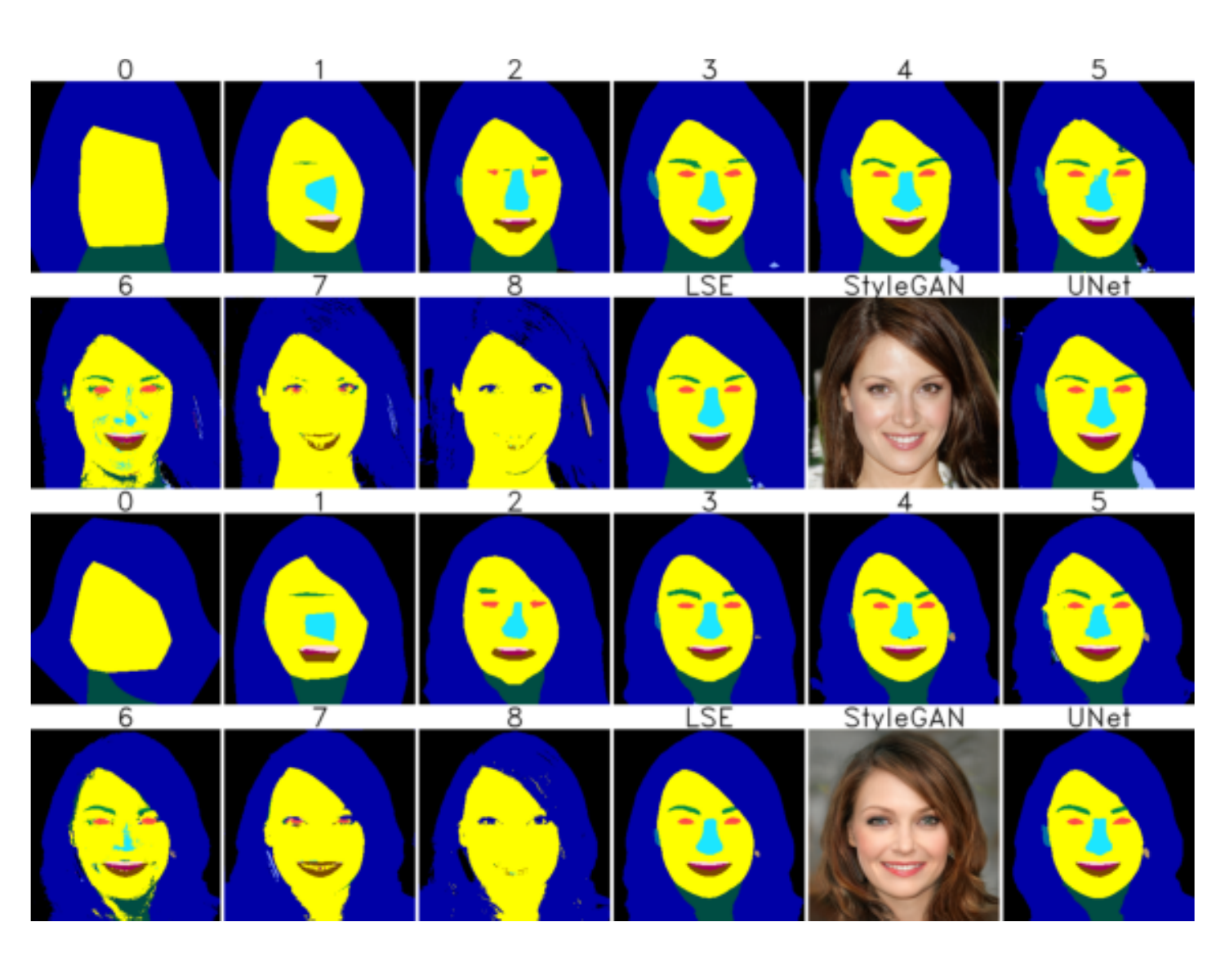}
    \includegraphics[trim=10 15 10 15,clip,width=0.83\linewidth]{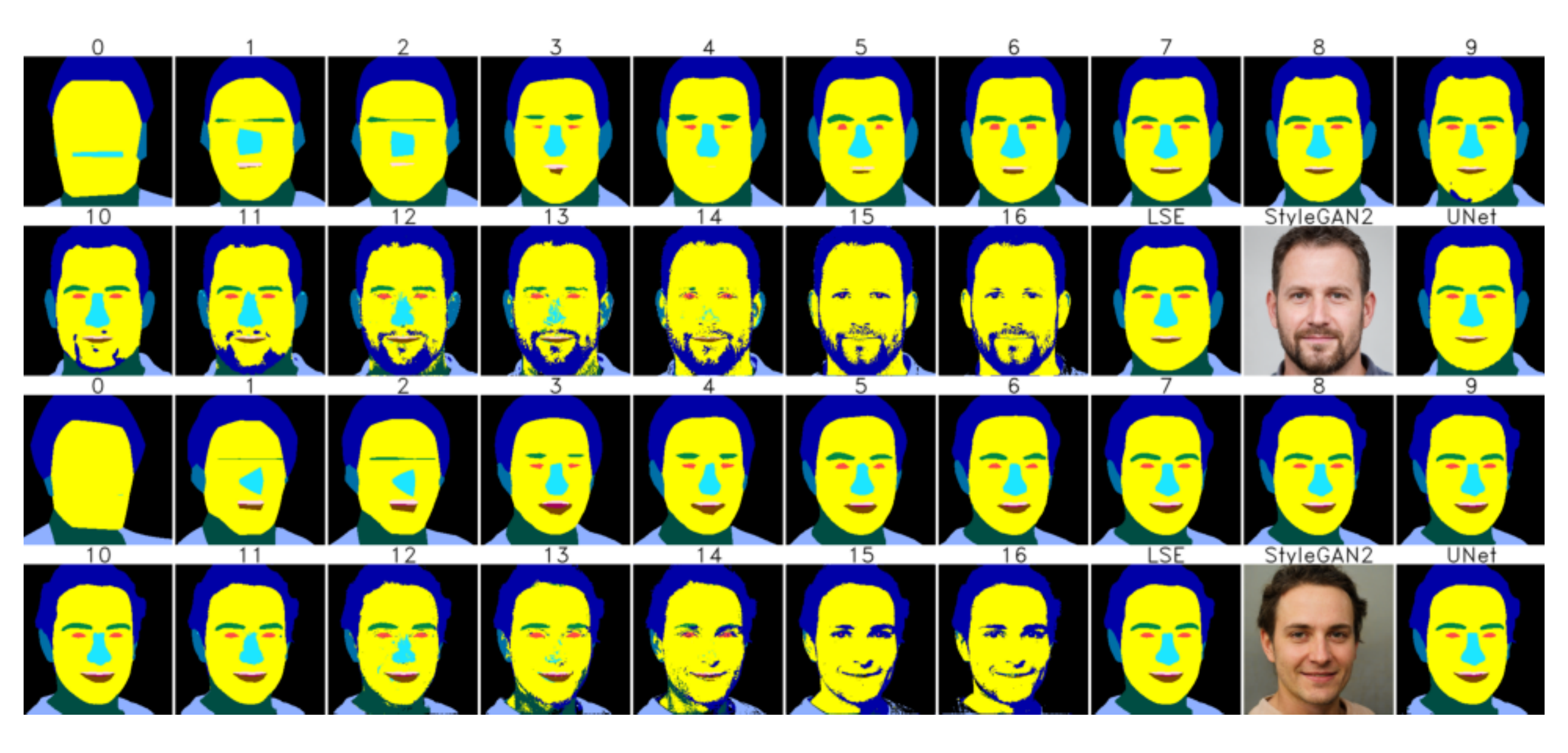}
    \caption{The layer-wise semantics extracted from PGGAN, StyleGAN, and StyleGAN2. Layer indices are shown in the headers.}
    \label{fig:layer}
\end{figure*}

\begin{table*}[t]
    \centering
    \resizebox{\linewidth}{!}{
    \begin{tabular}{c|cccccccccccccc}
    \whline{1.0pt}
     & skin & nose & eye-g & eye & brow & ear & mouth & u-lip & l-lip & hair & hat & ear-r & neck & cloth \\\hline
    & \multicolumn{14}{c}{StyleGAN2-FFHQ} \\ \hline
    LSE & 95.9\% & 94.7\% & 69.9\% & 91.0\% & 83.5\% & 80.5\% & 84.5\% & 87.8\% & 91.2\% & 92.9\% & 11.1\% & 22.8\% & 91.0\% & 72.5\% \\
    NSE-1 & 97.0\% & 95.4\% & 72.4\% & 92.1\% & 88.2\% & 83.0\% & 87.4\% & 91.6\% & 92.8\% & 94.2\% & 12.9\% & 34.0\% & 92.9\% & 75.9\% \\
    NSE-2 & 96.9\% & 95.3\% & 73.4\% & 92.0\% & 87.7\% & 82.8\% & 87.7\% & 90.9\% & 92.9\% & 94.1\% & 12.9\% & 28.6\% & 92.4\% & 72.8\% \\\hline
    & \multicolumn{14}{c}{StyleGAN-CelebAHQ} \\ \hline
    LSE & 93.9\% & 91.3\% & 25.7\% & 86.2\% & 75.9\% & 63.5\% & 75.6\% & 81.1\% & 85.4\% & 87.5\% & 0.0\% & 13.1\% & 84.5\% & 35.9\% \\
    NSE-1 & 95.8\% & 93.6\% & 22.8\% & 89.3\% & 83.2\% & 69.4\% & 78.8\% & 87.4\% & 88.7\% & 90.8\% & 0.3\% & 21.3\% & 88.0\% & 41.2\% \\
    NSE-2 & 96.0\% & 94.1\% & 22.1\% & 89.4\% & 84.7\% & 69.7\% & 79.0\% & 88.0\% & 89.5\% & 90.9\% & 0.0\% & 19.0\% & 87.8\% & 39.2\% \\\hline
    & \multicolumn{14}{c}{PGGAN-CelebAHQ} \\ \hline
    LSE & 92.7\% & 89.4\% & 19.7\% & 84.9\% & 71.7\% & 61.9\% & 72.4\% & 81.4\% & 84.7\% & 85.2\% & 5.0\% & 16.1\% & 79.8\% & 34.1\% \\
    NSE-1 & 93.8\% & 90.9\% & 22.0\% & 86.3\% & 78.4\% & 63.0\% & 71.6\% & 83.0\% & 85.6\% & 86.4\% & 6.3\% & 20.3\% & 81.5\% & 37.0\% \\
    NSE-2 & 94.1\% & 92.0\% & 20.8\% & 86.2\% & 78.9\% & 64.4\% & 73.0\% & 83.9\% & 86.4\% & 86.9\% & 6.2\% & 21.3\% & 82.2\% & 37.4\% \\
    \whline{1.0pt}
    \end{tabular}}

    \caption{The IoU for each category (excluding background category) of LSE, NSE-1 and NSE-2. The ground-truth used in the IoU computation is obtained from UNet.}
    \label{tab:cat_face}
\end{table*}

During this work, we examined layer-wise semantics, which refers to the semantics extracted from each layers alone.
As the original training objective \eq{loss} only optimizes the summation from all layers, the semantics from each layer may not be a good segmentation individually.
To extract the layer-wise semantics better, we add a cross-entropy loss term on each layer.
Theoretically, the best possible layer-wise semantics should be obtained by training only on that layer.
However, the computational cost would then be prohibitive.
Thus, we choose to optimize all layer losses together.

To put it more formally, we denote the output of LSE on layer $i$ to be $\bS_i = \T_i \cdot \xx_i$ (the final segmentation is $\bS = \sum_{i=1}^{N-1} \mathsf{u}_{i}^\uparrow(\bS_i)$).
The training objective becomes

\begin{equation} \label{eq:layer_loss}
    \mathcal{L}_{l} = \mathcal{L}(\bS, Y) + \sum_{i=1}^{N-1} \alpha_i \mathcal{L}(\mathsf{u}_{i}^\uparrow(\bS_i), Y),
\end{equation}

where $Y$ is the segmentation label, $\mathcal{L}$ is the standard cross-entropy loss, and $\alpha_i$ is the coefficient for each layer.
In practice, we set $\alpha_i=0.1$.
The training procedure is exactly the same.

The visualizations of layer-wise semantics are shown in \figref{layer}.
Our main discoveries are twofold: \textbf{(1)} the semantic
layout in each layer becomes refined as the network layer progresses from input to output; 
\textbf{(2)} the most semantically rich layers are often near the middle layer of the network. 
However, it remains unclear how to make use of the layer-wise semantics and we choose to leave this question for future research.

\section{Supplementary results}\label{sec:supp_results}

\begin{figure*}
    \centering
    \subfloat[Face datasets.]{\includegraphics[trim=20 10 20 10,clip,width=0.33\linewidth]{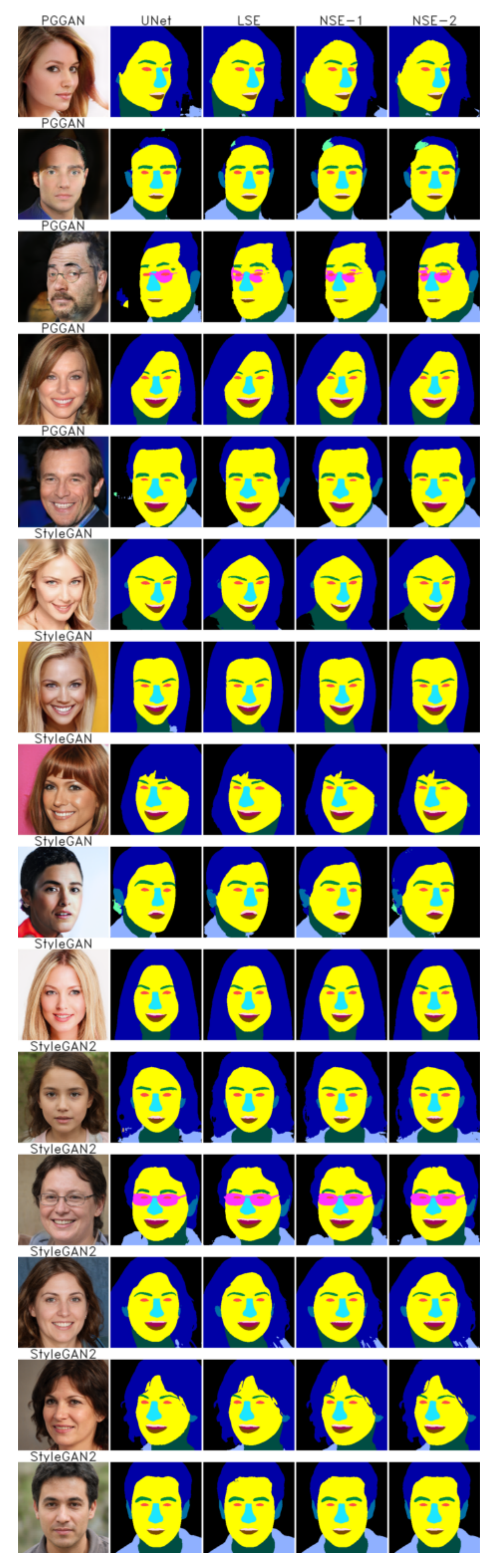}}
    ~
    \subfloat[LSUN-bedroom dataset.]{\includegraphics[trim=20 10 20 10,clip,width=0.33\linewidth]{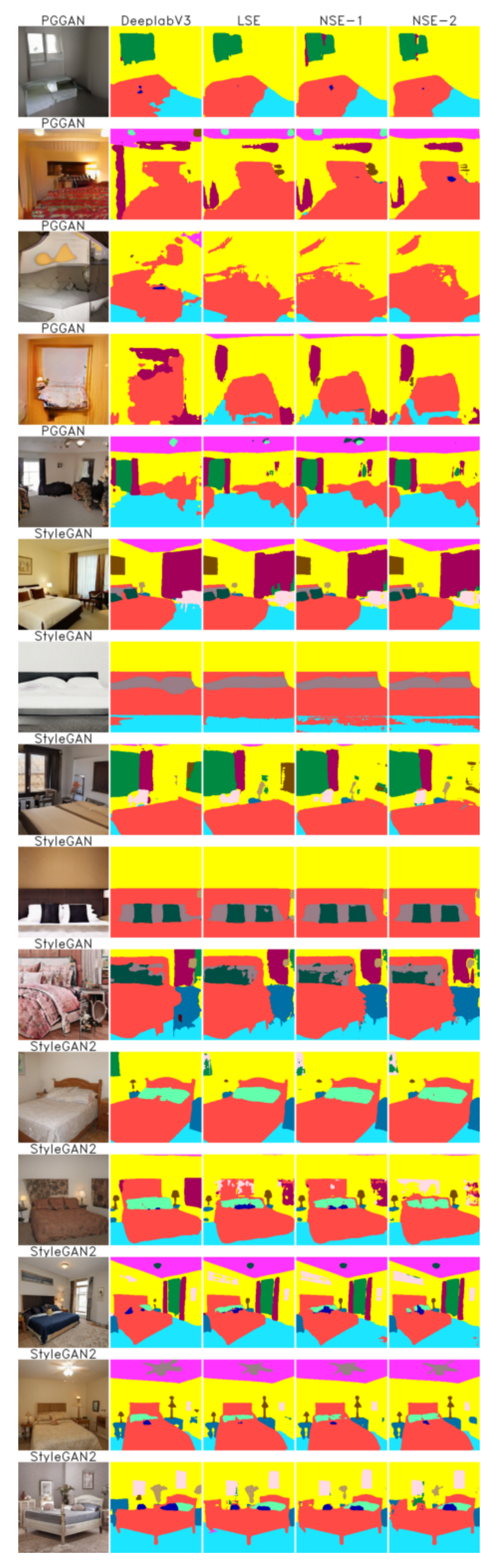}}
    ~
    \subfloat[LSUN-church dataset.]{\includegraphics[trim=20 10 20 10,clip,width=0.33\linewidth]{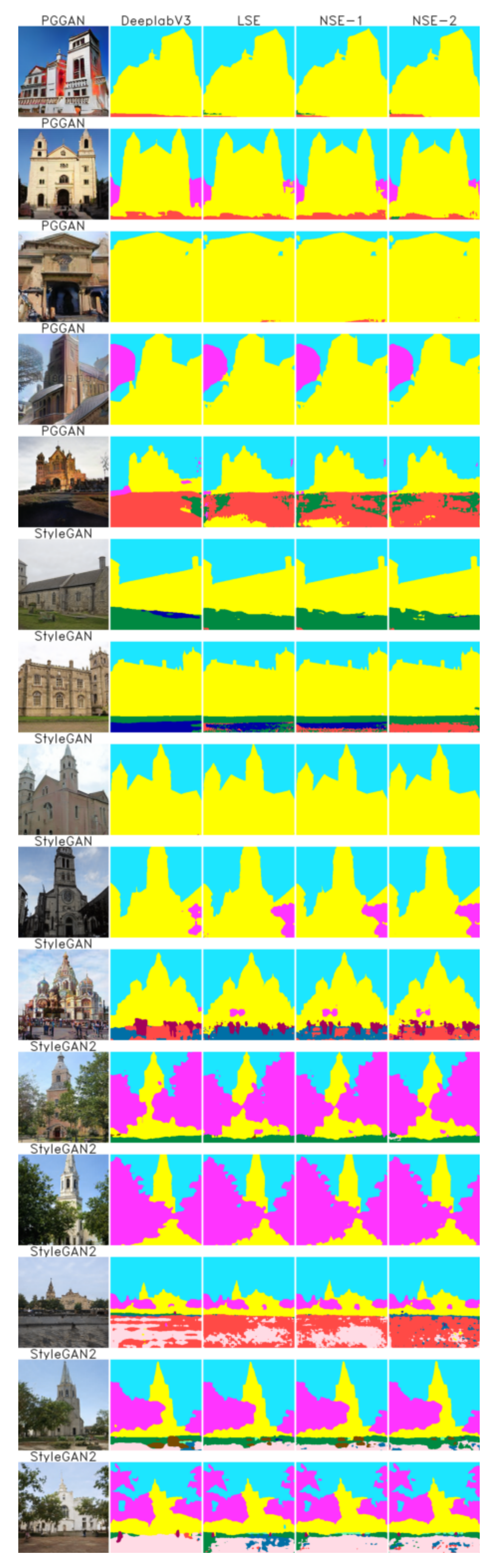}}
    \caption{Qualitative comparisions of LSEs and NSEs. For each GAN, 5 samples are shown.}
    \label{fig:supple_qualitative}
\end{figure*}

Additional qualitative results comparing LSEs and NSEs are shown in ~\figref{supple_qualitative}.

Category IoUs for bedroom and church models are summarized in \tabref{cat_bedroom_church}.
They are shown together with category IoUs from models trained with full ADE20K categories.
For face GANs, the results are shown in \tabref{cat_face}.

\begin{figure*}
    \centering
    \includegraphics[width=0.9\linewidth]{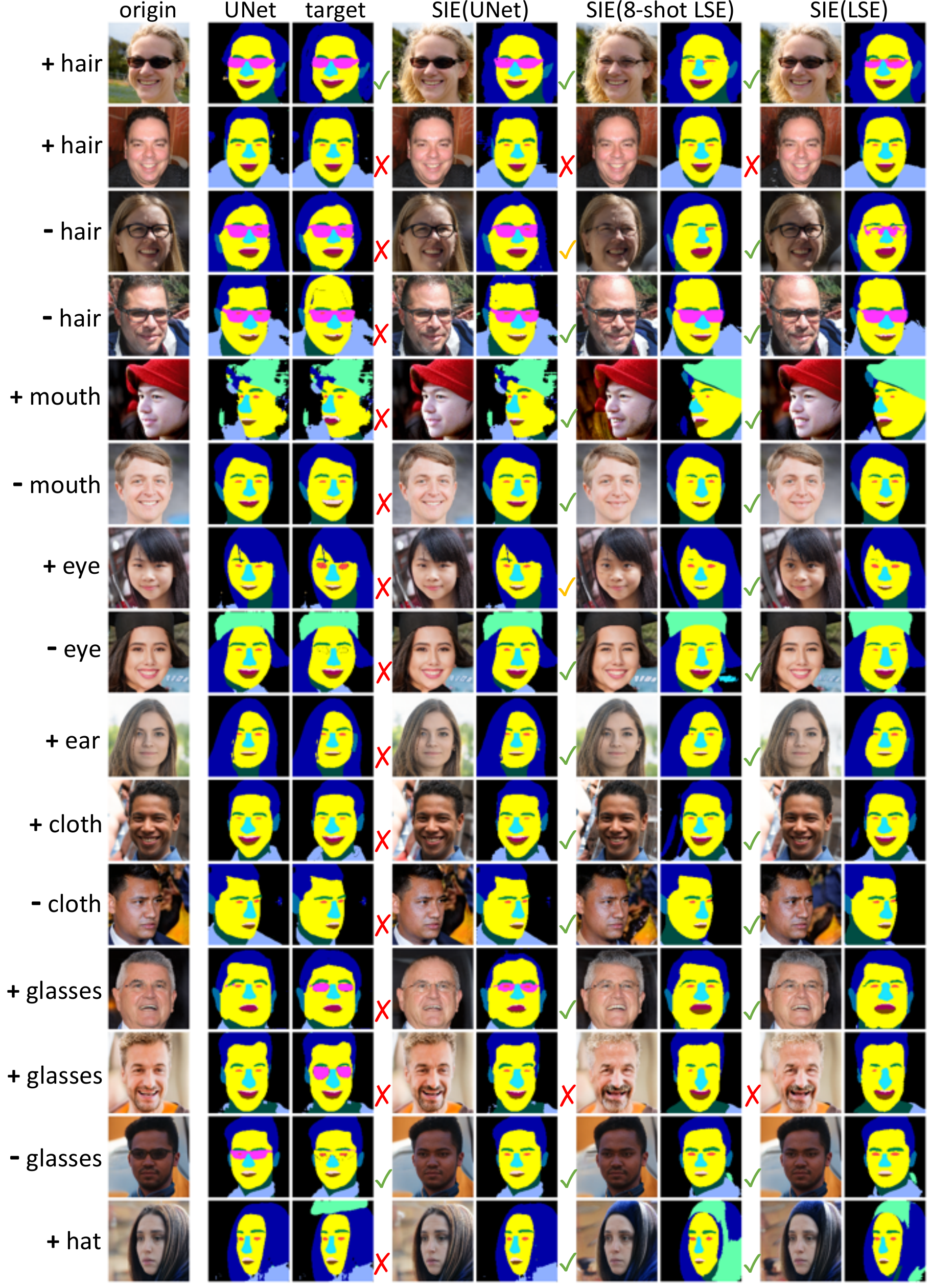}
    \caption{More SIE results on StyleGAN2-FFHQ. Annotations on the left are users' edit intentions. The following columns are original images, the face segmentation from UNet, the modified semantic mask by the user, the results from SIE(UNet), SIE(8-shot LSE), and SIE(LSE), respectively. The green ticks and red crosses represent whether the editing success or not. Other yellow ticks indicate that the image quality degrades.}
    \label{fig:sie_supp}
\end{figure*}

More results for Semantic Image Editing are shown in \figref{sie_supp}.

\begin{figure*}
    \centering
    \subfloat[SCS(1-shot LSE)]{\includegraphics[trim=0 0 810 0,clip,width=0.49\linewidth]{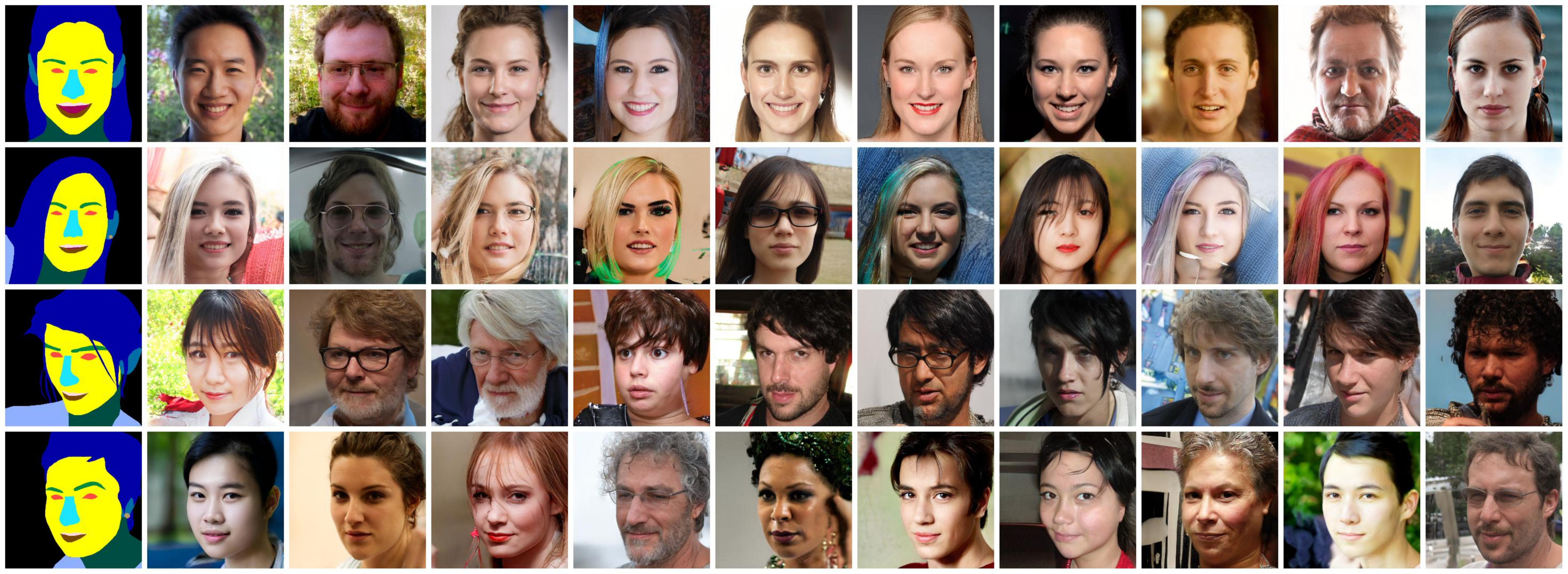}}
    ~
    \subfloat[SCS(4-shot LSE)]{\includegraphics[trim=0 0 810 0,clip,width=0.49\linewidth]{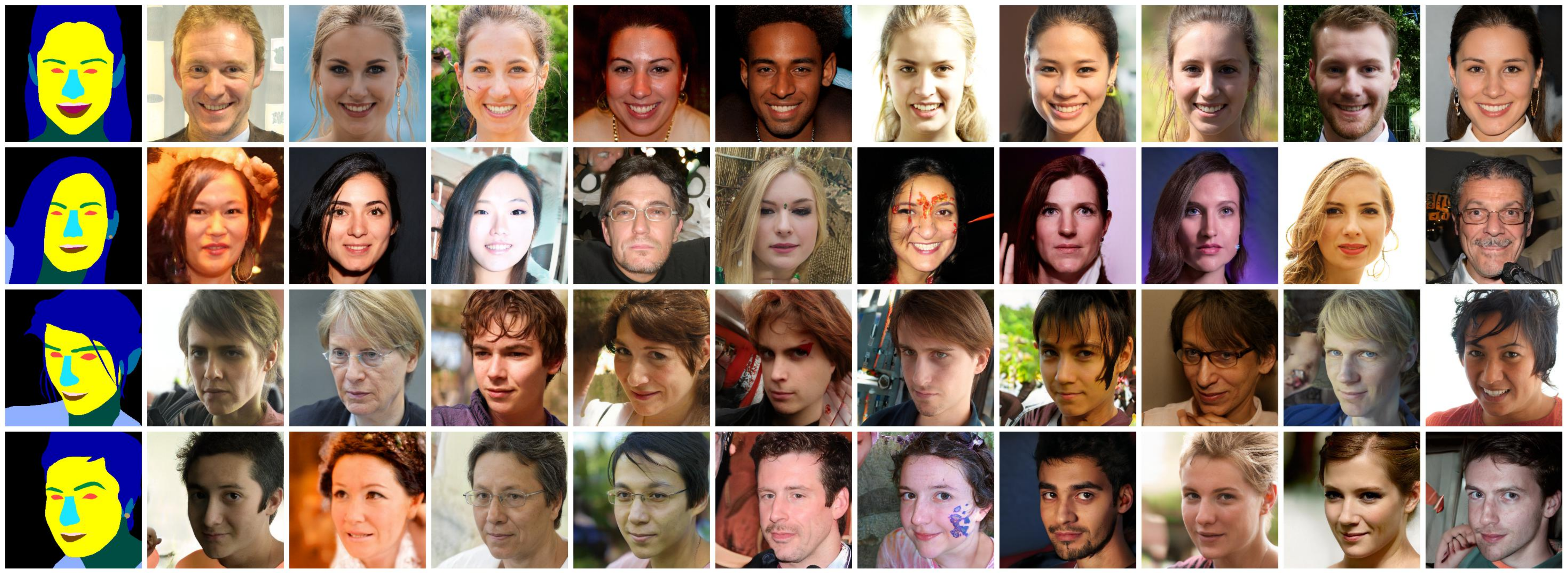}}
    \\
    \subfloat[SCS(8-shot LSE)]{\includegraphics[trim=0 0 810 0,clip,width=0.49\linewidth]{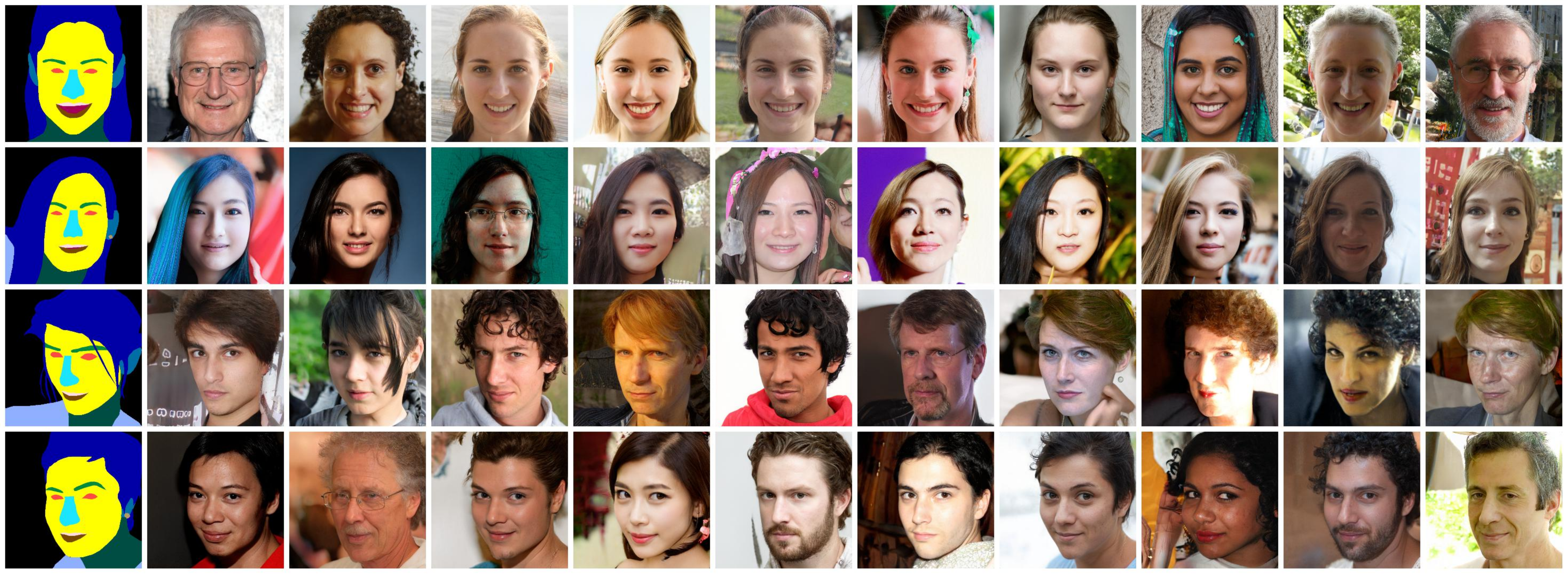}}
    ~
    \subfloat[SCS(16-shot LSE)]{\includegraphics[trim=0 0 810 0,clip,width=0.49\linewidth]{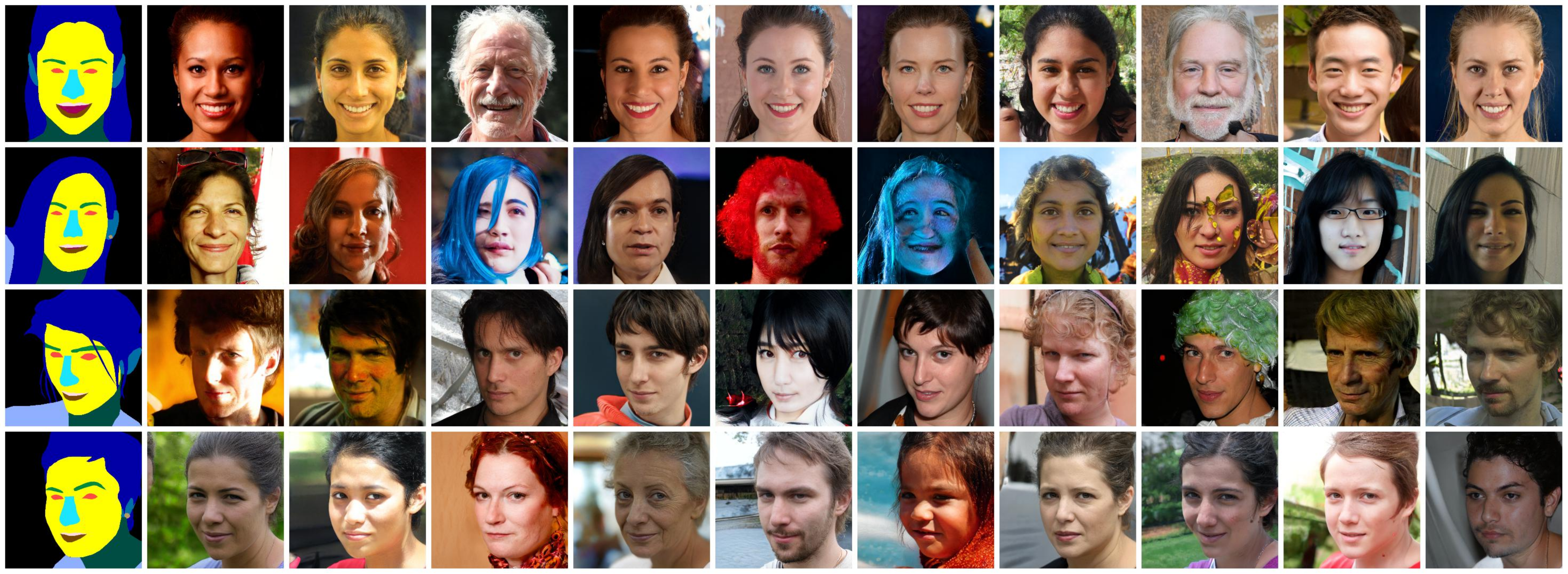}}
    \\
    \subfloat[SCS(UNet)]{\includegraphics[width=0.99\linewidth]{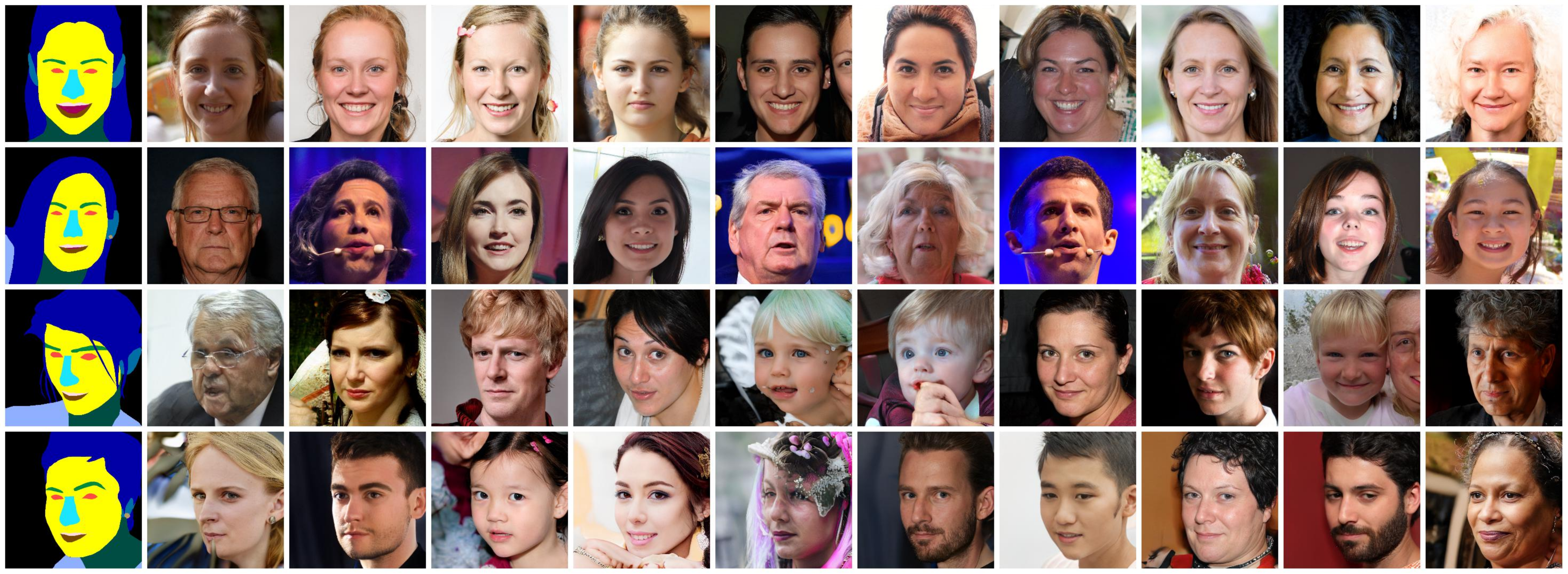}}
    \caption{The results of SCS on StyleGAN2-FFHQ using LSEs and UNet.}
    \label{fig:scs_ffhq_supp}
\end{figure*}

\begin{figure*}
    \centering
    \subfloat[SCS(1-shot LSE)]{\includegraphics[trim=0 0 810 0,clip,width=0.49\linewidth]{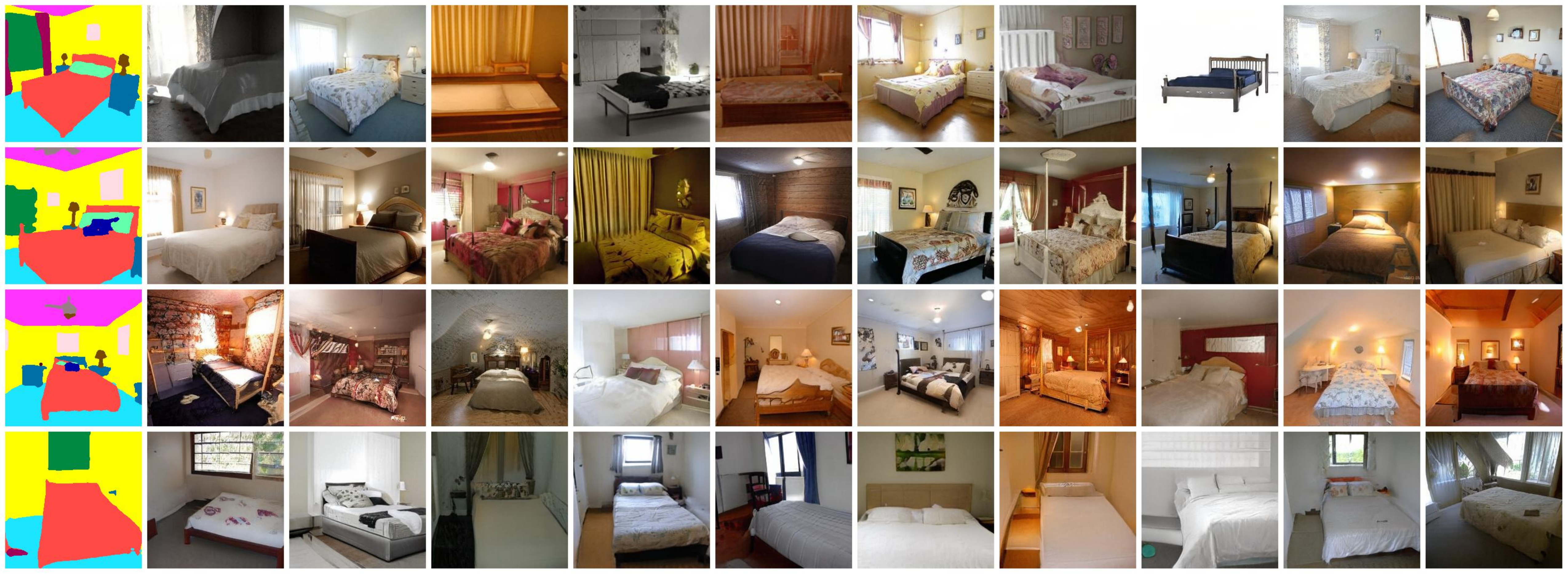}}
    ~
    \subfloat[SCS(4-shot LSE)]{\includegraphics[trim=0 0 810 0,clip,width=0.49\linewidth]{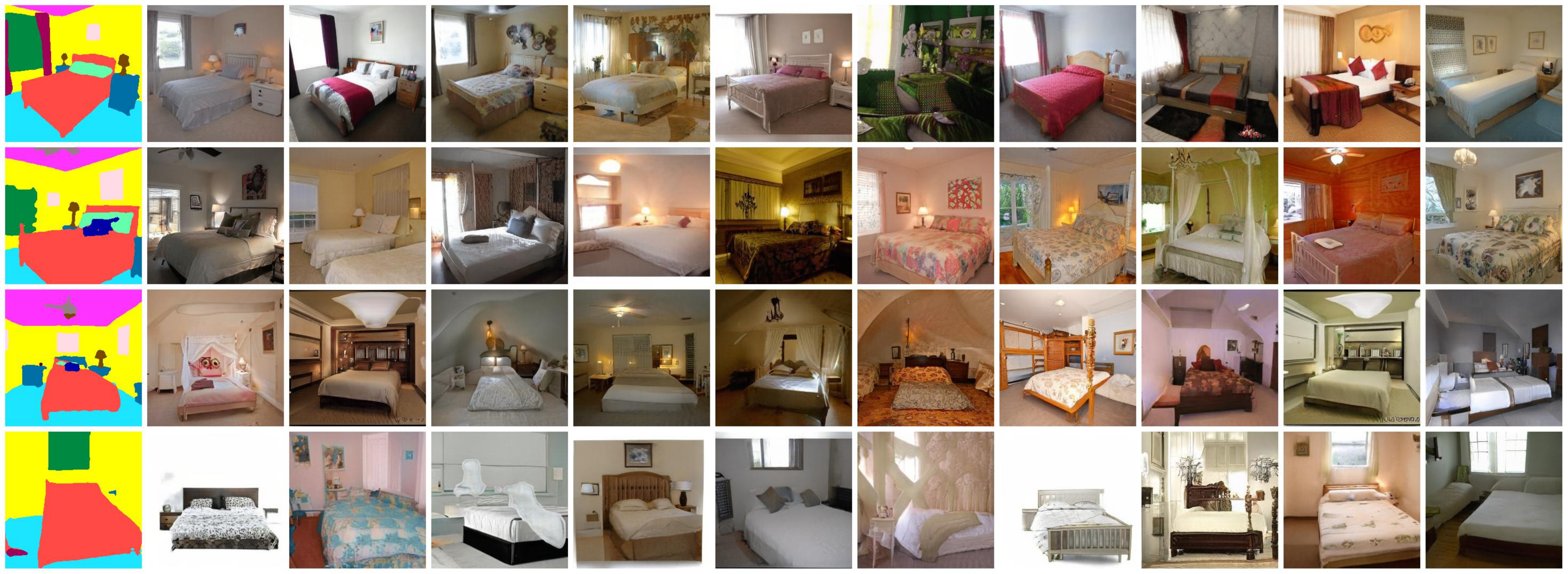}}
    \\
    \subfloat[SCS(8-shot LSE)]{\includegraphics[trim=0 0 810 0,clip,width=0.49\linewidth]{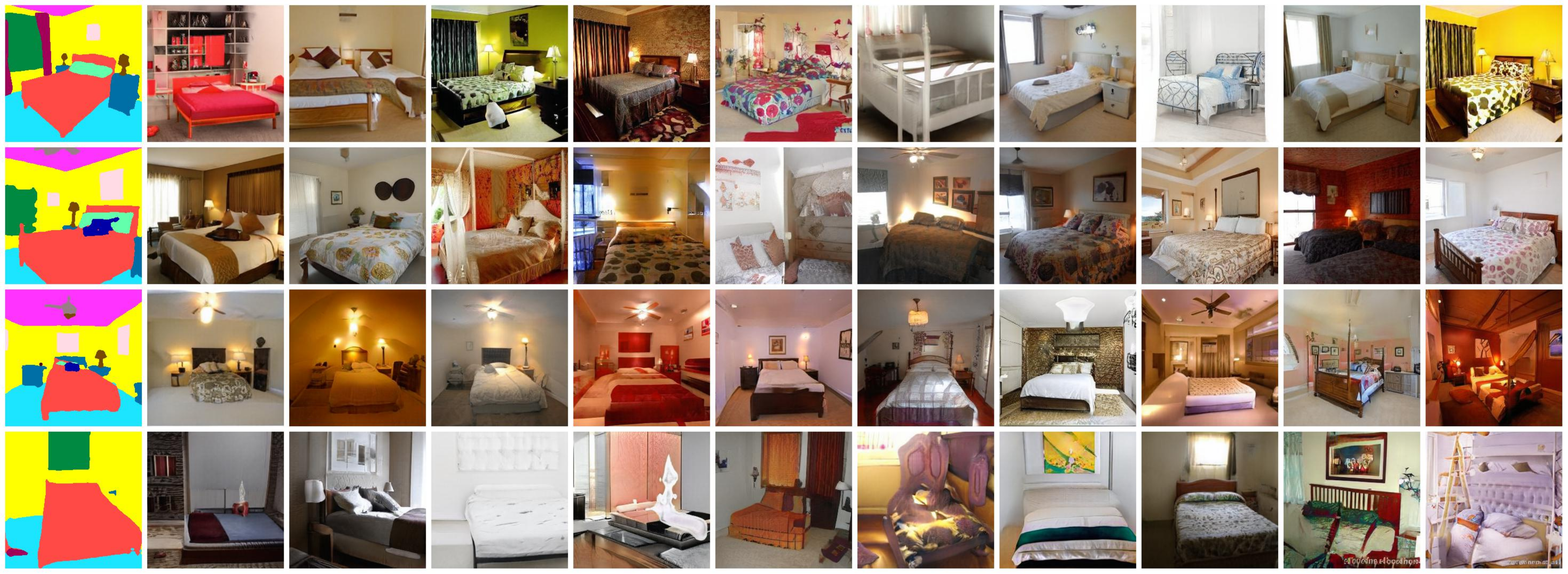}}
    ~
    \subfloat[SCS(16-shot LSE)]{\includegraphics[trim=0 0 810 0,clip,width=0.49\linewidth]{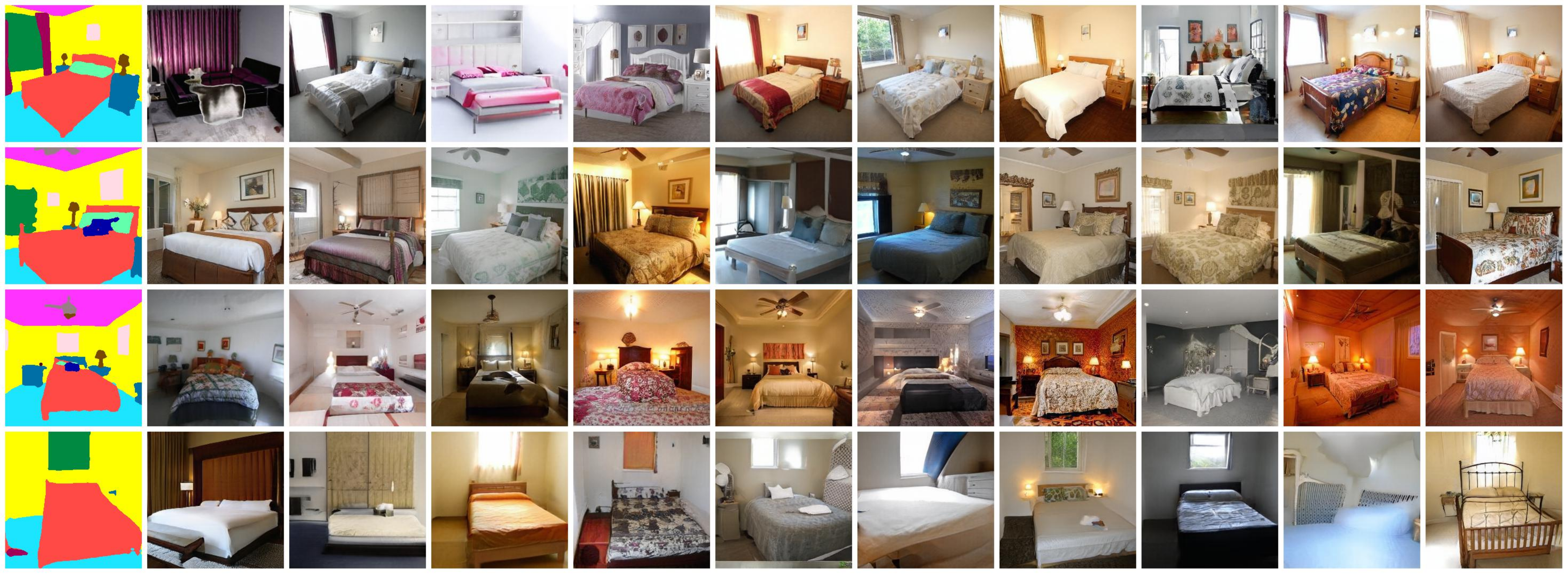}}
    \\
    \subfloat[SCS(DeepLabV3)]{\includegraphics[width=0.99\linewidth]{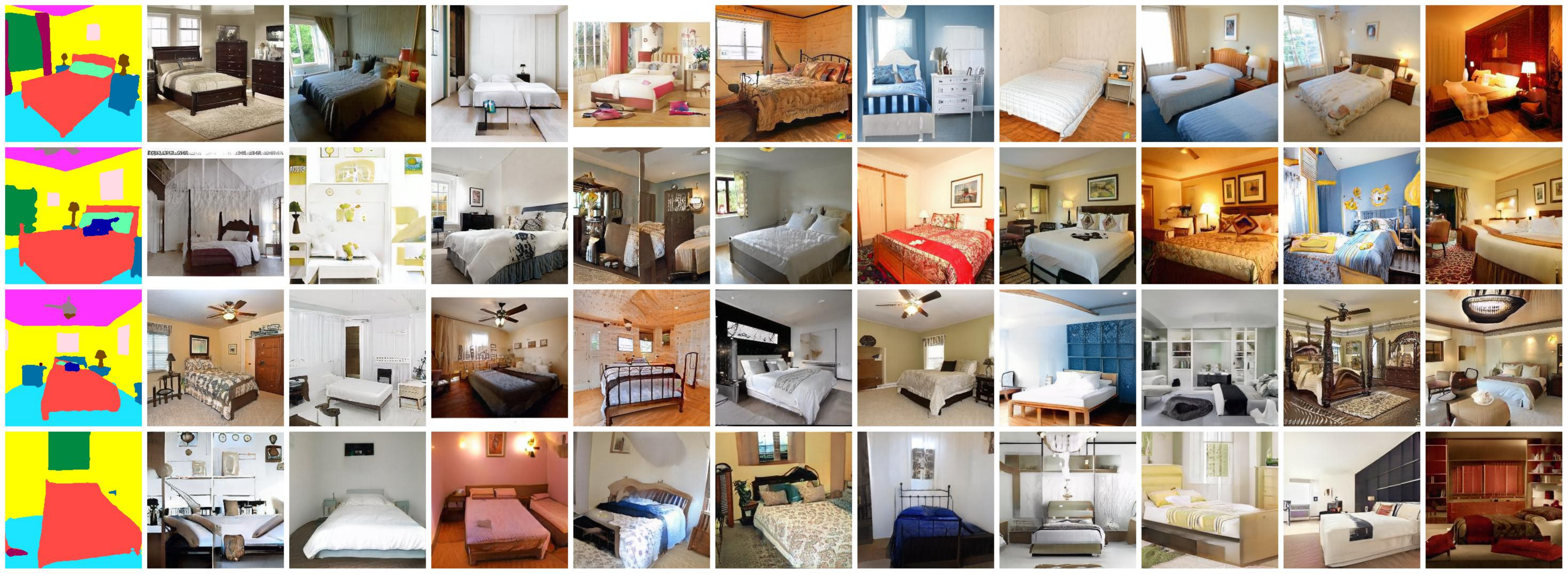}}
    \caption{The results of SCS on StyleGAN2-Bedroom using LSEs and DeepLabV3.}
    \label{fig:scs_bedroom_supp}
\end{figure*}

\begin{figure*}
    \centering
    \subfloat[SCS(1-shot LSE)]{\includegraphics[trim=0 0 810 0,clip,width=0.49\linewidth]{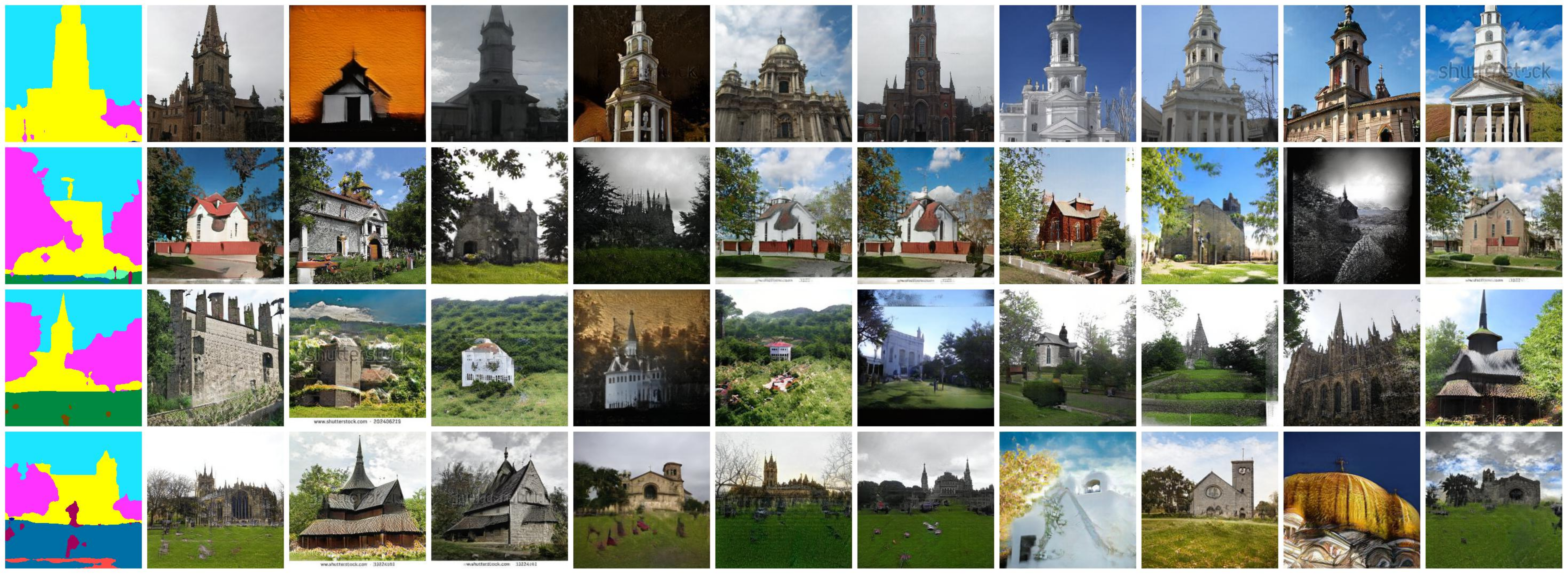}}
    ~
    \subfloat[SCS(4-shot LSE)]{\includegraphics[trim=0 0 810 0,clip,width=0.49\linewidth]{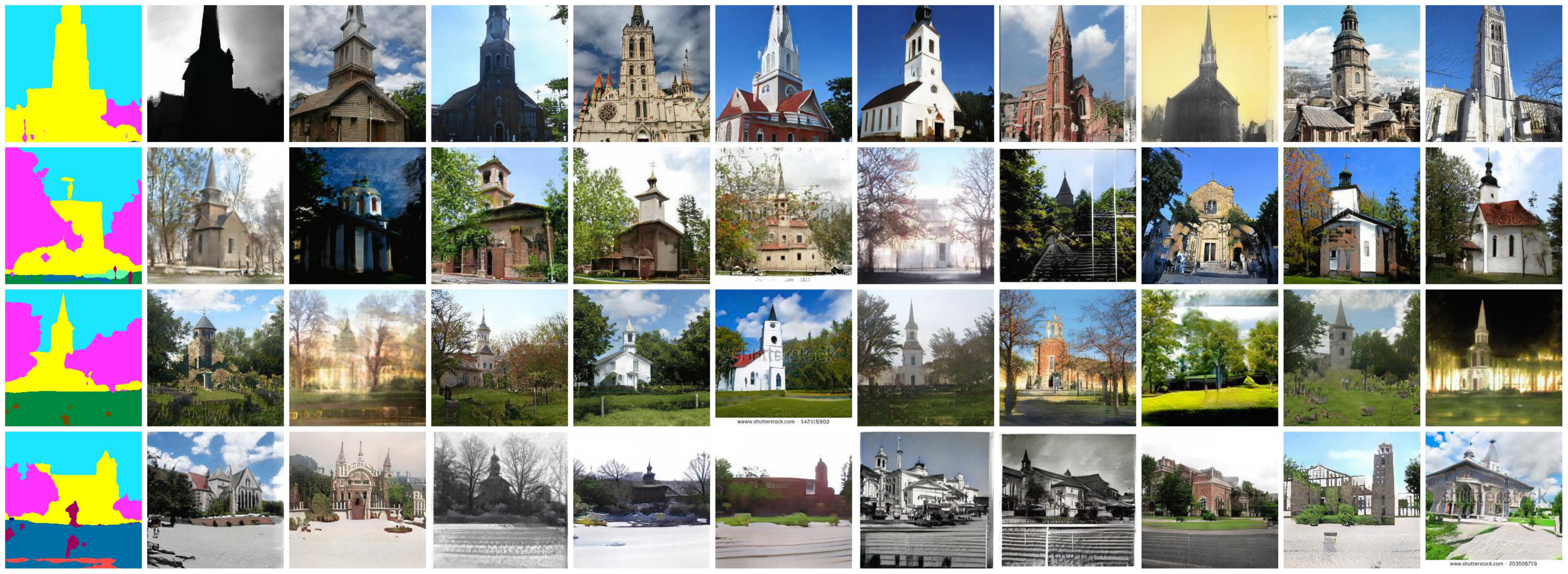}}
    \\
    \subfloat[SCS(8-shot LSE)]{\includegraphics[trim=0 0 810 0,clip,width=0.49\linewidth]{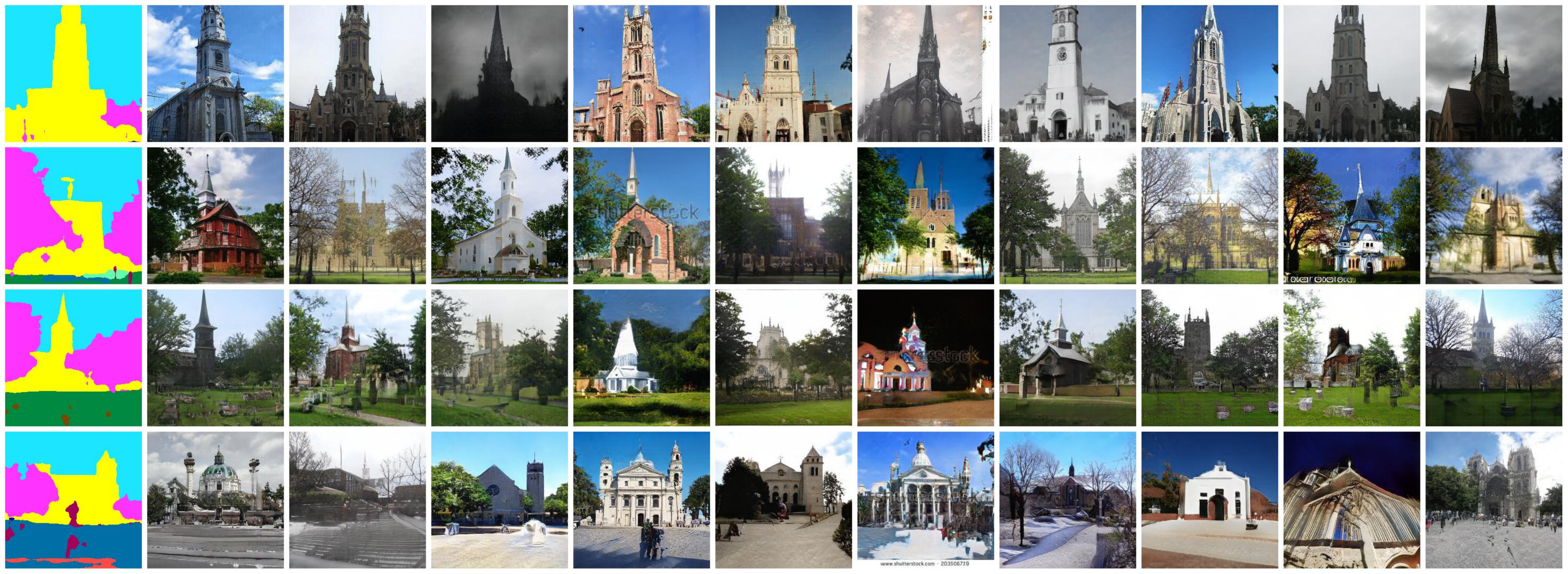}}
    ~
    \subfloat[SCS(16-shot LSE)]{\includegraphics[trim=0 0 810 0,clip,width=0.49\linewidth]{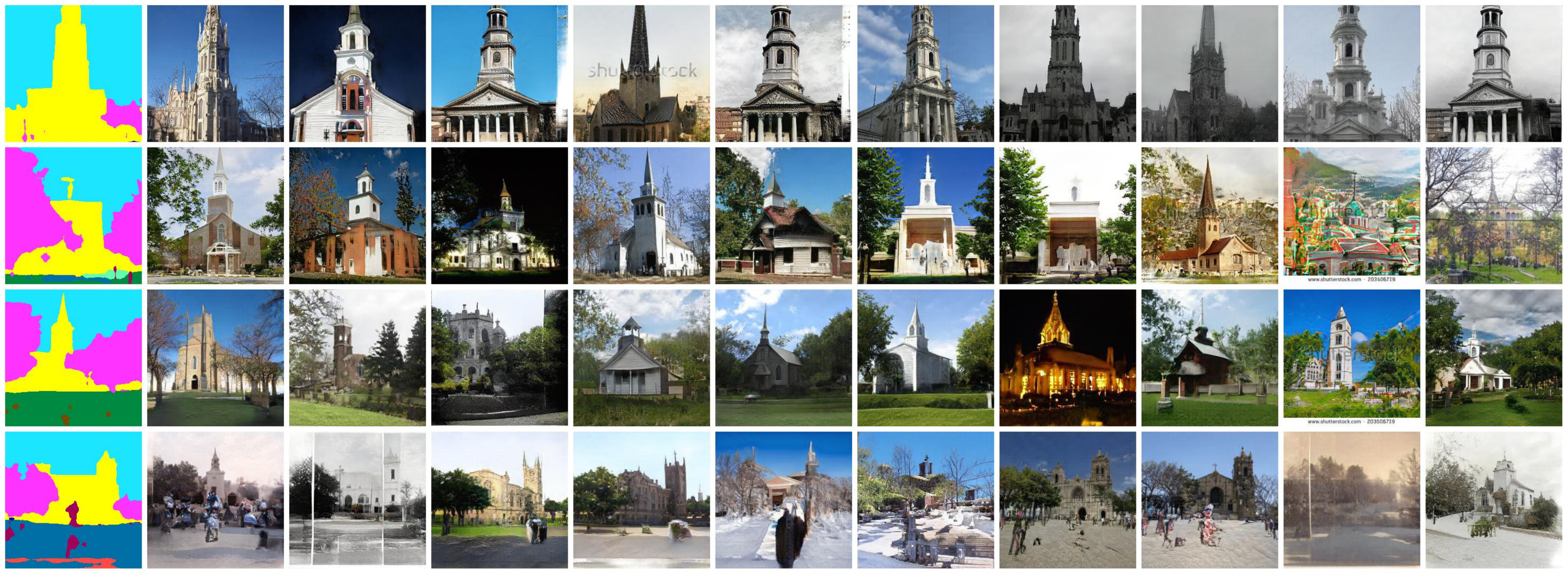}}
    \\
    \subfloat[SCS(DeepLabV3)]{\includegraphics[width=0.99\linewidth]{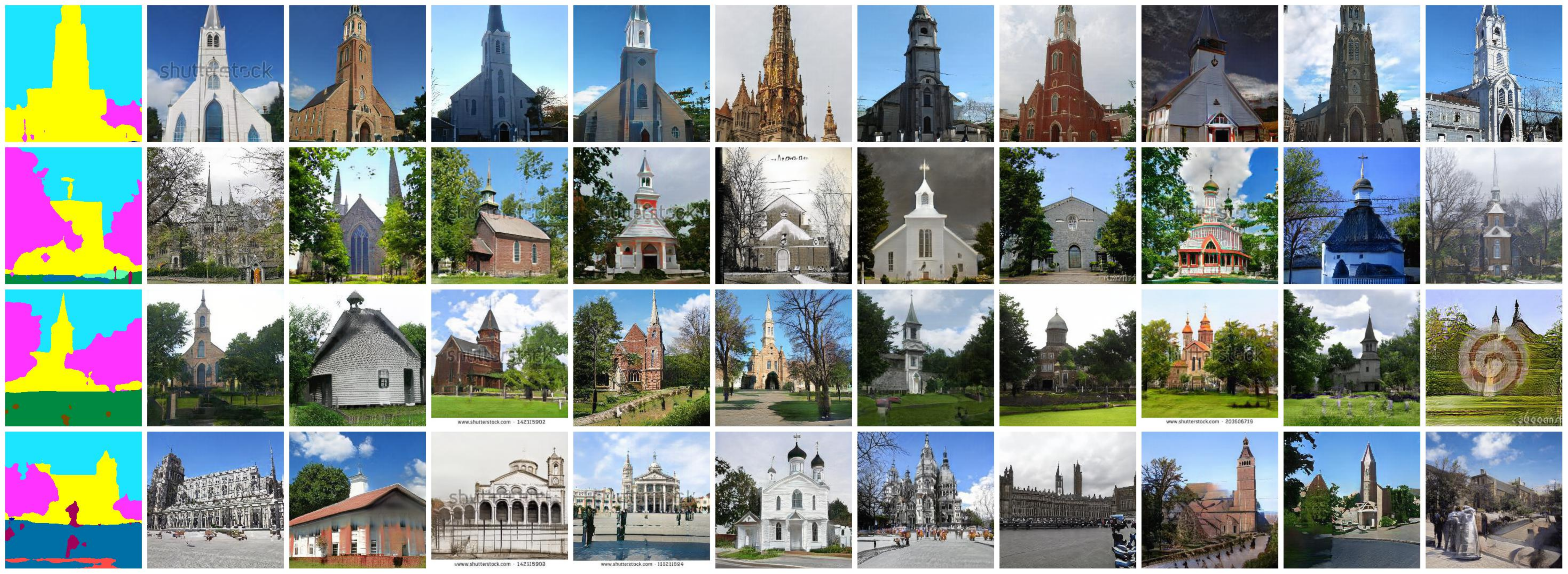}}
    \caption{The results of SCS on StyleGAN2-Church using LSEs and DeepLabV3.}
    \label{fig:scs_church_supp}
\end{figure*}

We present supplementary results for Semantic Conditional Sampling on facial images (\figref{scs_ffhq_supp}), bedroom images(\figref{scs_bedroom_supp}) and church images(\figref{scs_church_supp}).

\end{document}